\documentclass[english]{article}
\usepackage[preprint,nonatbib]{nips_2018_wider_nonotice}

\usepackage[T1]{fontenc}
\usepackage[latin9]{inputenc}
\usepackage{color,colortbl}
\usepackage{babel}
\usepackage{verbatim}
\usepackage{url}
\usepackage{amsmath}
\usepackage{amssymb}
\usepackage{graphicx}
\usepackage{setspace}
\usepackage{hyperref}
\usepackage{cancel}
\hypersetup{unicode=true,pdfusetitle,breaklinks=false}
\usepackage[hyperpageref]{backref}
\usepackage{appendix}

\graphicspath{{figures/}}

\usepackage[labelfont=bf]{caption}
\renewcommand{\arraystretch}{1.5}

\makeatletter
\newcommand{\be}{\begin{equation}}
\newcommand{\ee}{\end{equation}}

\allowdisplaybreaks \numberwithin{equation}{section}
\makeatother

\title{Scaling Laws for Neural Language Models}

\author{
    Jared Kaplan \thanks{Equal contribution. \newline \newline
    Contributions: Jared Kaplan and Sam McCandlish led the research.  Tom Henighan contributed the LSTM experiments.  Tom Brown, Rewon Child, and Scott Gray, and Alec Radford developed the optimized Transformer implementation.  Jeff Wu, Benjamin Chess, and Alec Radford developed the text datasets.  Dario Amodei provided guidance throughout the project. \newline \newline
    }\\
    Johns Hopkins University, OpenAI \\
    \texttt{jaredk@jhu.edu} \\
\And
    Sam McCandlish$^{\ast}$\\
    OpenAI \\
    \texttt{sam@openai.com} \\
\AND
    Tom Henighan \\
    OpenAI \\
    \texttt{henighan@openai.com} \\
\And
    Tom B. Brown \\
    OpenAI \\
    \texttt{tom@openai.com} \\
\And
    Benjamin Chess\\
    OpenAI \\
    \texttt{bchess@openai.com} \\
\And
    Rewon Child\\
    OpenAI \\
    \texttt{rewon@openai.com} \\
\And
    Scott Gray\\
    OpenAI \\
    \texttt{scott@openai.com} \\
\And
    Alec Radford\\
    OpenAI \\
    \texttt{alec@openai.com} \\
\And
    Jeffrey Wu\\
    OpenAI \\
    \texttt{jeffwu@openai.com} \\
\And
    Dario Amodei \\
    OpenAI \\
    \texttt{damodei@openai.com} \\
}

\begin{document}
\maketitle

\begin{abstract}
We study empirical scaling laws for language model performance on the cross-entropy loss.
The loss scales as a power-law with model size, dataset size, and the amount of compute used for training, with some trends spanning more than seven orders of magnitude.
Other architectural details such as network width or depth have minimal effects within a wide range.
Simple equations govern the dependence of overfitting on model/dataset size and the dependence of training speed on model size.
These relationships allow us to determine the optimal allocation of a fixed compute budget.
Larger models are significantly more sample-efficient, such that optimally compute-efficient training involves training very large models on a relatively modest amount of data and stopping significantly before convergence.

\end{abstract}

\newpage\tableofcontents{}

\section{Introduction}

Language provides a natural domain for the study of artificial intelligence, as the vast majority of reasoning tasks can be efficiently expressed and evaluated in language, and the world's text provides a wealth of data for unsupervised learning via generative modeling.  Deep learning has recently seen rapid progress in language modeling, with state of the art models \cite{radford2018improving,1810.04805,1906.08237,DBLP:journals/corr/abs-1907-11692, 1910.10683} approaching human-level performance on many specific tasks \cite{wang2019superglue}, including the composition of coherent multi-paragraph prompted text samples \cite{radford2019language}.

One might expect language modeling performance to depend on model architecture, the size of neural models, the computing power used to train them, and the data available for this training process.  In this work we will empirically investigate the dependence of language modeling loss on all of these factors, focusing on the Transformer architecture \cite{OriginalTransformer,liu2018generating}.  The high ceiling and low floor for performance on language tasks allows us to study trends over more than seven orders of magnitude in scale.

Throughout we will observe precise power-law scalings for performance as a function of training time, context length, dataset size, model size, and compute budget.  

\begin{figure}
\noindent \centering{} 
\includegraphics[width=\textwidth]{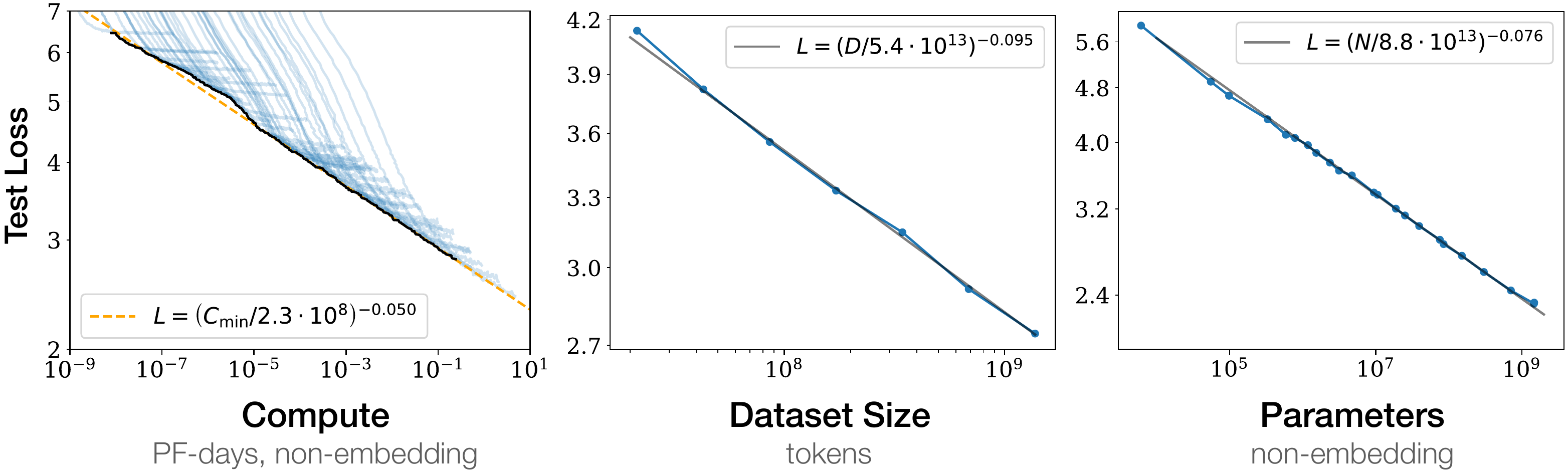}

\caption[Summary of simple power laws.]{Language modeling performance improves smoothly as we increase the model size, datasetset size, and amount of compute\footnotemark~used for training.  For optimal performance all three factors must be scaled up in tandem. Empirical performance has a power-law relationship with each individual factor when not bottlenecked by the other two.  \label{fig:BasicPowerLaws}}
\end{figure}

\subsection{Summary}

\footnotetext{Here we display predicted compute when using a sufficiently small batch size.  See Figure \ref{fig:ComputeEfficientAdjusted} for comparison to the purely empirical data.}

Our key findings for Transformer language models are are as follows:

\paragraph{Performance depends strongly on scale, weakly on model shape:} Model performance depends most strongly on scale, which consists of three factors: the number of model parameters $N$ (excluding embeddings), the size of the dataset $D$, and the amount of compute $C$ used for training.  Within reasonable limits, performance depends very weakly on other architectural hyperparameters such as depth vs.~width. (Section \ref{sec:Empirical})

\paragraph{Smooth power laws:} Performance has a power-law relationship with each of the three scale factors $N, D, C$ when not bottlenecked by the other two, with trends spanning more than six orders of magnitude (see Figure \ref{fig:BasicPowerLaws}). We observe no signs of deviation from these trends on the upper end, though performance must flatten out eventually before reaching zero loss. (Section \ref{sec:Empirical})

\paragraph{Universality of overfitting:} Performance improves predictably as long as we scale up $N$ and $D$ in tandem, but enters a regime of diminishing returns if either $N$ or $D$ is held fixed while the other increases.  The performance penalty depends predictably on the ratio $N^{0.74}/D$, meaning that every time we increase the model size 8x, we only need to increase the data by roughly 5x to avoid a penalty. (Section \ref{sec:ChartingOverfitting})

\paragraph{Universality of training:} Training curves follow predictable power-laws whose parameters are roughly independent of the model size.  By extrapolating the early part of a training curve, we can roughly predict the loss that would be achieved if we trained for much longer. (Section \ref{sec:ScalingSizeandSteps})

\paragraph{Transfer improves with test performance:} When we evaluate models on text with a different distribution than they were trained on, the results are strongly correlated to those on the training validation set with a roughly constant offset in the loss -- in other words, transfer to a different distribution incurs a constant penalty but otherwise improves roughly in line with performance on the training set. (Section \ref{sec:GeneralizationtoOtherDistributions})

\begin{figure}
\noindent \centering{} \includegraphics[width=0.99\textwidth]{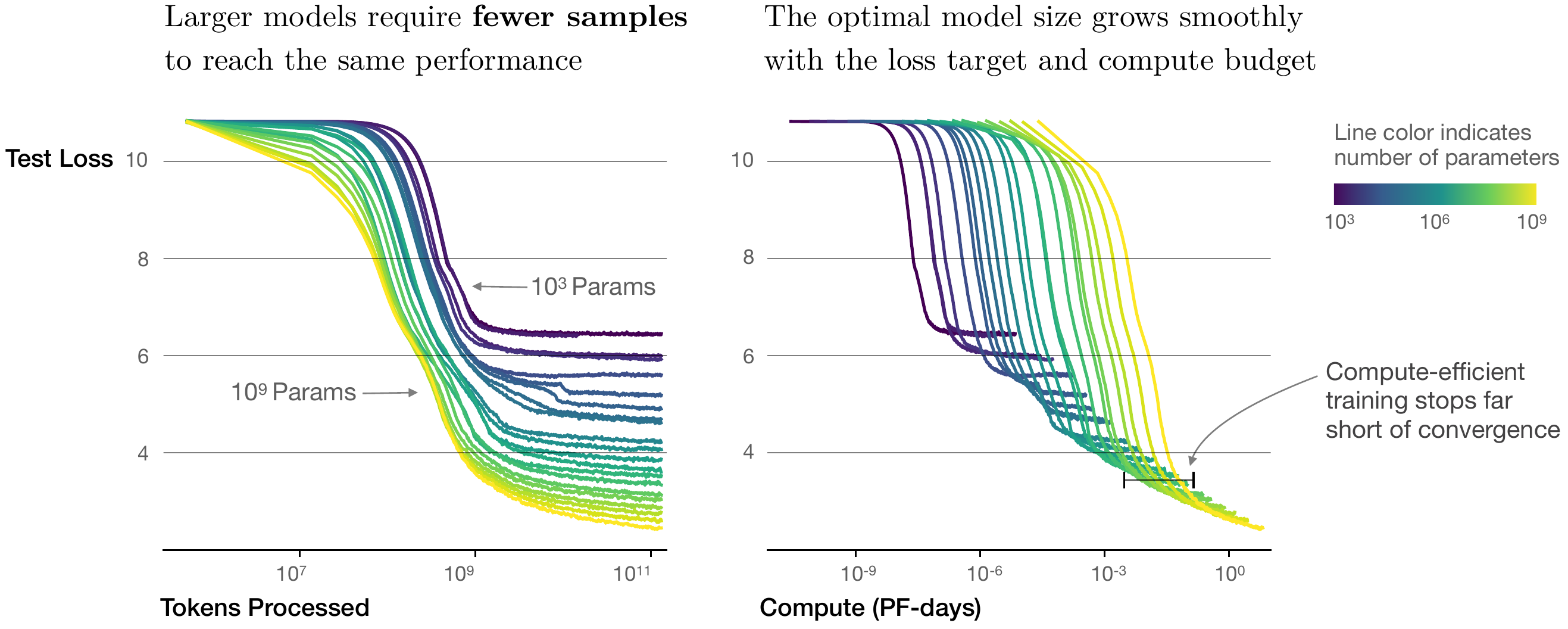}
\caption[Illustration of sample efficiency and compute efficiency.]{We show a series of language model training runs, with models ranging in size from $10^3$ to $10^9$ parameters (excluding embeddings). \label{fig:EfficiencyIllustration}}
\end{figure}

\paragraph{Sample efficiency:} Large models are  more sample-efficient than small models, reaching the same level of performance with fewer optimization steps (Figure \ref{fig:EfficiencyIllustration}) and using  fewer data points (Figure \ref{fig:LossvsModelDatasetSize}).

\begin{figure}
\noindent \centering{} \includegraphics[height=0.32\textwidth]{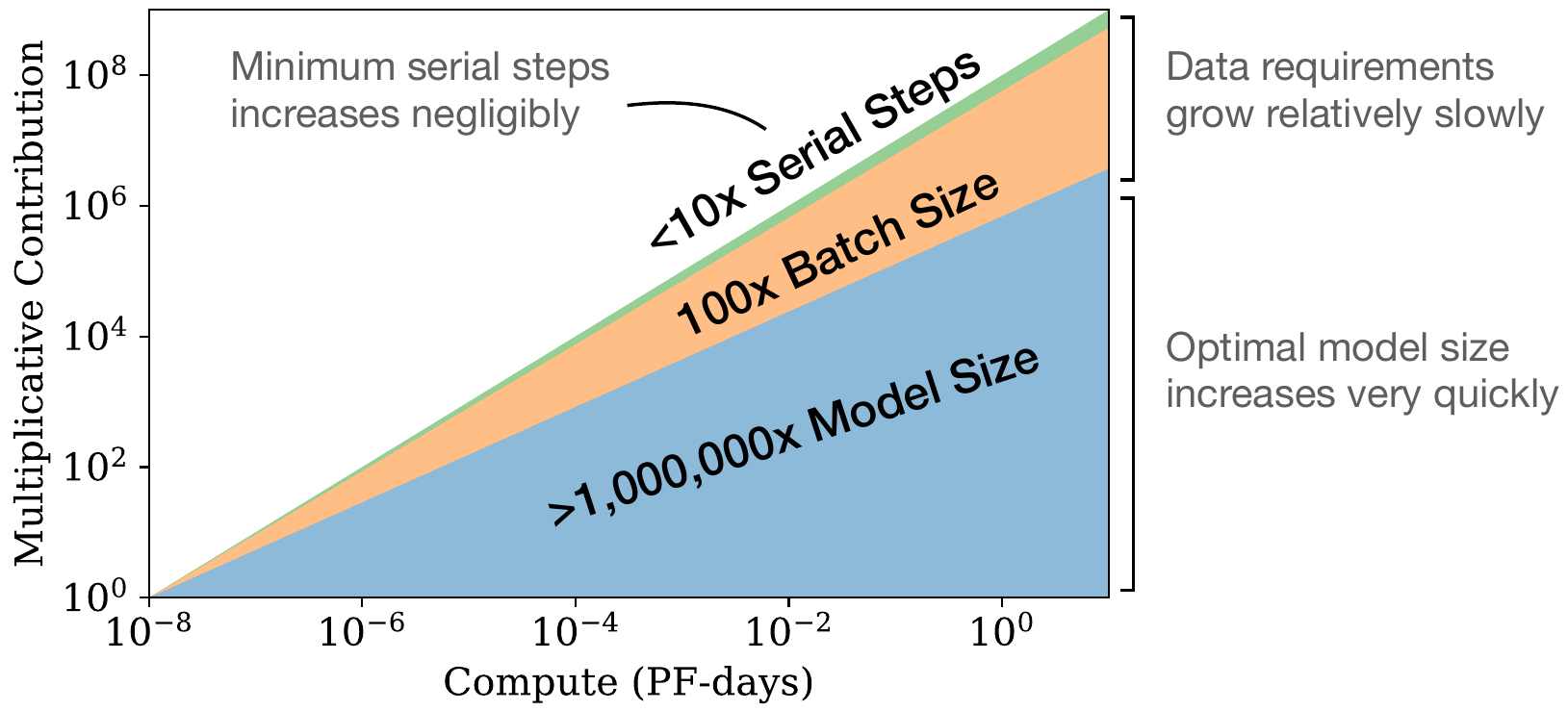}
\caption[How to scale up model size, batch size, and serial steps]{As more compute becomes available, we can choose how much to allocate towards training larger models, using larger batches, and training for more steps.  We illustrate this for a billion-fold increase in compute.  For optimally compute-efficient training, most of the increase should go towards increased model size.  A relatively small increase in data is needed to avoid reuse.  Of the increase in data, most can be used to increase parallelism through larger batch sizes, with only a very small increase in serial training time required.  \label{fig:ContributionIllustration} }
\end{figure}

\paragraph{Convergence is inefficient:} When working within a fixed compute budget $C$ but without any other restrictions on the model size $N$ or available data $D$, we attain optimal performance by training \emph{very large models} and stopping \emph{significantly short of convergence} (see Figure \ref{fig:ContributionIllustration}).  Maximally compute-efficient training would therefore be far more sample efficient than one might expect based on training small models to convergence, with data requirements growing very slowly as $D \sim C^{0.27}$ with training compute. (Section \ref{sec:OptimalCompute})

\paragraph{Optimal batch size:} The ideal batch size for training these models is roughly a power of the loss only, and continues to be determinable by measuring the gradient noise scale \cite{1812.06162}; it is roughly 1-2 million tokens at convergence for the largest models we can train. (Section \ref{sec:OptimalBatchSize})

Taken together, these results show that language modeling performance improves smoothly and predictably as we appropriately scale up model size, data, and compute.  We expect that larger language models will perform better and be more sample efficient than current models.

\subsection{Summary of Scaling Laws}

The test loss of a Transformer trained to autoregressively model language can be predicted using a power-law  when performance is limited by only either the number of non-embedding parameters $N$, the dataset size $D$, or the optimally allocated compute budget $C_{\rm min}$ (see Figure \ref{fig:BasicPowerLaws}):
\begin{enumerate}
\setlength\itemsep{0.5em}
\item For models with a limited number of parameters, trained to convergence on sufficiently large datasets:
\be L(N) = \left(N_{\mathrm{c}}/N\right)^{\alpha_N};~~ \alpha_N \sim 0.076, \quad N_{\mathrm{c}} \sim 8.8 \times 10^{13}~\text{(non-embedding parameters)} \label{eq:crit_n} \ee
\item For large models trained with a limited dataset with early stopping:
\be L(D) = \left(D_{\mathrm{c}}/D\right)^{\alpha_D};~~ \alpha_D \sim 0.095, \quad D_{\mathrm{c}} \sim 5.4 \times 10^{13}~\text{(tokens)} \label{eq:crit_d} \ee
\item When training with a limited amount of compute, a sufficiently large dataset, an optimally-sized model, and a sufficiently small batch size (making optimal\footnote{We also observe an empirical power-law trend with the training compute $C$ (Figure \ref{fig:BasicPowerLaws}) while training at fixed batch size, but it is the trend with $C_{\rm min}$ that should be used to make predictions.  They are related by equation \eqref{eq:AdjustedCompute}. } use of compute):
\be L(C_{\rm min}) = \left(C_{\mathrm{c}}^{\rm min} / C_{\rm min}\right)^{\alpha_C^{\rm min}};~~ \alpha_C^{\rm min} \sim 0.050, \quad C_{\mathrm{c}}^{\rm min} \sim 3.1 \times 10^{8}~\text{(PF-days)} \label{eq:crit_c} \ee
\end{enumerate}

These relations hold across eight orders of magnitude in $C_{\rm min}$, six orders of magnitude in $N$, and over two orders of magnitude in $D$.  They depend very weakly on model shape and other Transformer hyperparameters (depth, width, number of self-attention heads), with specific numerical values associated with the Webtext2 training set \cite{radford2019language}.   The power laws $\alpha_{\rm N}, \alpha_{\rm D}, \alpha_{C}^{\rm min}$ specify the degree of performance improvement expected as we scale up $N$, $D$, or $C_{\rm min}$; for example, doubling the number of parameters yields a loss that is smaller by a factor $2^{-\alpha_N}=0.95$. The precise numerical values of $N_{\mathrm{c}}, C_{\rm c}^{\rm min},$ and $D_{\mathrm{c}}$ depend on the vocabulary size and tokenization and hence do not have a fundamental meaning.

The critical batch size, which determines the speed/efficiency tradeoff for data parallelism (\cite{1812.06162}), also roughly obeys a power law in $L$:
\begin{equation}
B_{\rm crit}\left(L\right)=\frac{B_{\ast}}{L^{1/\alpha_{B}}},\qquad B_{\ast} \sim 2\cdot 10^8 \text{ tokens},\ \ \alpha_{B} \sim 0.21
\label{eq:critical-batch-size}
\end{equation}

\begin{figure}
\noindent \centering{} 
\includegraphics[width=0.48\textwidth]{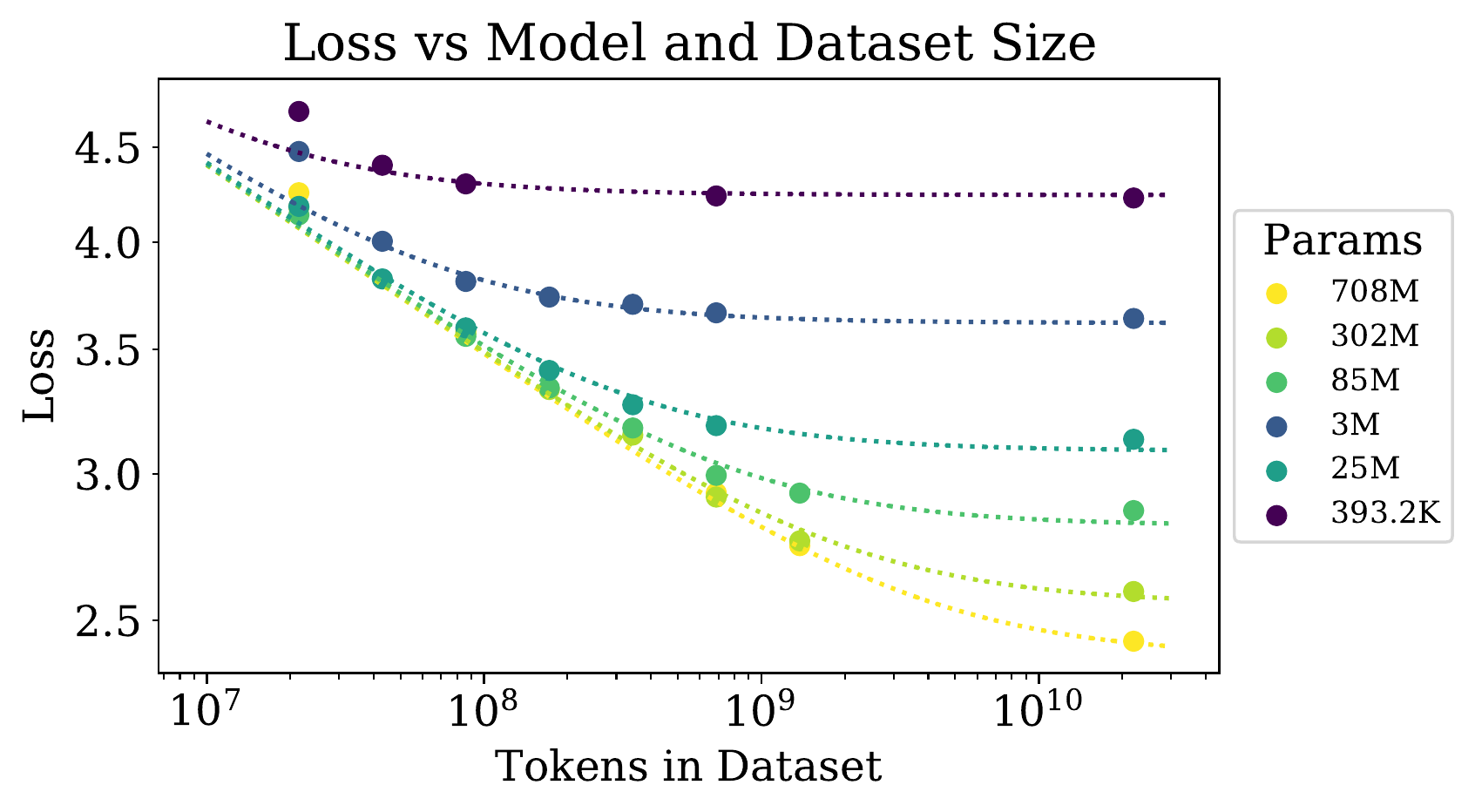}
\includegraphics[width=0.48\textwidth]{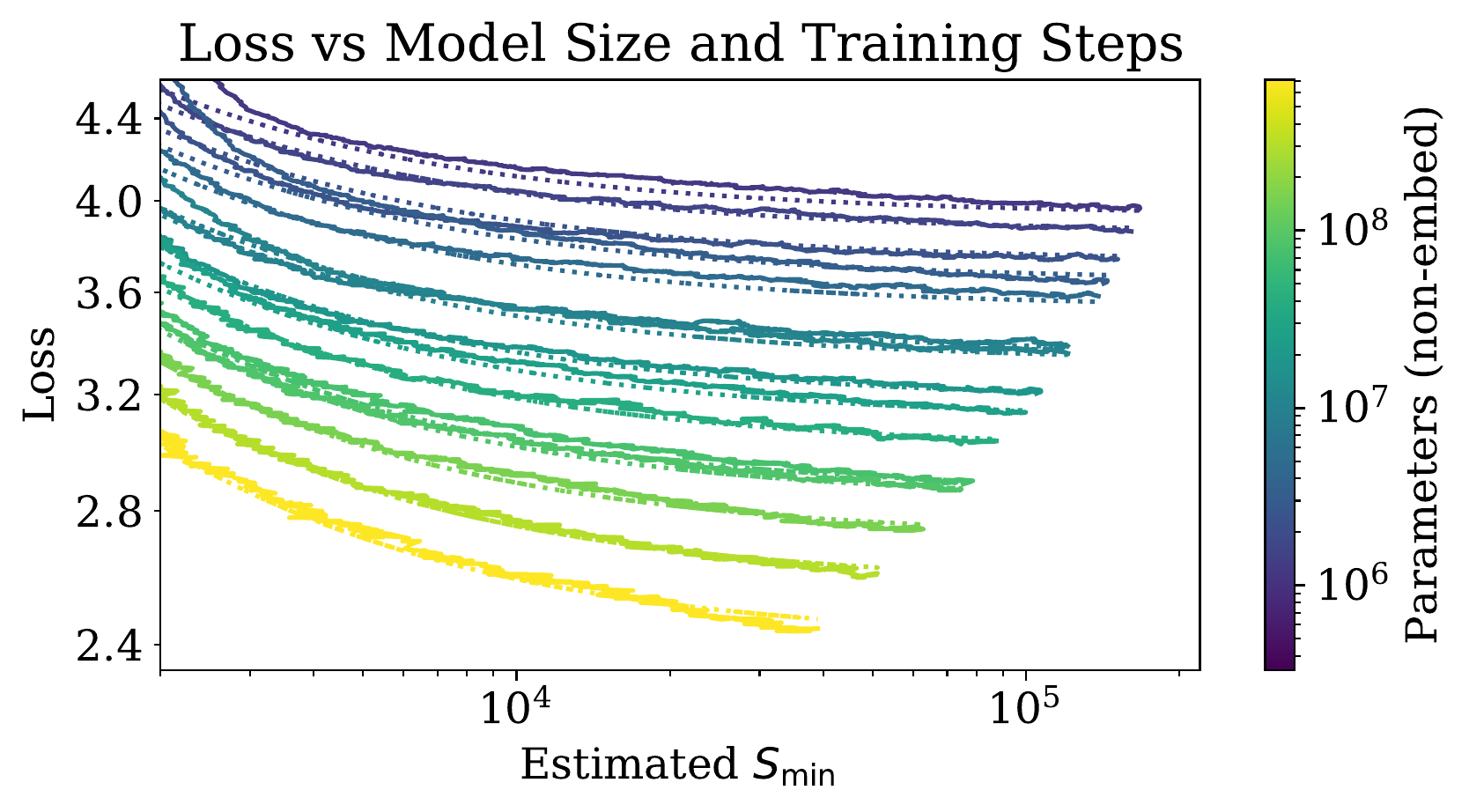}
\caption[Performance when varying model and data size, or model and training steps, simultaneously]{
{\bf Left}: The early-stopped test loss $L(N, D)$ varies predictably with the dataset size $D$ and model size $N$ according to Equation \eqref{eq:FundamentalLikelihioodvsModelandDataSize}.  
{\bf Right}:  After an initial transient period, learning curves for all model sizes $N$ can be fit with Equation \eqref{eq:FundamentalLikelihioodvsModelandSteps}, which is parameterized in terms of $S_{\rm min}$, the number of steps when training at large batch size (details in Section \ref{sec:OptimalBatchSize}).  \label{fig:LearningCurveFitsandResiduals}
\label{fig:LossvsModelDatasetSize}}
\end{figure}

Equation \eqref{eq:crit_n} and \eqref{eq:crit_d} together suggest that as we increase the model size, we should increase the dataset size sublinearly according to $D \propto N^{\frac{\alpha_N}{\alpha_D}} \sim N^{0.74}$. In fact, we find that there is a single equation combining \eqref{eq:crit_n} and \eqref{eq:crit_d} that governs the simultaneous dependence on $N$ and $D$ and governs the degree of overfitting:
\be
\label{eq:FundamentalLikelihioodvsModelandDataSize}
L(N, D) 
= \left[ \left( \frac{N_c}{N} \right)^{\frac{\alpha_N}{\alpha_D}} + \frac{D_c}{D}  \right]^{\alpha_D}
\ee
with fits pictured on the left in figure \ref{fig:LossvsModelDatasetSize}.  We conjecture that this functional form may also parameterize the trained log-likelihood for other generative modeling tasks.

When training a given model for a finite number of parameter update steps $S$ in the infinite data limit, after an initial transient period, the learning curves can be accurately fit by  (see the right of figure \ref{fig:LearningCurveFitsandResiduals})
\be
\label{eq:FundamentalLikelihioodvsModelandSteps}
L(N, S) = \left( \frac{N_c}{N} \right)^{\alpha_N}  + \left( \frac{S_c }{S_{\rm min}(S)} \right)^{\alpha_S}
\ee
where $S_c \approx 2.1 \times 10^3$ and $\alpha_S \approx 0.76$, and $S_{\rm min}(S)$ is the minimum possible number of optimization steps (parameter updates)
estimated  using Equation \eqref{eq:AdjustedSteps}.

When training within a fixed compute budget $C$, but with no other constraints, Equation \eqref{eq:FundamentalLikelihioodvsModelandSteps} leads to the prediction that the optimal model size $N$, optimal batch size $B$, optimal number of steps $S$, and dataset size $D$ should grow as 
\be
\label{eq:OptimalModelSizeTrainingTimevsCompute}
N \propto C^{\alpha_{C}^{\rm min} /\alpha_{N}}, \quad
B \propto C^{\alpha_C^{\rm min} / \alpha_B}, \quad
S \propto C^{\alpha_C^{\rm min} / \alpha_S}, \quad
D = B\cdot S \quad
\ee
with
\be
\alpha_{C}^{\rm min} = 1/\left(1/\alpha_{S}+1/\alpha_{B}+1/\alpha_{N}\right) 
\ee
which closely matches the empirically optimal results $N \propto C_{\rm min}^{0.73}$, $B \propto C_{\rm min}^{0.24}$, and $S \propto C_{\rm min}^{0.03}$.  As the computational budget $C$ increases, it should be spent primarily on larger models, without dramatic increases in training time or dataset size (see Figure \ref{fig:ContributionIllustration}).  This also implies that as models grow larger, they  become increasingly sample efficient.  In practice, researchers typically train  smaller models for  longer than would be maximally compute-efficient because of  hardware constraints.
Optimal performance depends on total compute as a power law (see Equation \eqref{eq:crit_c}).

We  provide some basic theoretical motivation for Equation \eqref{eq:FundamentalLikelihioodvsModelandDataSize}, an analysis of learning curve fits and their implications for training time, and a breakdown of our results per token.  We also make some brief comparisons to LSTMs and recurrent Transformers \cite{DBLP:journals/corr/abs-1807-03819}.

\subsection{Notation}

We use the following notation:
\begin{itemize}
\item $L$ -- the cross entropy loss in nats.  Typically it will be averaged over the tokens in a context, but in some cases we report the loss for specific tokens within the context.
\item $N$ -- the number of model parameters, \emph{excluding all vocabulary and positional embeddings}  
\item $C \approx 6 N B S$ -- an estimate of the total non-embedding training compute, where $B$ is the batch size, and $S$ is the number of training steps (ie parameter updates). We quote numerical values in PF-days, where one PF-day $ = 10^{15} \times 24 \times 3600 = 8.64 \times 10^{19}$ floating point operations.  
\item $D$ -- the dataset size in tokens
\item $B_{\rm crit}$ -- the critical batch size \cite{1812.06162}, defined and discussed in Section \ref{sec:OptimalBatchSize}.  Training at the critical batch size provides a roughly optimal compromise between time and compute efficiency.
\item $C_{\rm min}$ -- an estimate of the minimum amount of non-embedding compute to reach a given value of the loss.  This is the training compute that would be used if the model were trained at a batch size much less than the critical batch size.  
\item $S_{\rm min}$ -- an estimate of the minimal number of training steps needed to reach a given value of the loss.  This is also the number of training steps that would be used if the model were trained at a batch size much greater than the critical batch size.
\item $\alpha_X$ -- power-law exponents for the scaling of the loss as $L(X) \propto 1/X^{\alpha_X}$ where $X$ can be any of $N, D, C, S, B, C^{\rm min}$.
\end{itemize}

\section{Background and Methods}

We train language models on WebText2, an extended version of the WebText \cite{radford2019language} dataset, tokenized using byte-pair encoding \cite{BPE} with a vocabulary size $n_{\rm vocab} = 50257$.  We optimize the autoregressive log-likelihood (i.e. cross-entropy loss) averaged over a 1024-token context, which is also our principal performance metric.  We record the loss on the WebText2 test distribution and on a selection of other text distributions.  We primarily train decoder-only \cite{liu2018generating, radford2018improving} Transformer \cite{OriginalTransformer} models, though we also train LSTM models and Universal Transformers \cite{DBLP:journals/corr/abs-1807-03819}  for comparison.

\subsection{Parameter and Compute Scaling of Transformers}
\label{sec:ParameterComputeCounts}

\begin{table}[t!]
\centering
\begin{tabular}{|l|l|l|}
\hline 
\textbf{Operation}  & \textbf{Parameters}  & \textbf{FLOPs per Token}\tabularnewline
\hline 
\hline 
Embed  & $\left(n_{{\rm vocab}} + n_{{\rm ctx}}\right)d_{{\rm model}}$  & $4d_{{\rm model}}$\tabularnewline
\hline 
Attention: QKV  & $n_{{\rm layer}}d_{{\rm model}}3d_{{\rm attn}}$  & $2n_{{\rm layer}}d_{{\rm model}}3d_{{\rm attn}}$\tabularnewline
\hline 
Attention: Mask & ---  & $2n_{{\rm layer}}n_{{\rm ctx}}d_{{\rm attn}}$\tabularnewline
\hline 
Attention: Project & $n_{{\rm layer}}d_{{\rm attn}}d_{{\rm model}}$  & $2n_{{\rm layer}}d_{{\rm attn}}d_{{\rm embd}}$\tabularnewline
\hline 
Feedforward  & $n_{{\rm layer}}2d_{{\rm model}}d_{{\rm ff}}$ & $2n_{{\rm layer}}2d_{{\rm model}}d_{{\rm ff}}$\tabularnewline
\hline 
De-embed  & ---  & $2d_{{\rm model}}n_{{\rm vocab}}$\tabularnewline
\hline 
\hline 
\textbf{Total (Non-Embedding)} & $N=2d_{{\rm model}}n_{{\rm layer}}\left(2d_{{\rm attn}}+d_{{\rm ff}}\right)$  & $C_{\mathrm{forward}}=2N+2n_{{\rm layer}}n_{{\rm ctx}}d_{{\rm attn}}$\tabularnewline
\hline 
\end{tabular}
\vspace{1em}
\caption[Parameter and compute counts for Transformer]{Parameter counts and compute (forward pass) estimates for a Transformer model.  Sub-leading terms such as nonlinearities, biases, and layer normalization are omitted. \label{tab:TableTransformerParamsFLOPs}}
\end{table}

 We parameterize the Transformer architecture using hyperparameters $n_{\rm layer}$ (number of layers), $d_{{\rm model}}$ (dimension of the residual stream), $d_{\rm ff}$ (dimension of the intermediate feed-forward layer), $d_{\rm attn}$ (dimension of the attention output), and $n_{\rm heads}$ (number of attention heads per layer).  We include $n_{\rm ctx}$ tokens in the input context, with $n_{\rm ctx} = 1024$ except where otherwise noted.

We use $N$ to denote the model size, which we define as the number of \emph{non-embedding} parameters
\begin{align}
\label{eq:ModelSizeDefinition}
N & \approx  2d_{{\rm model}}n_{{\rm layer}}\left(2d_{{\rm attn}}+d_{{\rm ff}}\right) & \nonumber \\
& = 12 n_{\rm layer} d_{{\rm model}}^2 \quad \text{ with the standard } \quad d_{\rm attn} = d_{\rm ff}/4 = d_{{\rm model}} &
\end{align}
where we have excluded biases and other sub-leading terms.
Our models also have $n_{\rm vocab} d_{{\rm model}}$ parameters in an embedding matrix, and use $n_{\rm ctx} d_{{\rm model}}$ parameters for positional embeddings, but we do not include these when discussing the `model size' $N$; we will see that this produces significantly cleaner scaling laws.

Evaluating a forward pass of the Transformer involves roughly
\be
\label{eq:ApproximateTotalCompute}
C_{\rm forward} \approx 2N + 2n_{{\rm layer}}n_{{\rm ctx}}d_{{\rm model}}
\ee
add-multiply operations, where the factor of two comes from the multiply-accumulate operation used in matrix multiplication.  A more detailed per-operation parameter and compute count is included in Table \ref{tab:TableTransformerParamsFLOPs}.

For contexts and models with $d_{{\rm model}} > n_{\rm ctx} / 12$, the context-dependent computational cost per token is a relatively small fraction of the total compute.
Since we primarily study models where $d_{\rm model} \gg n_{\rm ctx}/12$, we do not include context-dependent terms in our training compute estimate.  Accounting for the backwards pass (approximately twice the compute as the forwards pass), we then define the estimated non-embedding compute as $C \approx 6 N$ floating point operators per training token.

\subsection{Training Procedures}

Unless otherwise noted, we train  models with the Adam optimizer \cite{kingma2014adam} for a fixed $2.5 \times 10^5$  steps with a batch size of $512$ sequences of $1024$ tokens.
Due to memory constraints, our largest models (more than 1B parameters) were trained with Adafactor \cite{DBLP:journals/corr/abs-1804-04235}.
We experimented with a variety of learning rates and schedules, as discussed in Appendix \ref{app:OptimizationDetailsandErrorAnalysis}.
We found that results at convergence were largely independent of learning rate schedule.  Unless otherwise noted, all training runs included in our data used a learning rate schedule with a 3000 step linear warmup followed by a cosine decay to zero.

\subsection{Datasets}
We train our models on an extended version of the WebText dataset described in \cite{radford2019language}.  The original WebText dataset was a web scrape of outbound links from Reddit through December 2017 which received at least 3 karma. In the second version, WebText2, we added outbound Reddit links from the period of January to October 2018, also with a minimum of 3 karma. The karma threshold served as a heuristic for whether people found the link interesting or useful. The text of the new links was extracted with the Newspaper3k python library. In total, the dataset consists of 20.3M documents containing 96 GB of text and $1.62 \times 10^{10}$ words (as defined by \texttt{wc}). We then apply the reversible tokenizer described in \cite{radford2019language}, which yields $2.29 \times 10^{10}$ tokens.  We reserve $6.6 \times 10^{8}$ of these tokens for use as a test set, and we also test on similarly-prepared samples of Books Corpus \cite{Zhu_2015}, Common Crawl \cite{commoncrawl}, English Wikipedia, and a collection of publicly-available Internet Books.

\section{Empirical Results and Basic Power Laws}
\label{sec:Empirical}

To characterize language model scaling we train a wide variety of models, varying a number of factors including:
\begingroup
\renewcommand{\arraystretch}{1.1}
\begin{itemize}
\item Model size (ranging in size from 768 to 1.5 billion non-embedding parameters)
\item Dataset size (ranging from 22 million to 23 billion tokens)
\item Shape (including depth, width, attention heads, and feed-forward dimension)
\item Context length (1024 for most runs, though we also experiment with shorter contexts)
\item Batch size ($2^{19}$ for most runs, but we also vary it to measure the critical batch size)
\end{itemize}
\endgroup

In this section we will display data along with empirically-motivated fits, deferring theoretical analysis to later sections.  

\subsection{Approximate Transformer Shape and Hyperparameter Independence}
\label{sec:ShapeIndependence}

\begin{figure}
\noindent \centering{} 
\includegraphics[width=\textwidth]{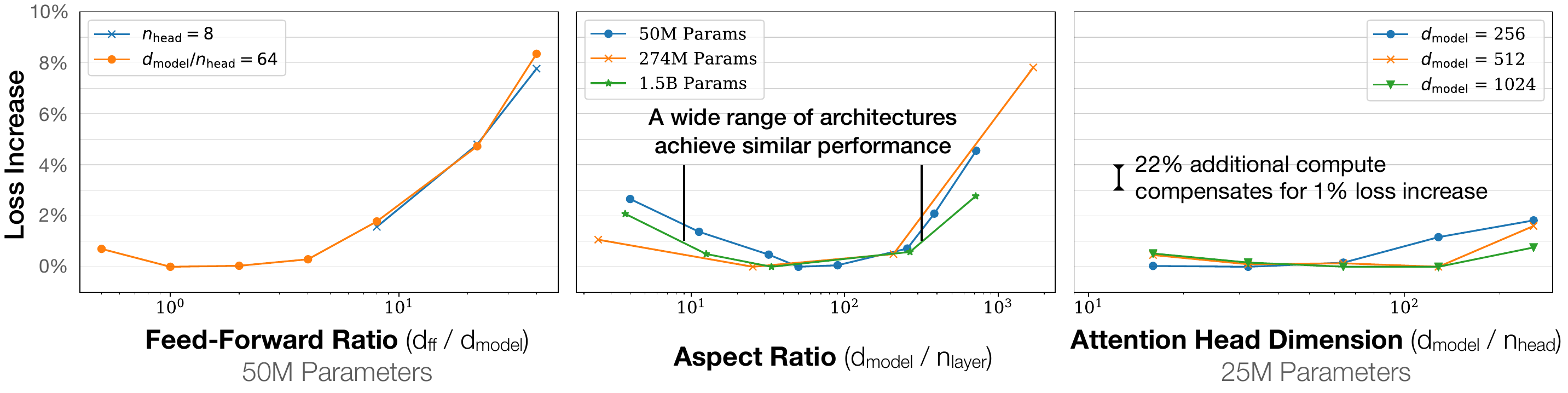}
 \caption[Weak dependence of performance on hyperparameter tuning]{Performance depends very mildly on model shape when the total number of non-embedding parameters $N$ is held fixed.  The loss varies only a few percent over a wide range of shapes.  Small differences in parameter counts are compensated for by using the fit to $L(N)$ as a baseline.  Aspect ratio in particular can vary by a factor of 40 while only slightly impacting performance; an $(n_{\mathrm{layer}}, d_{\mathrm{model}}) = (6, 4288)$ reaches a loss within 3\% of the $(48, 1600)$ model used in \cite{radford2019language}.  \label{fig:HeadsLayersIndependence}}
\end{figure}

Transformer performance depends very weakly on the shape parameters $n_{\rm layer}, n_{\rm heads}$, and $d_{\rm ff}$ when we hold the total non-embedding parameter count $N$ fixed.  
To establish these results we  trained models with fixed size while varying a single hyperparameter. This was simplest for the case of $n_{\rm heads}$.  When varying $n_{\rm layer}$, we simultaneously varied $d_{{\rm model}}$ while keeping $N \approx 12 n_{\rm layer} d_{{\rm model}}^2$ fixed.  Similarly, to vary $d_{\rm ff}$ at fixed model size we also simultaneously varied the $d_{{\rm model}}$ parameter, as required by the parameter counts in Table \ref{tab:TableTransformerParamsFLOPs}.  Independence of $n_{\rm layers}$ would follow if deeper Transformers effectively behave as ensembles of shallower models, as has been suggested for ResNets \cite{ResNetsEnsemblesShallow}.  The results are shown in Figure \ref{fig:HeadsLayersIndependence}.

\subsection{Performance with Non-Embedding Parameter Count $N$} 
\label{sec:PerformancevsModelSize}

\begin{figure}
\noindent \centering{}
\includegraphics[width=0.45\textwidth]{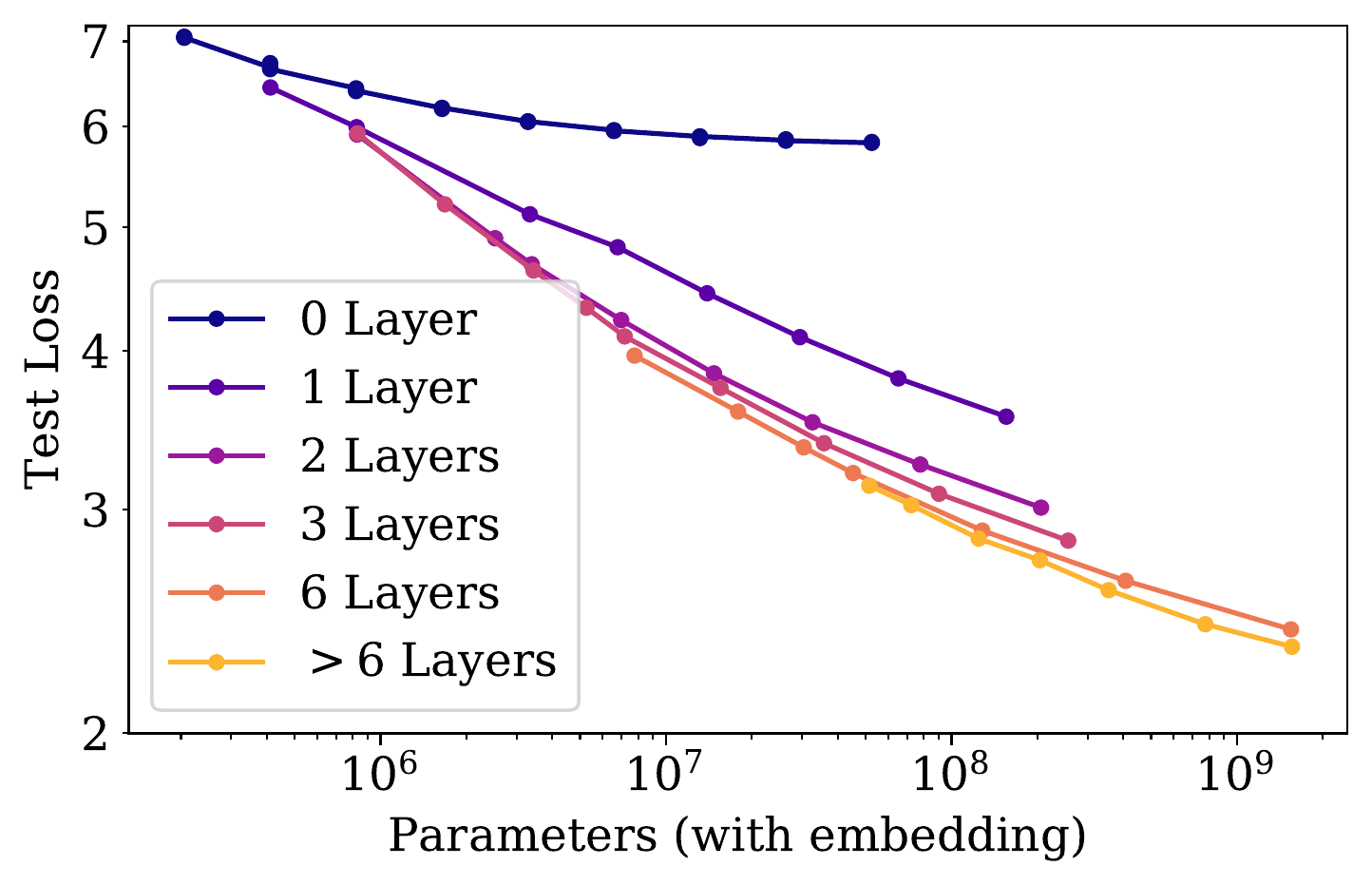}
\includegraphics[width=0.45\textwidth]{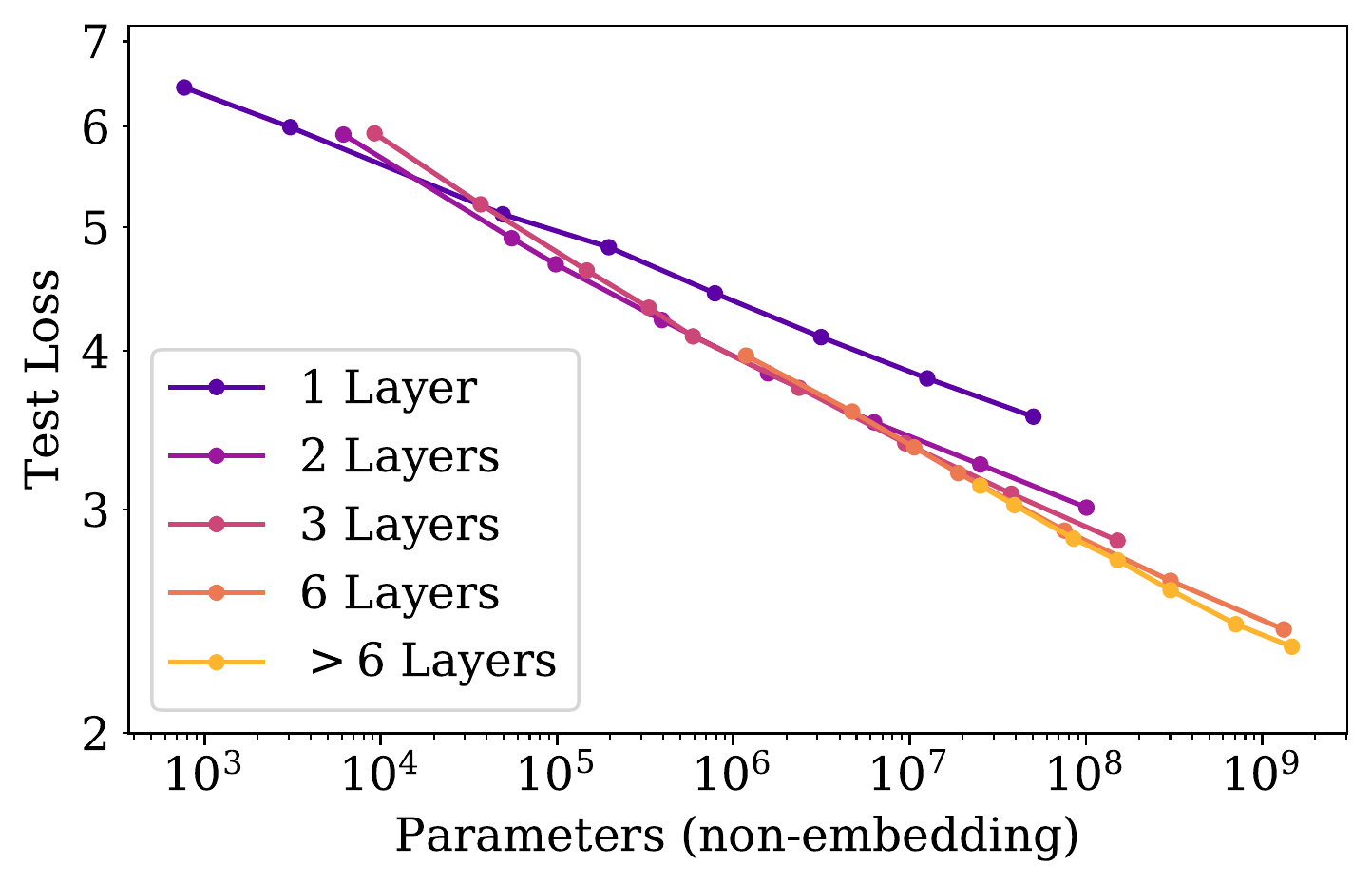}
\caption[Comparison of performance trend when including or excluding embeddings]{
{\bf Left:} When we include embedding parameters, performance appears to depend strongly on the number of layers in addition to the number of parameters.
{\bf Right:} When we exclude embedding parameters, the performance of models with different depths converge to a single trend. Only models with fewer than 2 layers or with extreme depth-to-width ratios deviate significantly from the trend.  \label{fig:PerformancevsModelSizeBody}}
\end{figure}

In Figure \ref{fig:PerformancevsModelSizeBody} we display the performance of a wide variety of models, ranging from small models with shape $(n_{\rm layer}, d_{{\rm model}}) = (2, 128)$ through billion-parameter models, ranging in shape from $(6, 4288)$ through $(207, 768)$.  Here we have trained to near convergence on the full WebText2 dataset and observe no overfitting (except possibly for the very largest models).

As shown in  Figure \ref{fig:BasicPowerLaws}, we find a steady trend with non-embedding parameter count $N$, which can be fit to the first term of Equation \eqref{eq:FundamentalLikelihioodvsModelandDataSize}, so that
\be
L(N) \approx \left( \frac{N_c}{N} \right)^{ \alpha_N }
\ee  
To observe these trends it is crucial to study performance as a function of $N$; if we instead use the total parameter count (including the embedding parameters) the trend is somewhat obscured (see Figure \ref{fig:PerformancevsModelSizeBody}).  This suggests that the embedding matrix can be made smaller without impacting performance, as has been seen in recent work \cite{lan2019albert}.

Although these models have been trained on the WebText2 dataset, their test loss on a variety of other datasets is also a power-law in $N$ with nearly identical power, as shown in Figure \ref{fig:GeneralizationVsModelSize}.

\subsubsection{Comparing to LSTMs and Universal Transformers}

\begin{figure}
\begin{centering}
\includegraphics[width=\columnwidth]{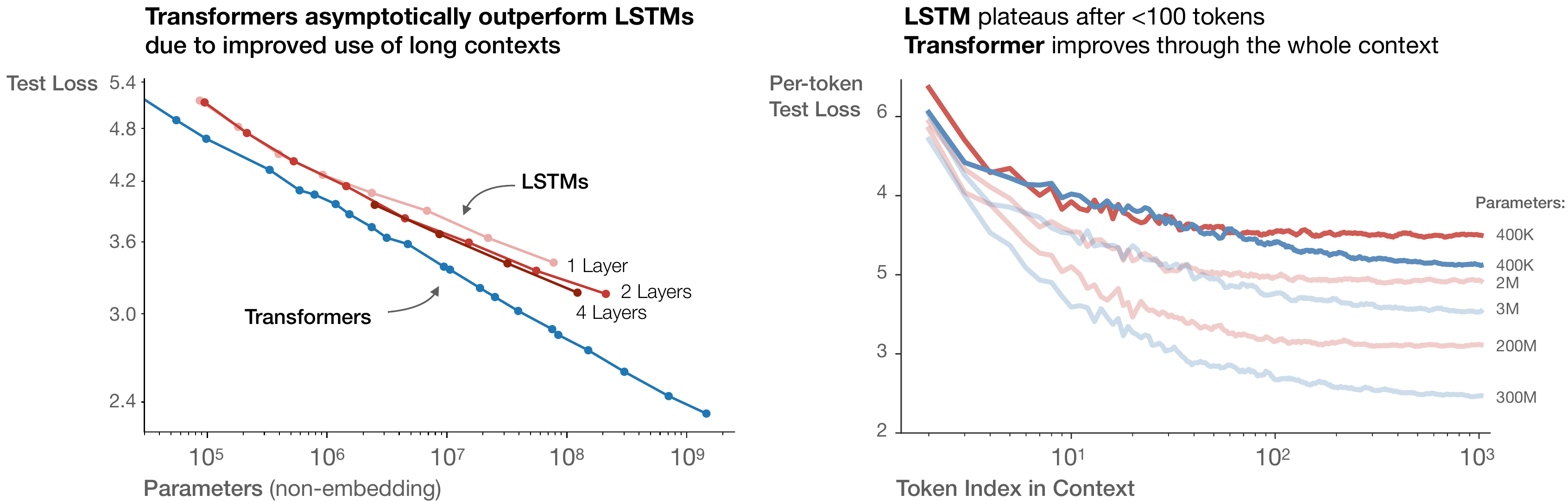}
\caption[LSTM and Transformer performance comparison]{
\label{fig:LSTMvsTransformers}}  
\end{centering}
\end{figure}

In Figure \ref{fig:LSTMvsTransformers} we compare LSTM and Transformer performance as a function of non-embedding parameter count $N$. The LSTMs were trained with the same dataset and context length.  We see from these figures that the LSTMs perform as well as Transformers for tokens appearing early in the context, but cannot match the Transformer performance for later tokens.  We present power-law relationships between performance and context position Appendix \ref{sec:ContextDependence}, where increasingly large powers for larger models suggest improved ability to quickly recognize patterns.

We also compare the performance of standard Transformers to recurrent Transformers \cite{DBLP:journals/corr/abs-1807-03819} in Figure \ref{fig:RecurrentTransformers} in the appendix.  These models re-use parameters, and so perform slightly better as a function of $N$, at the cost of additional compute per-parameter.

\subsubsection{Generalization Among Data Distributions}
\label{sec:GeneralizationtoOtherDistributions}

\begin{figure}
\noindent \centering{} \includegraphics[width=0.48\textwidth]{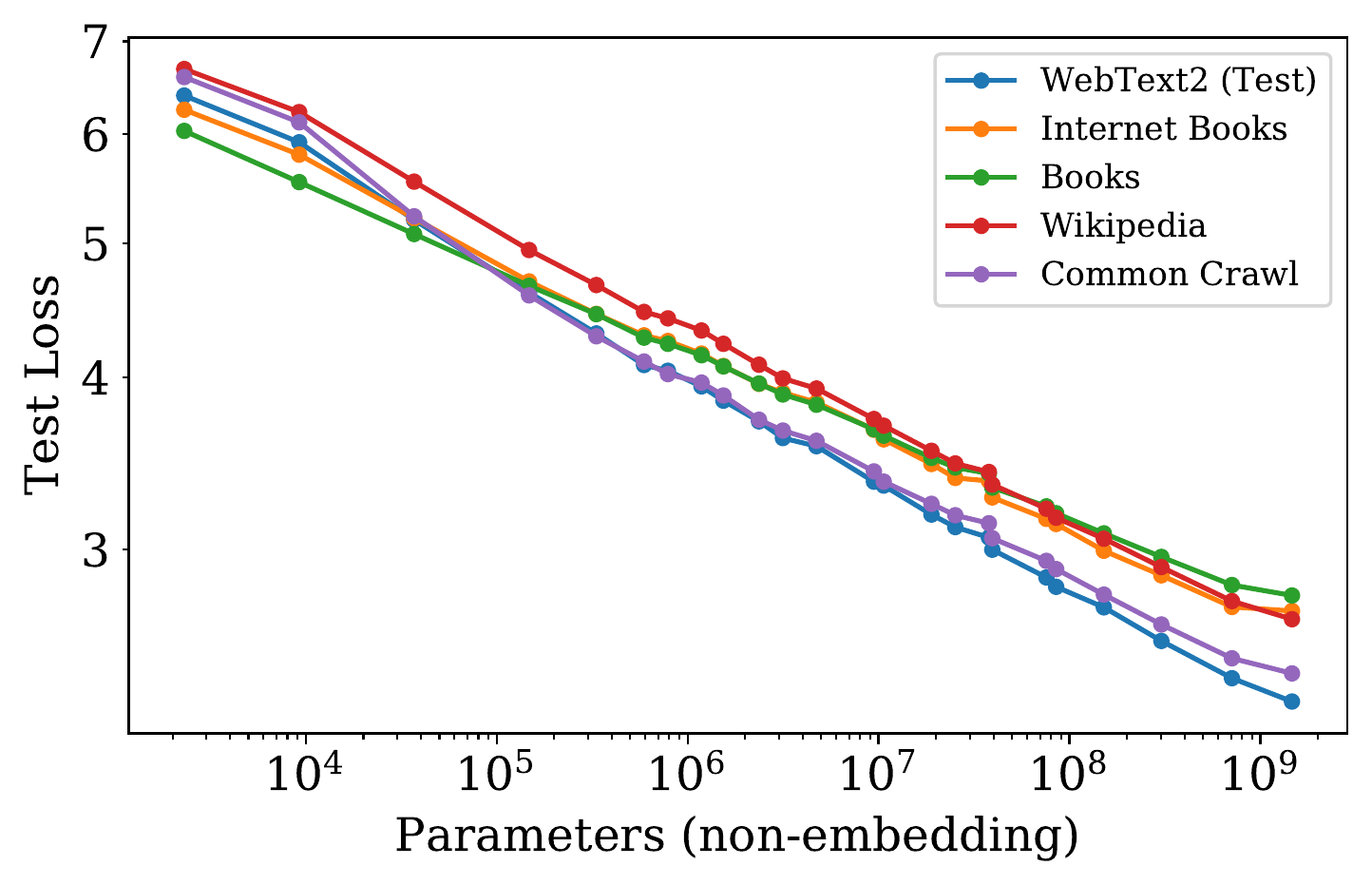}\hfill
\noindent \centering{} \includegraphics[width=0.48\textwidth]{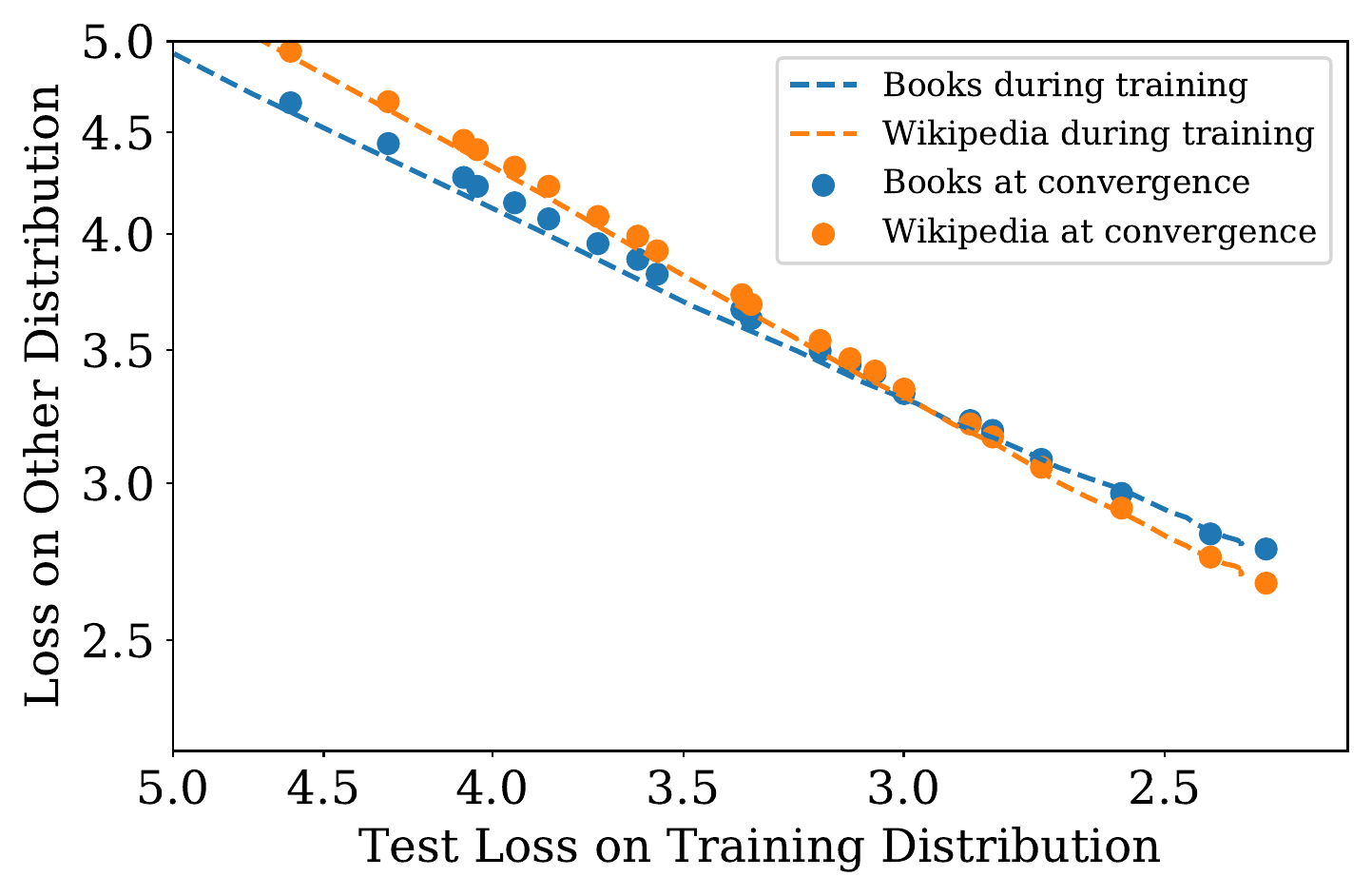}
\caption[Generalization to other test datasets]{
\textbf{Left:} Generalization performance to other data distributions improves smoothly with model size, with only a small and very slowly growing offset from the WebText2 training distribution.
\textbf{Right:} Generalization performance depends only on training distribution performance, and not on the phase of training.  We compare generalization of converged models (points) to that of a single large model (dashed curves) as it trains.
\label{fig:GeneralizationVsModelSize}}
\end{figure}

We have also tested our models on a set of additional text data distributions.  The test loss on these datasets as a function of model size is shown in Figure \ref{fig:GeneralizationVsModelSize}; in all cases the models were trained only on the WebText2 dataset.  We see that the loss on these other data distributions improves smoothly with model size, in direct parallel with the improvement on WebText2.  We find that generalization depends almost exclusively on the in-distribution validation loss, and does not depend on the duration of training or proximity to convergence. We also observe no dependence on model depth (see Appendix \ref{sec:DepthVsGeneralization}).

\subsection{Performance with Dataset Size and Compute}

We display empirical trends for the test loss as a function of dataset size $D$ (in tokens) and training compute $C$ in Figure \ref{fig:BasicPowerLaws}.    

For the trend with $D$ we trained a model with $(n_{\rm layer}, n_{\rm embd}) = (36, 1280)$ on fixed subsets of the WebText2 dataset.  We stopped training once the test loss ceased to decrease.  We see that the resulting test losses can be fit with simple power-law
\be
L(D) \approx \left( \frac{D_c}{D} \right)^{\alpha_D}
\ee
in the dataset size.  The data and fit appear in Figure \ref{fig:BasicPowerLaws}.

The total amount of non-embedding compute used during training can be estimated as $C = 6 N B S$, where $B$ is the batch size, $S$ is the number of parameter updates, and the factor of $6$ accounts for the forward and backward passes.  Thus for a given value of $C$ we can scan over all models with various $N$ to find the model with the best performance on step $S = \frac{C}{6 B S}$.  Note that in these results \emph{the batch size $B$ remains fixed for all models}, which means that these empirical results are not truly optimal.  We will account for this in later sections using an adjusted $C_{\rm min}$ to produce cleaner trends.

The result appears as the heavy black line on the left-hand plot in Figure \ref{fig:BasicPowerLaws}.  It can  be fit with 
\be
L(C) \approx \left( \frac{C_c}{C} \right)^{\alpha_C}
\ee
The figure also includes images of individual learning curves to clarify when individual models are optimal.  We will study the optimal allocation of compute more closely later on.
The data strongly suggests that sample efficiency improves with model size, and we also illustrate this directly in Figure \ref{fig:SampleEfficiency} in the appendix.

\section{Charting the Infinite Data Limit and Overfitting}
\label{sec:ChartingOverfitting}

In Section \ref{sec:Empirical} we found a number of basic  scaling laws for language modeling performance.  Here we will study the performance of a model of size $N$ trained on a dataset with $D$ tokens while varying $N$ and $D$ simultaneously.  We will empirically demonstrate that the optimally trained test loss accords with the scaling law of Equation \eqref{eq:FundamentalLikelihioodvsModelandDataSize}.  This  provides guidance on how much data we would need to train models of increasing size while keeping overfitting under control.

\subsection{Proposed $L(N,D)$ Equation}

\begin{figure}
\noindent \centering{} 
\includegraphics[width=0.48\textwidth]{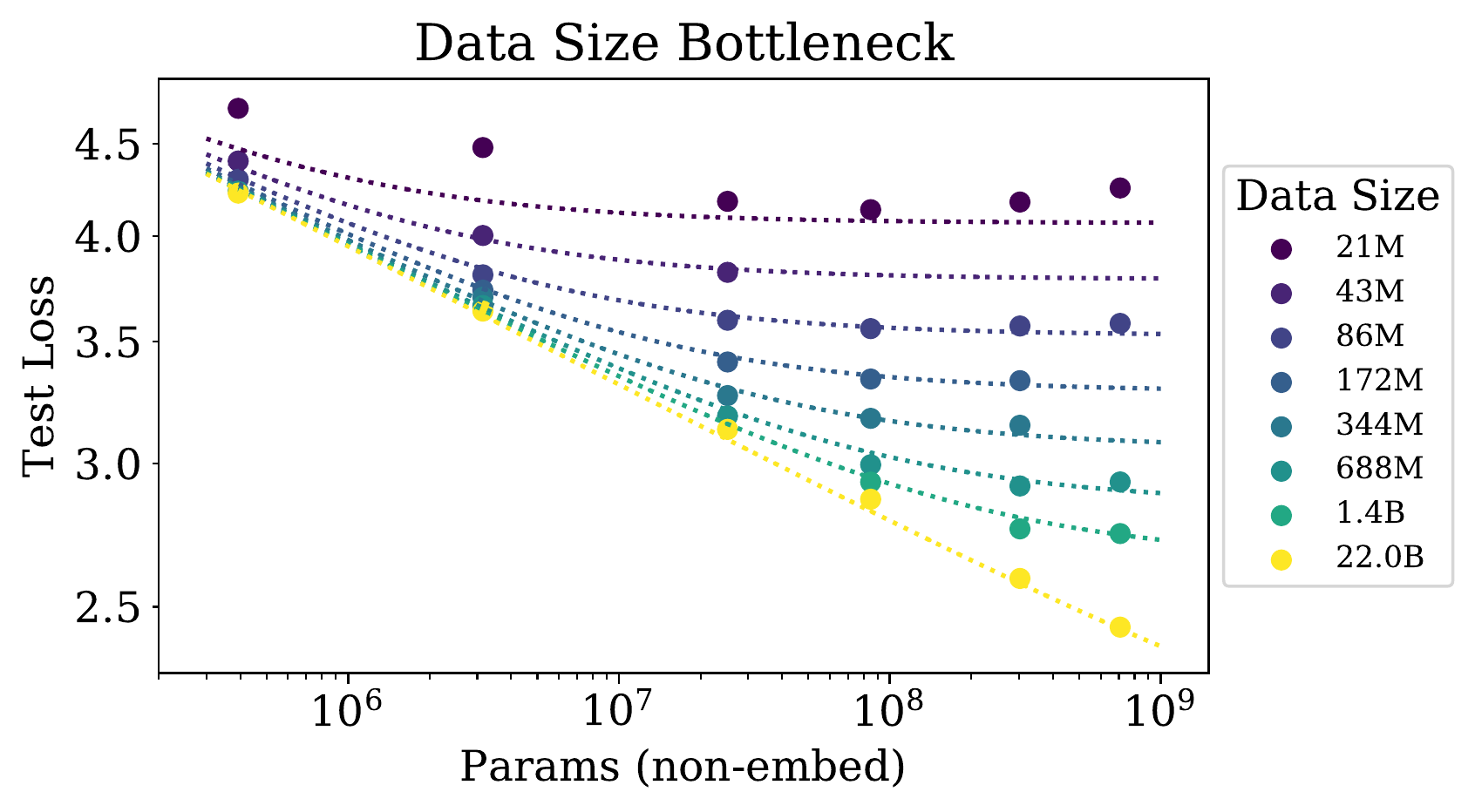}\hfill
\includegraphics[width=0.48\textwidth]{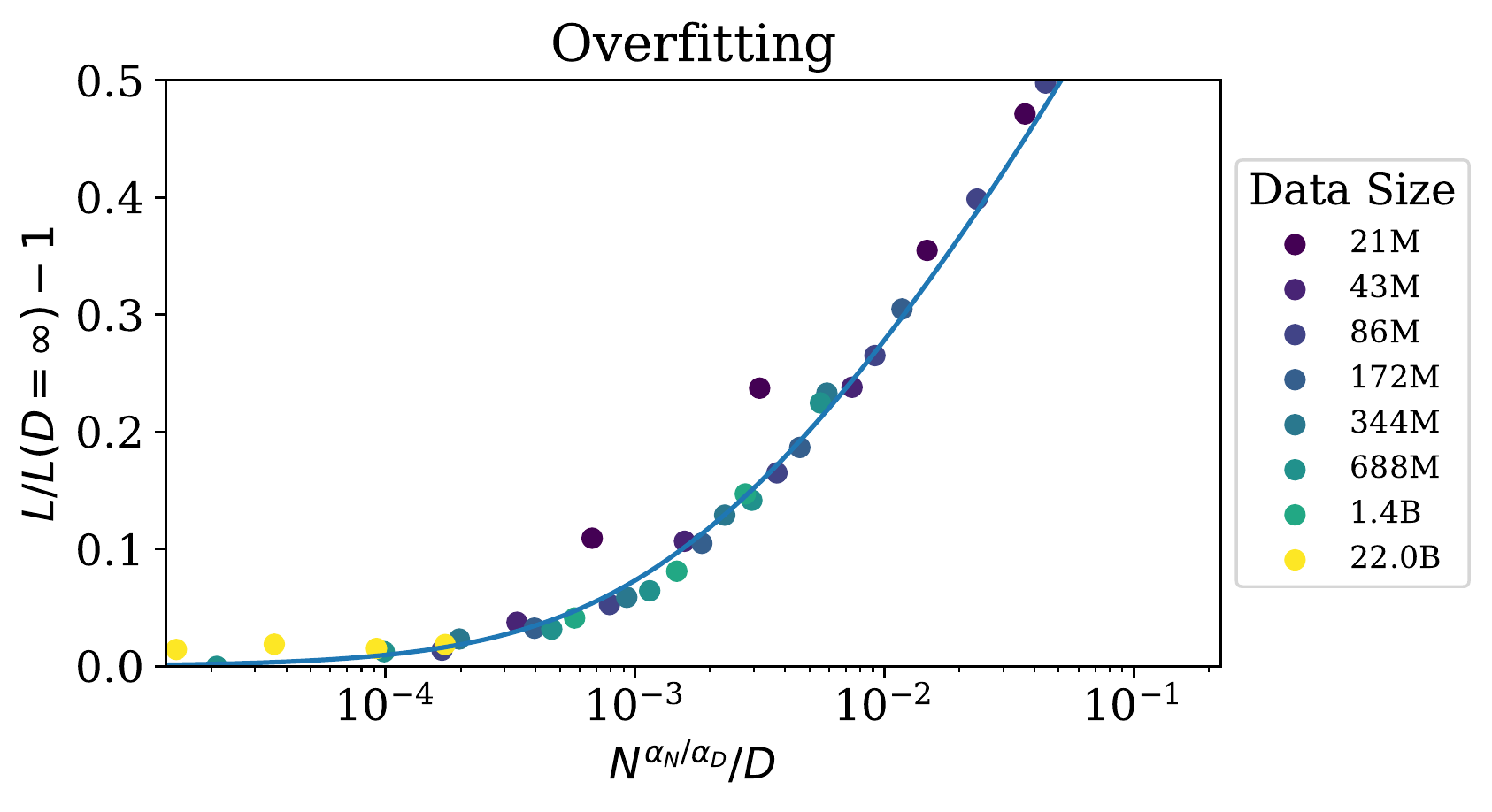}
\caption[Universality of overfitting]{
The early-stopped test loss $L(N, D)$ depends predictably on the dataset size $D$ and model size $N$ according to Equation \eqref{eq:FundamentalLikelihioodvsModelandDataSize}.
{\bf Left}:
For large $D$, performance is a straight power law in $N$. For a smaller fixed $D$, performance stops improving as $N$ increases and the model begins to overfit. (The reverse is also true, see Figure \ref{fig:LossvsModelDatasetSize}.)
{\bf Right}:  The extent of overfitting depends predominantly on the ratio $N^{\frac{\alpha_N}{\alpha_D}}/D$, as predicted in equation (\ref{eq:OverfittingPrediction}).  The line is our fit to that equation.
\label{fig:DatasetModelSizevsPerformance}}
\end{figure}

We have chosen the parameterization \eqref{eq:FundamentalLikelihioodvsModelandDataSize} (repeated here for convenience):
\be
L(N, D) 
= \left[ \left( \frac{N_c}{N} \right)^{\frac{\alpha_N}{\alpha_D}} + \frac{D_c}{D}  \right]^{\alpha_D}
\ee
using three principles:
\begin{enumerate}
\item Changes in vocabulary size or tokenization are expected to rescale the loss by an overall factor.  The parameterization of $L(N,D)$ (and all models of the loss) must naturally allow for such a rescaling.
\item Fixing $D$ and sending $N \to \infty$,  the overall loss should approach $L(D)$.  Conversely, fixing $N$ and sending $D \to \infty$ the loss must approach $L(N)$.
\item $L(N,D)$ should be analytic at $D=\infty$, so that it has a series expansion in $1/D$ with integer powers.  Theoretical support for this principle is significantly weaker than for the first two.
\end{enumerate}

Our choice of $L(N,D)$ satisfies the first requirement because we can rescale $N_c, D_c$ with changes in the vocabulary.  This also implies that the values of $N_c, D_c$ have no fundamental meaning.

Since we stop training early when the test loss ceases to improve and optimize all models in the same way, we expect that larger models should always perform better than smaller models.  But with fixed finite $D$, we also do not expect any model to be capable of approaching the best possible loss (ie the entropy of text).  Similarly, a model with fixed size will be capacity-limited.  These considerations motivate our second principle.
Note that knowledge of $L(N)$ at infinite $D$ and $L(D)$ at infinite $N$ fully determines all the parameters in $L(N,D)$.

The third principle is more speculative.  There is a simple and general reason one might expect overfitting to scale $\propto 1/D$ at very large $D$.  Overfitting should be related to the variance or the signal-to-noise ratio of the dataset \cite{1710.03667}, and this scales as $1/D$.  This expectation should hold for any smooth loss function, since we expect to be able to expand the loss about the $D \to \infty$ limit.   However, this argument assumes that $1/D$ corrections dominate over other sources of variance, such as the finite batch size and  other limits on the efficacy of optimization.  Without empirical confirmation, we would not be very confident of its applicability.

Our third principle explains the asymmetry between the roles of $N$ and $D$ in Equation \eqref{eq:FundamentalLikelihioodvsModelandDataSize}.  Very similar symmetric expressions\footnote{For example, one might have used $L(N,D) =  \left[ \left( \frac{N_c}{N} \right)^{\alpha_N} + \left( \frac{D_c}{D} \right)^{\alpha_D}  \right]^\beta$, but this does not  have a $1/D$ expansion.} are possible, but they would not have a $1/D$ expansion with integer powers, and would require the introduction of an additional parameter.

In any case, we will see that our equation for $L(N,D)$ fits the data  well, which is the most important justification for our $L(N,D)$ ansatz.

\subsection{Results}

We regularize all our models with 10\% dropout, and by tracking test loss and stopping once it is no longer decreasing.   The results are displayed in Figure \ref{fig:DatasetModelSizevsPerformance}, including a fit to the four parameters $\alpha_N, \alpha_D, N_c, D_c$ in Equation \eqref{eq:FundamentalLikelihioodvsModelandDataSize}:

\begin{table}[h!]
\centering
\vspace{-0.5em}
\begin{tabular}{|c| c | c | c | c|} 
 \hline
Parameter & $\alpha_N$ & $\alpha_D$ & $N_c$ & $D_c$ \\ [0.5ex] 
 \hline\hline
Value  & $0.076$ & $0.103$ & $6.4 \times 10^{13}$ & $1.8 \times 10^{13}$ \\ 
 \hline
\end{tabular}
\vspace{0.5em}
\caption{Fits to $L(N, D)$}
\vspace{-1em}
\end{table}

 We obtain an excellent fit, with the exception of the runs where the dataset has been reduced by a factor of $1024$, to about $2 \times 10^7$ tokens.  With such a small dataset, an epoch consists of only 40 parameter updates.  Perhaps such a tiny dataset represents a different regime for language modeling, as overfitting happens very early in training (see Figure \ref{fig:OverfittingandEarlyStopping}).  Also note that the parameters differ very slightly from those obtained in Section \ref{sec:Empirical}, as here we are fitting the full $L(N,D)$ rather than just $L(N, \infty)$ or $L(\infty, D)$.

To chart the borderlands of the infinite data limit, we can directly study the extent of overfitting.  For all but the largest models, we see no sign of overfitting when training with the full 22B token WebText2 dataset, so we can take it as representative of $D=\infty$.  Thus we can compare finite $D$ to the infinite data limit by defining
\be
\delta L(N, D) \equiv \frac{L(N, D)}{L(N, \infty)} - 1
\ee
and studying it as a function of $N, D$.   In fact, we see empirically that $\delta L$ depends only a specific combination of $N$ and $D$, as shown in Figure \ref{fig:OverfittingandEarlyStopping}.  This follows from the scaling law of Equation \eqref{eq:FundamentalLikelihioodvsModelandDataSize}, which implies
\be
\label{eq:OverfittingPrediction}
\delta L \approx  \left( 1 +  \left(\frac{N}{N_c} \right)^{\frac{\alpha_N}{\alpha_D}} \frac{D_c}{D} \right)^{\alpha_D} - 1 
\ee
Note that at large $D$ this formula also has a series expansion in powers of $1/D$.

We estimate that the variation in the loss with different random seeds is roughly $0.02$, which means that to avoid overfitting when training to within that threshold of convergence we require 
\be
D \gtrsim (5 \times 10^3) \, N^{0.74}
\ee
With this relation, models smaller than $10^9$ parameters can be trained with minimal overfitting on the 22B token WebText2 dataset, but our largest models will encounter some mild overfitting.  More generally, this relation shows that dataset size may grow sub-linearly in model size while avoiding overfitting.  Note however that this does not typically represent maximally compute-efficient training.  We should also emphasize that we have not optimized regularization (eg the dropout probability) while varying dataset and model size.

\section{Scaling Laws with Model Size and Training Time}
\label{sec:ScalingSizeandSteps}

In this section we will demonstrate that a simple scaling law provides a good description for the loss as a function of model size $N$ and training time.  First we will explain how to use the results of \cite{1812.06162}  to define a universal training step $S_{\rm min}$, which accounts for the fact that most of our models have not been trained at an optimal batch size.   Then we will demonstrate that we can fit the model size and training time dependence of the loss using Equation \eqref{eq:FundamentalLikelihioodvsModelandSteps}.  Later we will use these results to predict the optimal allocation of training compute between model size and training time, and then confirm that prediction.

\subsection{Adjustment for Training at $B_{\rm crit}(L)$}
\label{sec:OptimalBatchSize}

\begin{figure}
\noindent \centering{} 
\includegraphics[width=0.60\textwidth]{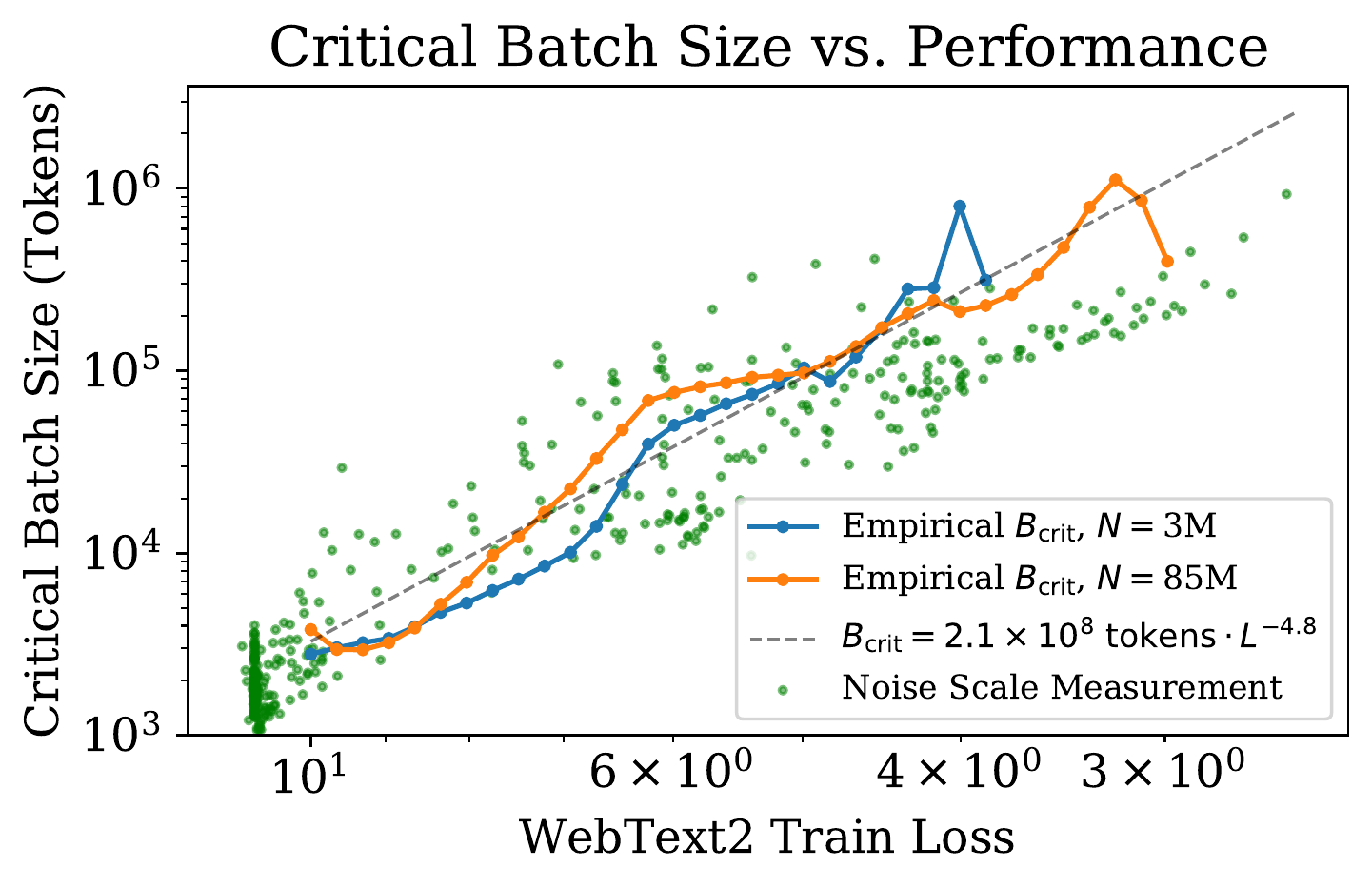}
\caption[Critical batch size]{The critical batch size $B_{\rm crit}$ follows a power law in the loss as performance increase, and does not depend directly on the model size.  We find that the critical batch size approximately doubles for every $13\%$ decrease in loss.  $B_{\rm crit}$  is measured empirically from the data shown in Figure \ref{fig:BatchPareto}, but it is also roughly predicted by the gradient noise scale, as in \cite{1812.06162}.  \label{fig:OptimalBatchSize}}
\end{figure}

A simple empirical theory for the batch size dependence of training was developed in \cite{1812.06162} (see also \cite{1811.03600, DBLP:journals/corr/abs-1907-04164}).  It was argued that there is a critical batch size $B_{\rm crit}$ for training; for $B$ up to $B_{\rm crit}$  the batch size can be increased with very minimal degradation in compute-efficiency, whereas for $B > B_{\rm crit}$ increases in $B$ result in diminishing returns.  It was also argued that the gradient noise scale provides a simple prediction for $B_{\rm crit}$, and that neither depends directly on model size except through the value of the loss that has been attained.  These results can be used to predict how training time and compute will vary with the batch size.  To utilize both training time and compute as effectively as possible, it is best to train with a batch size $B \approx B_{\rm crit}$.  Training at $B \gg B_{\rm crit}$  minimizes the number of training steps, while $B \ll B_{\rm crit}$ minimizes the use of compute.

More specifically, it was demonstrated that for a wide variety of neural network tasks, the number of training steps $S$ and the number of data examples processed $E = B S$ satisfy the simple relation
\be
\label{eq:TimeComputeTradeoff}
\left( \frac{S}{S_{\rm min}} -1 \right) \left( \frac{E}{E_{\rm min}} - 1 \right) = 1
\ee
when training to any fixed value of the loss $L$.  Here $S_{\rm min}$ is the minimum number of steps necessary to reach $L$, while $E_{\rm min}$ is the minimum number of data examples that must be processed. 

We demonstrate the relation \eqref{eq:TimeComputeTradeoff} for Transformers in Figure \ref{fig:BatchPareto} in the appendix. This relation defines the critical batch size 
\be
\label{eq:DefinitionBcrit}
B_{\rm crit}(L) \equiv \frac{E_{\rm min}}{S_{\rm min}}
\ee
which is a function of the target value of the loss.  Training at the critical batch size makes a roughly optimal time/compute tradeoff, requiring $2 S_{\rm min}$ training steps and processing $E = 2 E_{\rm min}$ data examples.  

In Figure \ref{fig:OptimalBatchSize} we have plotted the critical batch size and gradient noise scale\footnote{Although the critical batch size roughly matches the gradient noise scale, we are using a direct measurements of $B_{\rm crit}$ from Figures \ref{fig:BatchPareto} and \ref{fig:OptimalBatchSize} for all our later analyses.  } as a function of training loss for two different models.  We see that $B_{\rm crit}(L)$ is  independent of model size, and only depends on the loss $L$.  So the predictions of \cite{1812.06162} continue to hold for Transformer language models.  The critical batch size can be fit with a power-law in the loss
\be
B_{\rm crit}(L) \approx \frac{B_*}{L^{1/\alpha_B}}
\ee
where $B_* \approx 2 \times 10^8$ and $\alpha_B \approx 0.21$.  

We have chosen this parameterization for $B_{\rm crit}(L)$ because as the loss approaches its minimum value $L_{\rm min}$, the gradient noise scale is expected to diverge, and we expect $B_{\rm crit}$ to track this noise scale.  We do not know $L_{\rm min}$, as we see no sign that our models are approaching it, but $L_{\rm min} > 0$ since the entropy of natural language is non-zero.  Since apparently $L_{\rm min}$ is much smaller than the values of $L$ we have achieved, we used a parameterization where $B_{\rm crit}$ diverges as $L \to 0$. 

We will use $B_{\rm crit}(L)$ to estimate the relation between the number of training steps $S$ while training at batch size $B = 2^{19}$ tokens and the number of training steps while training at $B \gg B_{\rm crit}$.  This is simply
\be
\label{eq:AdjustedSteps}
S_{\rm min}(S) \equiv \frac{S}{1 + B_{\rm crit}(L)/B} \qquad (\text{minimum steps, at } B \gg B_{\rm crit})
\ee
for any given target value $L$ for the loss.  This also defines a critical value of the compute needed to train to $L$ with a model of size $N$ if we were to train at $B \ll B_{\rm crit}(L)$. This is
\be
\label{eq:AdjustedCompute}
C_{\rm min}(C) \equiv \frac{C  }{1 + B/B_{\rm crit}(L) } \qquad (\text{minimum compute, at } B \ll B_{\rm crit})
\ee
where $C = 6 N BS$ estimates the (non-embedding) compute used at batch size $B$.

\subsection{Results for $L(N, S_{\rm min})$ and Performance with Model Size and Compute}

\begin{figure}
\noindent \centering{}
\includegraphics[width=0.47\textwidth]{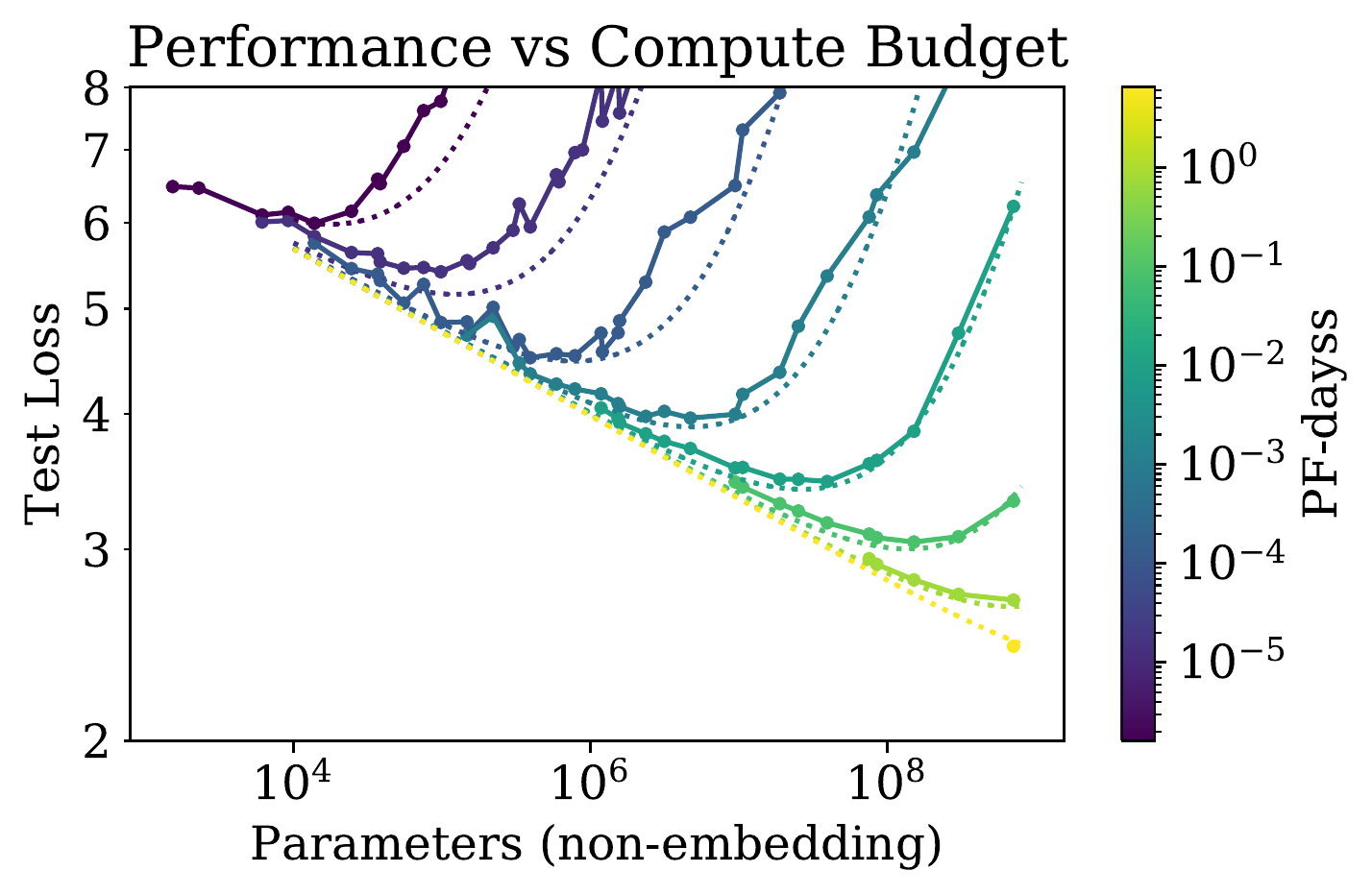}\hfill
\includegraphics[width=0.47\textwidth]{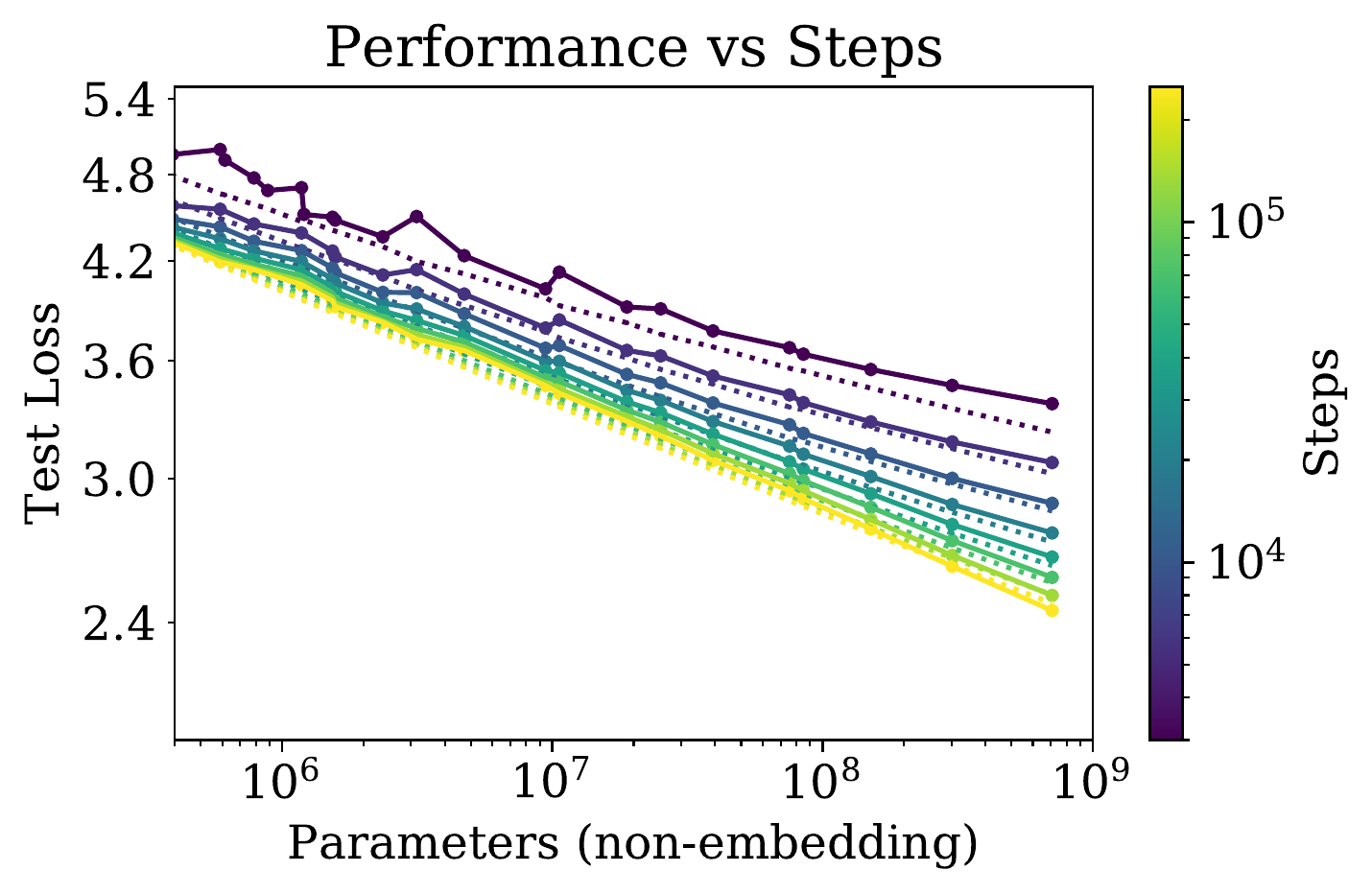} 
\caption[Performance versus compute budget or number of parameter updates]{
When we hold either total compute or number of training steps fixed, performance follows $L(N,S)$ from Equation \eqref{eq:FundamentalLikelihioodvsModelandSteps2}.  Each value of compute budget has an associated optimal model size that maximizes performance.  Mediocre fits at small $S$ are unsurprising, as the power-law equation for the learning curves breaks down very early in training. \label{fig:ComputevsParamsvsPerformance}}
\end{figure}

Now we will use $S_{\rm min}$ defined in Equation \eqref{eq:AdjustedSteps} to obtain a  simple and universal fit for the dependence of the loss on model size and training time in the infinite data limit.  We will fit the stable, Adam-optimized training runs using Equation \eqref{eq:FundamentalLikelihioodvsModelandSteps}, repeated here for convenience:
\be
\label{eq:FundamentalLikelihioodvsModelandSteps2}
L(N, S_{\rm min}) = \left( \frac{N_c}{N} \right)^{\alpha_N}  + \left( \frac{S_c }{S_{\rm min}} \right)^{\alpha_S}
\ee
for the loss.  We include all training steps after the warmup period of the learning rate schedule, and find  a fit to the data with the parameters:

\begin{table}[h!]
\centering
\begin{tabular}{|c| c | c | c | c| } 
 \hline
Parameter & $\alpha_N$ & $\alpha_S$ & $N_c$ & $S_c$   \\ [0.5ex] 
 \hline\hline
Value  & $0.077$ & $0.76$ & $6.5 \times 10^{13}$ & $2.1 \times 10^3$  \\ 
 \hline
\end{tabular}
\vspace{0.5em}
\caption{Fits to $L(N, S)$}
\vspace{-1em}
\end{table}

With these parameters, we obtain the learning curve fits in Figure \ref{fig:LearningCurveFitsandResiduals}.  Though the fits are imperfect, we believe they are quite compelling given the simplicity of Equation \eqref{eq:FundamentalLikelihioodvsModelandSteps2}.  

The data and fits can be visualized in a different and more interesting way, as shown in Figure \ref{fig:ComputevsParamsvsPerformance}.  There we study the test loss as a function of model size while fixing either the total non-embedding compute $C$ used in training, or the number of steps $S$.  For the fits we use Equation \eqref{eq:AdjustedCompute} and \eqref{eq:AdjustedSteps} along with the parameters above and Equation \eqref{eq:FundamentalLikelihioodvsModelandSteps2}.

The power-law dependence of the loss on $S_{\rm min}$ reflects the interplay of optimizer dynamics and the loss landscape.    Since the fits are best late in training, when the loss may be approximately quadratic, the power-law should provide information about the spectrum of the Hessian of the loss.  Its universality suggests that the Hessian eigenvalue density is roughly independent of model size.  

\subsection{Lower Bound on Early Stopping Step}
\label{sec:EarlyStop}

The results for $L(N,S_{\rm min})$ can be used to derive a lower-bound (and rough estimate) of the step at which early stopping should occur when training is data limited.  
It is motivated by the idea that finite and infinite $D$ learning curves for a given model will be very similar until we reach $S_{\rm min} \approx S_{\rm stop}$. Thus overfitting should be proportional to the correction from simply ending  training at $S_{\rm stop}$.  This will underestimate $S_{\rm stop}$, because in reality the test loss will decrease more slowly when we have a finite $D$, and therefore we will require more training steps to reach the optimal test loss at finite $D$.  
This line of reasoning leads to the inequality
\be
\label{eq:EarlyStopInequality}
S_{\rm stop}(N,D)  \gtrsim \frac{S_c}{\left[ L(N,D) - L(N, \infty) \right]^{1 / \alpha_S}}
\ee
where $L(N, \infty)$ is the converged loss, evaluated with infinite available data.  This inequality and its comparison to the empirical data is displayed in Figure \ref{fig:OverfittingandEarlyStopping} in the appendix.  In that figure, the values of $S_{\rm stop}$ and $L(N,D)$ are empirical (though $S_{\rm stop}$ is adjusted to mimic training at $B \gg B_{\rm crit}$), while $L(N, \infty)$ is computed from the fit to $L(N,D)$ evaluated at $D=\infty$.

\begin{figure}
\noindent \centering{}
\includegraphics[width=0.98\textwidth]{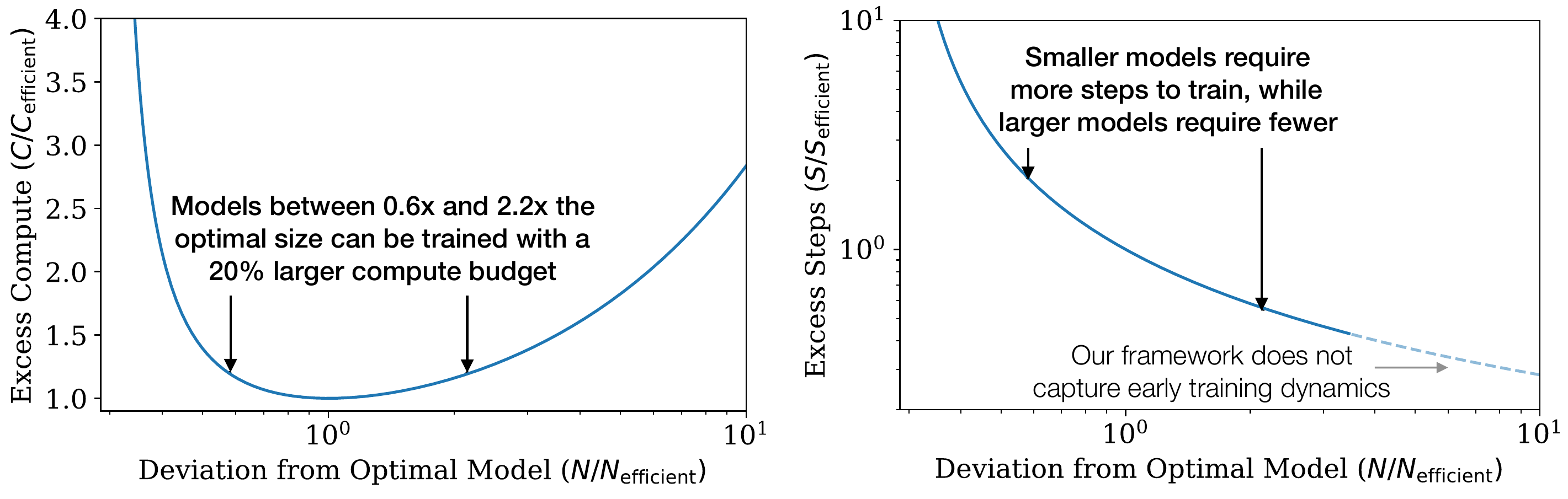}\
\caption[Training on suboptimal models]{\textbf{Left:} Given a fixed compute budget, a particular model size is optimal, though somewhat larger or smaller models can be trained with minimal additional compute. \textbf{Right:} Models larger than the compute-efficient size require fewer steps to train, allowing for potentially faster training if sufficient additional parallelism is possible. Note that this equation should not be trusted for very large models, as it is only valid in the power-law region of the learning curve, after initial transient effects.  \label{fig:SubOptimalModels}}
\end{figure}

\section{Optimal Allocation of the Compute Budget}
\label{sec:OptimalCompute}

We displayed the \emph{empirical} trend of performance as a function of the computation used during training in the top-right of Figure \ref{fig:BasicPowerLaws}.  However, this result involved training at a fixed batch size $B$, whereas we know that in fact we could train more efficiently\footnote{One might ask why we did not simply train at $B_{\rm crit}$ in the first place.  The reason is that it depends not only on the model but also on the target value of the loss we wish to achieve, and so is a moving target.} by training at the batch size $B_{\rm crit}$ discussed in Section \ref{sec:OptimalBatchSize}. Large and small values of the loss could have been achieved with fewer samples or fewer steps, respectively, and correcting for this inefficiency by standardizing to the critical batch size results in cleaner and more predictable trends.

\begin{figure}
\centering{}
\includegraphics[width=0.5\textwidth]{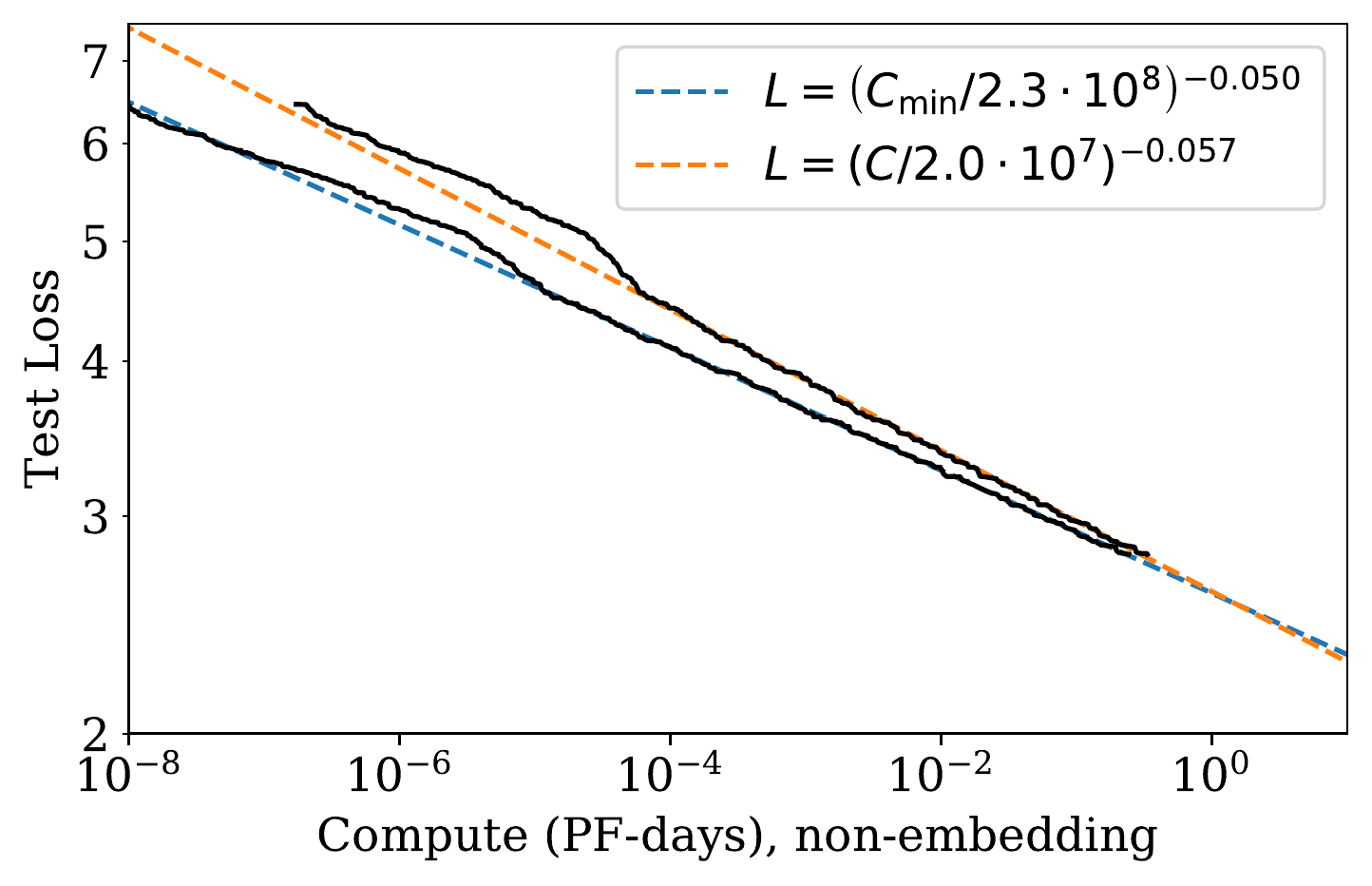}  
\caption[Comparison between empirical and adjusted compute trends]{
When adjusting performance to simulate training far below the critical batch size, we find a somewhat altered power law for $L(C_{\rm min})$ when compared with the fully empirical results.  The conspicuous lump at $10^{-5}$ PF-days marks the transition from 1-layer to 2-layer networks; we exclude 1-layer networks in the power-law fits.  It is the $L(C_{\rm min})$ trend that we expect to provide a reliable extrapolation for larger compute.
\label{fig:ComputeEfficientAdjusted}}
\end{figure}

In this section we will adjust for this oversight. More importantly, we will use the results of Section \ref{sec:ScalingSizeandSteps} to determine the optimal \emph{allocation} of compute between model size $N$ and the quantity of data processed during training, namely $2 B_{\rm crit}  S_{\rm min}$.  We will determine this allocation both empirically and theoretically, by using the equation for $L(N, S_{\rm min})$, and we will demonstrate that these methods agree.

\subsection{Optimal Performance and Allocations}

Let us first study the loss as a function of the optimally allocated compute from Equation \eqref{eq:AdjustedCompute}.  The result is plotted in Figure \ref{fig:ComputeEfficientAdjusted}, along with a power-law fit.  We see that as compared to the compute plot of Figure \ref{fig:BasicPowerLaws}, the new fit with $C_{\rm min}$ is somewhat improved.  

\begin{figure}
\noindent \centering{}
\includegraphics[width=0.48\textwidth]{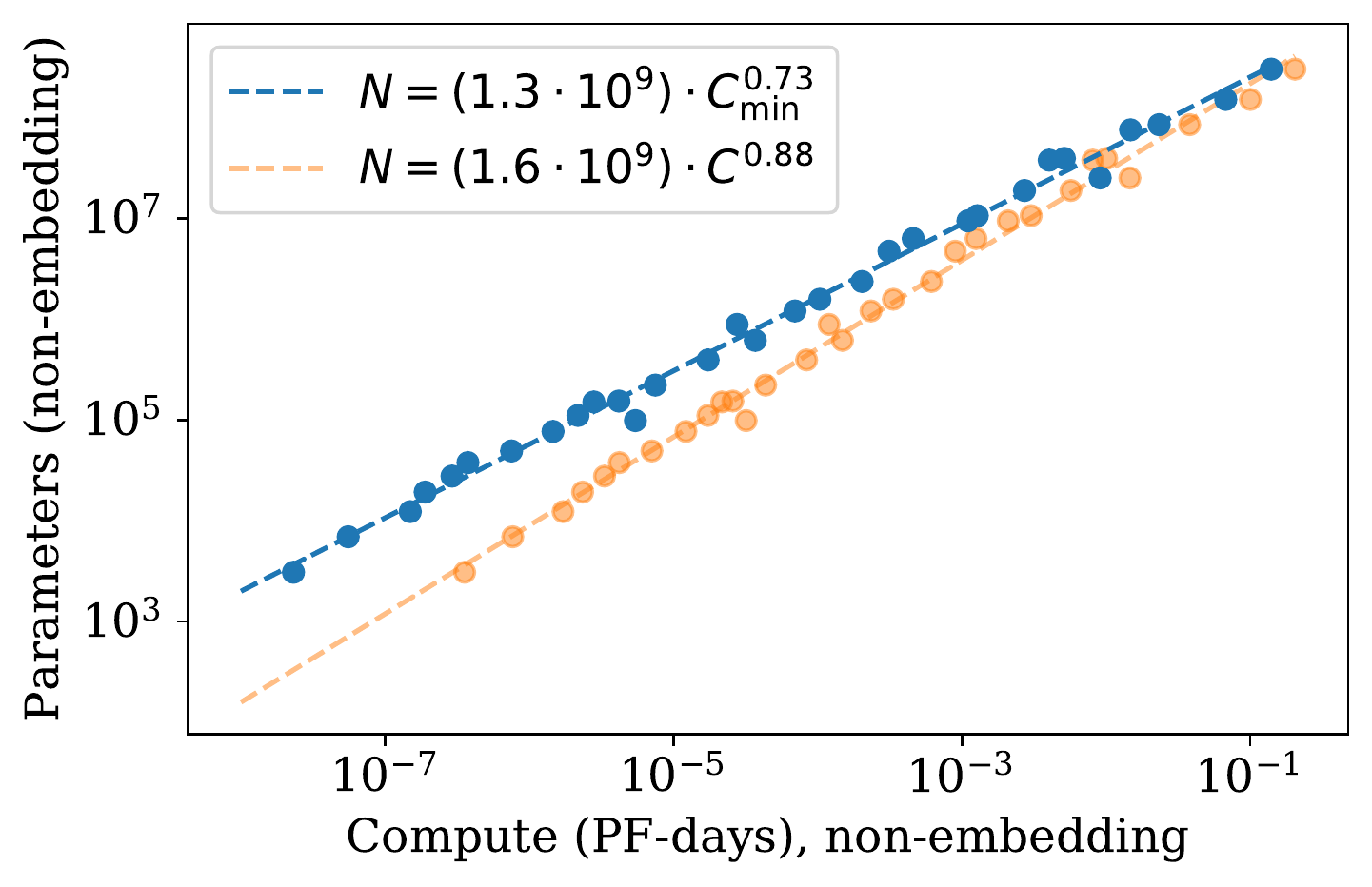}\hfill
\includegraphics[width=0.48\textwidth]{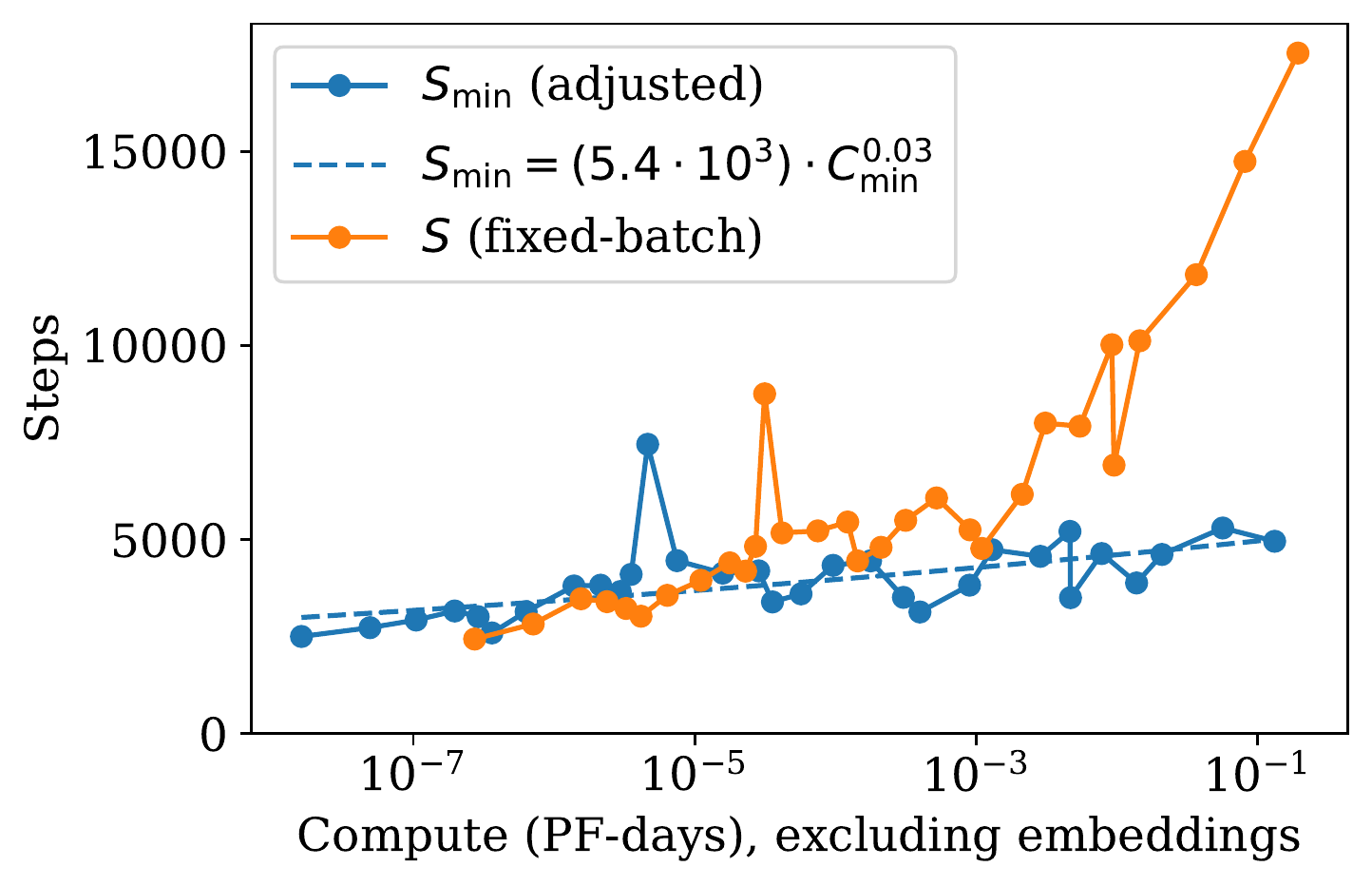}
\caption[Optimal model size and serial number of steps versus compute budget]{
\textbf{Left:} Each value of the compute budget $C_{\rm min}$ has an associated optimal model size $N$.  Optimal model size grows very rapidly with $C_{\rm min}$, increasing by 5x for each 10x increase in compute.  The number of data examples processed makes up the remainder of the increase, growing relatively modestly by only 2x.
\textbf{Right:} The batch-adjusted number of optimization steps also grows very slowly, if at all, meaning that most of the growth in data examples processed can be used for increased batch sizes.
\label{fig:ComputevsPerformance}}
\end{figure}

Given $L(C_{\rm min})$, it is natural to ask for the optimal model size $N(C_{\rm min})$ that provides the minimal loss with a given quantity of training compute.  The optimal model size is shown in Figure \ref{fig:ComputevsPerformance}.  We observe that $N(C_{\rm min})$ can be fit very well with a power-law
\be
N(C_{\rm min}) \propto (C_{\rm min})^{0.73}.
\ee
In Figure \ref{fig:SubOptimalModels}, we show the effect of training models of sub-optimal sizes (see Appendix \ref{sec:suboptimal-models}).

By definition $C_{\rm min} \equiv 6 N B_{\rm crit} S$, and so we can use $N(C_{\rm min})$ to extract further results.  In particular, since prior fits show $B \propto L^{-4.8}$ and $L \propto C_{\rm min}^{-0.05}$, we can conclude that $B_{\rm crit} \propto C_{\rm min}^{0.24}$.  This leads us to conclude that the optimal number of steps will only grow very slowly with compute, as
\be
S_{\rm min} \propto (C_{\rm min})^{0.03},
\ee
matching the empirical results in Figure \ref{fig:ComputevsPerformance}.  In fact the measured exponent is sufficiently small that our results may even be consistent with an exponent of zero.  

Thus we conclude that as we scale up language modeling with an optimal allocation of computation, we should predominantly increase the model size $N$, while simultaneously scaling up the batch size  via $B \propto B_{\rm crit}$ with negligible increase in the number of serial steps.  Since compute-efficient training uses relatively few optimization steps, additional work on speeding up early training dynamics may be warranted.

\subsection{Predictions from $L(N, S_{\rm min})$}

The results for $L(C_{\rm min})$ and the allocations can be predicted  from the $L(N, S_{\rm min})$ equation obtained in Section \ref{sec:ScalingSizeandSteps}.  Given our  equation for $L(N, S_{\rm min})$, we can substitute $S_{\rm min} = \frac{C_{\rm min}}{6 N B}$ and then find the minimum of the loss as a function of $N$, while fixing the training compute.  
We carry out this procedure in detail  in Appendix \ref{app:ComputeEfficientTraining}, where we also provide some additional predictions.  

For the loss as a function of training compute, we predict that
\be
L(C_{\rm min}) = \left( \frac{C_c^{\rm min}}{C_{\rm min}} \right)^{\alpha_C^{\rm min}}
\ee
where
\be
\alpha_C^{\rm min} \equiv \frac{1}{1/\alpha_S + 1/\alpha_B + 1/\alpha_N} \approx 0.054
\ee
in excellent agreement with the exponent of Figure \ref{fig:ComputeEfficientAdjusted}.  We also predict that
\be
N(C_{\rm min}) \propto (C_{\rm min})^{\alpha_C^{\rm min} / \alpha_N} \approx  (C_{\rm min})^{0.71}
\ee
which also matches the scaling of Figure \ref{fig:ComputevsPerformance} to within a few percent.  Our scaling laws provide a predictive framework for the performance of language modeling.

\subsection{Contradictions and a Conjecture}
\begin{figure}
\noindent \centering{}
\includegraphics[width=0.8\textwidth]{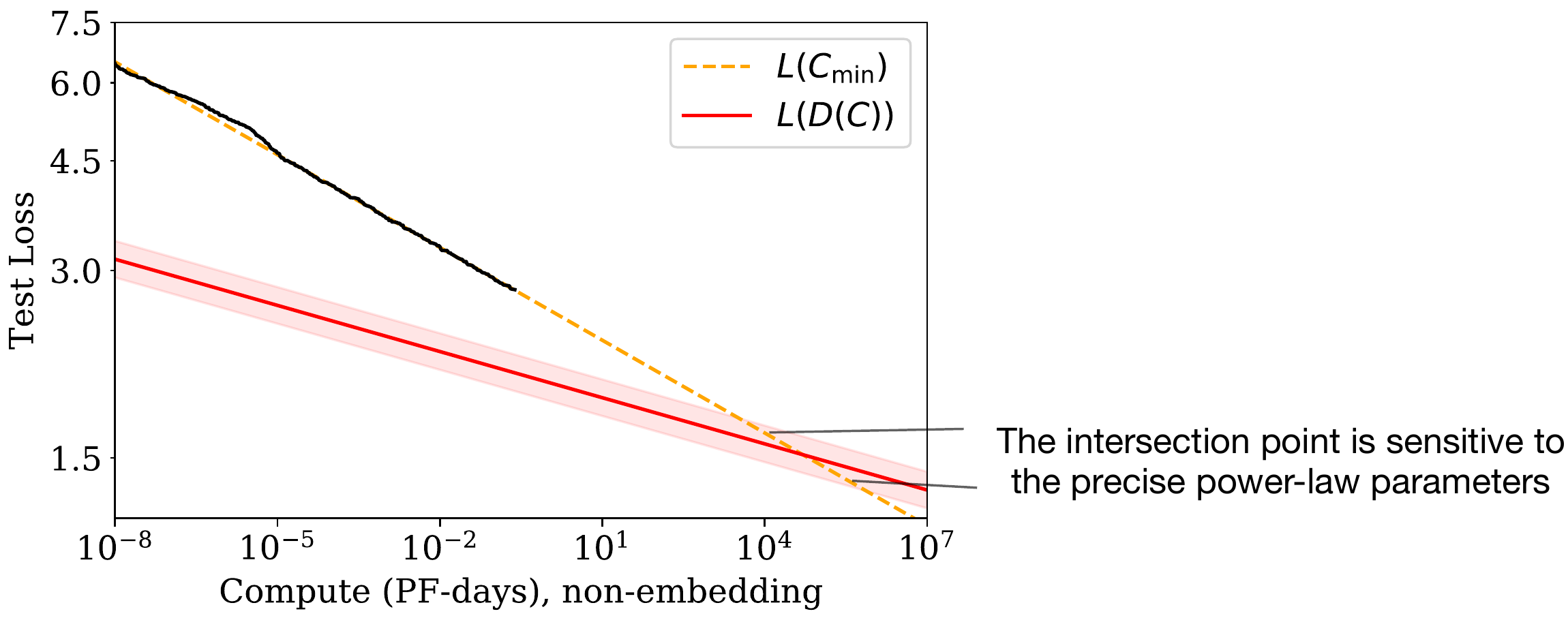}
\caption[Contradiction between compute and data trends]{Far beyond the model sizes we study empirically, we find a contradiction between our equations for $L(C_{\rm min})$ and $L(D)$ due to the slow growth of data needed for compute-efficient training.  The intersection marks the point before which we expect our predictions to break down.  The location of this point is highly sensitive to the precise exponents from our power-law fits. \label{fig:Contradiction}}
\end{figure}

We observe no signs of deviation from straight power-law trends at large values of compute, data, or model size. Our trends must eventually level off, though, since natural language has non-zero entropy. 

Indeed, the trends for compute-efficient training described in this section already contain an apparent contradiction. At scales several orders of magnitude above those documented here, the performance predicted by the $L(C_{\rm min})$ scaling law decreases below what should be possible given the slow growth in training data with compute.  This implies that our scaling laws must break down before this point, but we conjecture that the intersection point has a deeper meaning: it provides an estimate of the point at which Transformer language models reach maximal performance.

Since the amount of data used by compute-efficient training grows  slowly with the compute budget, the performance predicted by $L(C_{\rm min})$ eventually hits a lower bound set by the $L(D)$ power law (see Figure \ref{fig:Contradiction}).  Let us work this out in more detail.

To keep overfitting under control, the results of Section \ref{sec:ChartingOverfitting} imply that we should scale the dataset size as
\be
\label{eq:DataGrowthOverfitting}
D \propto N^{0.74} \propto C_{\rm min}^{0.54}
\ee
where we have used the compute-efficient $N(C_{\rm min})$ from Figure \ref{fig:ComputevsPerformance}.

Let us compare this to the data requirements of compute-efficient training.  If we train at the critical batch size (i.e. $C=2C_{\rm min}$) and never re-use data during training, we find that data usage grows with compute as
\be
D (C_{\rm min}) = \frac{2C_{\rm min}}{6 N(C_{\rm min})} \approx \left( 4 \times 10^{10}  \ {\rm tokens} \right) (C_{\rm min} / \mathrm{PF}{\text -}\mathrm{Day}  )^{0.26} 
\ee
This is the maximum rate at which the dataset size can productively grow with compute, since it means that we are only training for a single epoch.  But it grows the dataset much more slowly than in Equation \eqref{eq:DataGrowthOverfitting}.  It appears to imply that compute-efficient training will eventually run into a problem with overfitting, even if the training process never re-uses any data!

According to Figure \ref{fig:BasicPowerLaws}, we expect that when we are bottlenecked by the dataset size (ie by overfitting), the loss should scale as $L(D) \propto D^{-0.095}$.  This implies that the loss would scale with compute as $L(D(C_{\rm min})) \propto C_{\rm min}^{-0.03}$ once we are data-limited.  Once again, we have a contradiction, as this will eventually intersect with our prediction for $L(C_{\rm min})$ from Figure \ref{fig:ComputeEfficientAdjusted}, where we found a scaling $L(C_{\rm min}) \propto C_{\rm min}^{-0.050}$.  

The intersection point of $L(D(C_{\rm min}))$ and $L(C_{\rm min})$ occurs at
\be
C^* \sim 10^4~\mathrm{PF}{\text -}\mathrm{Days}
\quad
N^* \sim 10^{12}~\text{parameters},
\quad
D^* \sim 10^{12}~\text{tokens},
\quad
L^* \sim 1.7~\text{nats/token}
\label{eq:SpecialPoint}
\ee
though the numerical values are highly uncertain, varying by an order or magnitude in either direction depending on the precise values of the exponents from the power-law fits.  The most obvious interpretation is that our scaling laws break down at or before we reach this point, which is still many orders of magnitude away in both compute and model size.  

One might also conjecture that this intersection point has a deeper meaning.  If we cannot increase the model size beyond $N^*$ without qualitatively different data requirements, perhaps this means that once we reach $C_{\rm min}^*$ and $N^*$, we have extracted all of the reliable information available in natural language data.  In this interpretation, $L^*$ would provide a rough estimate for the entropy-per-token\footnote{Defining words using the \texttt{wc} utility, the WebText2 dataset has $1.4$ tokens per word and $4.3$ characters per token. } of natural language.  In this scenario, we would expect the loss trend to level off at or before $L^*$.

We can guess at the functional form of $L(C_{\rm min})$ as it levels off by considering a version of our training dataset with added noise.  For example, we could append a random string of tokens to each context shown to the model to artificially boost the loss by a constant additive factor. Then, the distance from the noise floor $L-L_{\rm noise}$ would be a more meaningful performance metric, with even a small decrease in this distance potentially representing a significant boost in qualitative performance. Since the artificial noise would affect all of our trends equally, the critical point of \ref{eq:SpecialPoint} would not change (aside from the absolute value of $L^*$), and may be meaningful even if it occurs after the leveling off.

\section{Related Work}

Power laws can arise from a wide variety of sources \cite{thurner2018introduction}.  Power-law scalings with model and dataset size in density estimation \cite{wasserman2006all} and  in random forest models \cite{biau2012analysis} may be connected with our results. These models suggest that power-law exponents may have a very rough interpretation as the inverse of the number of relevant features in the data.

Some early \cite{banko2001scaling, DBLP:journals/corr/cs-CL-0108005}   work found power-law scalings between performance and dataset size.  More recent work \cite{1712.00409, Hestness:2019:BHA:3293883.3295710} also investigated scaling between model size and data size; their work is perhaps the closest to ours in the literature\footnote{After this work was completed, \cite{rosenfeld2019constructive} also appeared, which makes similar predictions for the dependence of loss on both model and dataset size.}.  Note, however, that \cite{1712.00409} found super-linear scaling of dataset size with model size, whereas  we find a sub-linear scaling.  There are some parallels between our findings on optimal allocation of compute and \cite{1906.06669}, including power-law learning curves.  EfficientNets \cite{DBLP:journals/corr/abs-1905-11946} also appear to obey an approximate power-law relation between accuracy and model size.  Very recent work \cite{1909.12673} studies scaling with both dataset size and model size for a variety of datasets, and fits an ansatz similar to ours.

EfficientNet \cite{DBLP:journals/corr/abs-1905-11946}   advocates scaling depth and width exponentially (with different coefficients) for optimal performance of image models, resulting in a power-law scaling of width as a function of depth.    We find that for language models this power should be roughly one when scaling up (as width/depth should remain fixed). But more importantly, we find that the precise architectural hyperparameters are unimportant compared to the overall scale of the language model.  In \cite{ResNetsEnsemblesShallow} it was argued that deep models can function as ensembles of shallower models, which could potentially explain this finding.  Earlier work  \cite{Zagoruyko_2016} has compared width and depth, and found that wide ResNets can outperform deep ResNets on image classification.  
Some studies fix computation per data example, which tends to scale in proportion to the number of model parameters, whereas we investigate scaling with both model size and the quantity of training computation. 

Various works \cite{1710.03667,1812.11118} have investigated generalization in highly overparameterized models, finding a ``jamming transition'' \cite{1901.01608} when the model size reaches the dataset size (this may  require training many orders of magnitude beyond typical practice, and in particular does not use early stopping).  We do not observe such a transition, and find that the necessary training data scales sublinearly in the model size.
Expansions in the model size, particularly at large width \cite{jacot2018neural, 1902.06720}, may provide a useful framework for thinking about some of our scaling relations.  Our results on optimization, such as the shape of learning curves, can likely be explained using a noisy quadratic model, which can provide quite accurate predictions \cite{DBLP:journals/corr/abs-1907-04164} in realistic settings.  Making this connection quantitative will require a characterization of the Hessian spectrum \cite{DBLP:journals/corr/abs-1811-07062, DBLP:journals/corr/abs-1901-10159, unpublished-grd}.

\section{Discussion}

We have observed consistent scalings of language model log-likelihood loss with non-embedding parameter count $N$, dataset size $D$, and optimized training computation $C_{\rm min}$, as encapsulated in Equations \eqref{eq:FundamentalLikelihioodvsModelandDataSize} and \eqref{eq:FundamentalLikelihioodvsModelandSteps}.  Conversely, we find very weak dependence on many architectural and optimization hyperparameters.  Since scalings with $N, D, C_{\rm min}$ are power-laws, there are diminishing returns with increasing scale.

We were able to precisely model the dependence of the loss on $N$ and $D$, and alternatively on $N$ and $S$, when these parameters are varied simultaneously.  We used these relations to derive the compute scaling, magnitude of overfitting, early stopping step, and data requirements when training large language models. So our scaling relations go beyond mere observation to provide a predictive framework.  One might interpret these relations as analogues of the ideal gas law, which relates the macroscopic properties of a gas in a universal way, independent of most of the details of its microscopic consituents.

It is natural to conjecture that the scaling relations will apply to other generative modeling tasks with a maximum likelihood loss, and perhaps in other settings as well.    To this purpose, it will be  interesting to test these relations on other domains, such as images, audio, and video models, and perhaps also for random network distillation.  At this point we do not know which of our results depend on the structure of natural language data, and which are universal.  It would also be exciting to find a theoretical framework from which the scaling relations can be derived: a `statistical mechanics' underlying the `thermodynamics' we have observed.  Such a theory might make it possible to derive other more precise predictions, and provide a systematic understanding of the limitations of the scaling laws.

In the domain of natural language, it will be important to investigate whether continued improvement on the loss translates into improvement on relevant language tasks.  Smooth quantitative change can mask major qualitative improvements: ``more is different''.  For example, the smooth aggregate growth of the economy provides no indication of the specific technological developments that underwrite it. Similarly, the smooth improvements in language model loss may hide seemingly qualitative changes in capability.

Our results strongly suggest that larger models will continue to perform better, and will also be much more sample efficient than has been previously appreciated.  Big models may be more important than big data.  In this context, further investigation into model parallelism is warranted. Deep models can be trained using pipelining \cite{DBLP:journals/corr/abs-1811-06965}, which splits parameters depth-wise between devices, but eventually requires increased batch sizes as more devices are used.  Wide networks on the other hand are more amenable to parallelization \cite{shazeer2018meshtensorflow}, since large layers can be split between multiple workers with less serial dependency.  Sparsity \cite{DBLP:journals/corr/abs-1904-10509,gray2017gpu} or branching (e.g. \cite{Krizhevsky:2012:ICD:2999134.2999257}) may allow for even faster training of large networks through increased model parallelism.  And using methods like \cite{Wang_2017,wen2019autogrow}, which grow networks as they train, it might be possible to remain on the compute-efficient frontier for an entire training run.

\section*{Acknowledgements}

We would like to thank Shan Carter, Paul Christiano, Jack Clark, Ajeya Cotra, Ethan Dyer, Jason Eisner, Danny Hernandez, Jacob Hilton, Brice Menard, Chris Olah, and Ilya Sutskever for discussions and for feedback on drafts of this work.

\newpage
\appendix
\appendixpage
\addappheadtotoc

\section{Summary of Power Laws}

For easier reference, we provide a summary below of the key trends described throughout the paper.

\begin{table}[h!]
\centering
\vspace{-0.5em}
\begin{tabular}{|c|c|c|c|l|}
\hline 
\textbf{Parameters}  & \textbf{Data}  & \textbf{Compute}  & \textbf{Batch Size}  & \textbf{Equation}\tabularnewline
\hline 
\hline 
$N$  & $\infty$  & \multicolumn{1}{c|}{$\infty$ } & Fixed  & $L\left(N\right)=\left(N_{{\rm c}}/N\right)^{\alpha_{N}}$\tabularnewline
\hline 
$\infty$  & $D$  & \multicolumn{1}{c|}{Early Stop } & Fixed  & $L\left(D\right)=\left(D_{{\rm c}}/D\right)^{\alpha_{D}}$\tabularnewline
\hline 
Optimal  & $\infty$  & $C$  & Fixed  & $L\left(C\right)=\left(C_{{\rm c}}/C\right)^{\alpha_{C}}$ (naive)\tabularnewline
\hline 
$N_{{\rm opt}}$ & $D_{{\rm opt}}$ & $C_{{\rm min}}$  & $B\ll B_{{\rm crit}}$  & $L\left(C_{{\rm min}}\right)=\left(C_{{\rm c}}^{{\rm min}}/C_{{\rm min}}\right)^{\alpha_{C}^{{\rm min}}}$\tabularnewline
\hline 
$N$  & $D$  & \multicolumn{1}{c|}{Early Stop } & Fixed  & $L\left(N,D\right)=\left[\left(\frac{N_{{\rm c}}}{N}\right)^{\frac{\alpha_{N}}{\alpha_{D}}}+\frac{D_{c}}{D}\right]^{\alpha_{D}}$\tabularnewline
\hline 
$N$  & $\infty$  & $S$ steps & $B$  & $L\left(N,S\right)=\left(\frac{N_{{\rm c}}}{N}\right)^{\alpha_{N}}+\left(\frac{S_{{\rm c}}}{S_{{\rm min}}\left(S,B\right)}\right)^{\alpha_{S}}$\tabularnewline
\hline 
\end{tabular}
\vspace{0.5em}
\caption[Key trend equations]{}
\vspace{-1em}
\end{table}

The empirical fitted values for these trends are:

\begin{table}[h!]
\centering
\vspace{-0.5em}
\begin{tabular}{|l|l|}
\hline 
\textbf{Power Law}  & \textbf{Scale (tokenization-dependent)}\tabularnewline
\hline 
\hline 
$\alpha_{N}=0.076$  & $N_{{\rm c}}=8.8\times10^{13}$ params (non-embed)\tabularnewline
\hline 
$\alpha_{D}=0.095$  & $D_{{\rm c}}=5.4\times10^{13}$ tokens\tabularnewline
\hline 
$\alpha_{C}=0.057$  & $C_{{\rm c}}=1.6\times10^{7}$ PF-days\tabularnewline
\hline 
$\alpha_{C}^{{\rm min}}=0.050$  & $C_{{\rm c}}^{{\rm min}}=3.1\times10^{8}$ PF-days\tabularnewline
\hline 
$\alpha_{B}=0.21$  & $B_{\ast}=2.1\times10^{8}$ tokens\tabularnewline
\hline 
$\alpha_{S}=0.76$  & $S_{{\rm c}}=2.1\times10^{3}$ steps\tabularnewline
\hline 
\end{tabular}
\vspace{0.5em}
\caption[Key parameters to trend fits]{}
\vspace{-1em}
\end{table}

The optimal parameters for compute efficient training are given by:

\begin{table}[h!]
\centering
\vspace{-0.5em}
\begin{tabular}{|l|l|l|}
\hline 
\textbf{Compute-Efficient Value} & \textbf{Power Law} & \textbf{Scale}\tabularnewline
\hline 
\hline 
$N_{{\rm opt}}=N_{e}\cdot C_{{\rm min}}^{p_{N}}$ & $p_{N}=0.73$ & $N_{e}=1.3\cdot10^{9}$ params\tabularnewline
\hline 
$B \ll B_{{\rm crit}}=\frac{B_{\ast}}{L^{1/\alpha_{B}}}=B_{e}C_{{\rm min}}^{p_{B}}$ & $p_{B}=0.24$ & $B_{e}=2.0\cdot10^{6}$ tokens\tabularnewline
\hline 
$S_{{\rm min}}=S_{e}\cdot C_{{\rm min}}^{p_{S}}$ (lower bound) & $p_{S}=0.03$ & $S_{e}=5.4\cdot10^{3}$ steps\tabularnewline
\hline 
$D_{{\rm opt}}=D_{e}\cdot C_{{\rm min}}^{p_{D}}$ (1 epoch) & $p_{D}=0.27$ & $D_{e}=2\cdot10^{10}$ tokens\tabularnewline
\hline 
\end{tabular}
\vspace{0.5em}
\caption[Trends for compute-efficient training]{}
\vspace{-1em}
\end{table}

\section{Empirical Model of Compute-Efficient Frontier}
\label{app:ComputeEfficientTraining}

Throughout this appendix all values of $C, S,$ and $\alpha_C$ are adjusted for training at the critical batch size $B_{\rm crit}$.  We have left off the `adj' label to avoid cluttering the notation. 

\subsection{Defining Equations}
The power-law fit to the learning curves implies a simple prescription for compute-efficient training.
In this appendix, we will derive the optimal performance, model size, and number of training steps as a function of the compute budget.
We start with the Equation \eqref{eq:FundamentalLikelihioodvsModelandSteps}, repeated here for convenience:
\begin{equation}
L\left(N,S\right)=\left(\frac{N_{c}}{N}\right)^{\alpha_{N}}+\left(\frac{S_{c}}{S}\right)^{\alpha_{S}}.
\end{equation}
Here, $S$ represents the number of parameter updates when training \textbf{at the critical batch size} \cite{1812.06162}, which was defined in Equation \eqref{eq:DefinitionBcrit}\footnote{There is a slight ambiguity here: we can imagine training either at a constant batch size $B\left(L_{{\rm target}}\right)$, or we could instead train at a variable batch size $\tilde{B}\left(L\right)$, where $\tilde{B}$ is the instantaneous critical batch size (as opposed to $B$, which is the averaged version). These two prescriptions result in the same number of steps, so we can ignore this subtlety (see \cite{1812.06162}).}:
\begin{equation}
B\left(L\right)=\frac{B_{\ast}}{L^{1/\alpha_{B}}}.
\end{equation}
We would like to determine optimal training parameters for a fixed compute budget, so we replace $S=C/\left(6NB\left(L\right)\right)$, where $C$ is the number of FLOPs used in the training run:
\begin{equation}
L\left(N,C\right)=\left(\frac{N_{c}}{N}\right)^{\alpha_{N}}+\left(6B_{\ast}S_{c}\frac{N}{L^{1/\alpha_{B}}C}\right)^{\alpha_{S}}.\label{eq:loss-params-compute}
\end{equation}
Now, we set $\partial_{N}L\big|_{C}=0$ to find the condition for optimality:
\begin{align}
0 & =\frac{\partial L}{\partial N}\big|_{C}\nonumber \\
 & =-\frac{\alpha_{N}}{N}\left(\frac{N_{c}}{N}\right)^{\alpha_{N}}+\frac{\alpha_{S}}{N}\left(6B_{\ast}S_{c}\frac{N}{L^{1/\alpha_{B}}C}\right)^{\alpha_{S}}\left(1-5\frac{N}{L}\cancel{\frac{\partial L}{\partial N}\big|_{C}}\right)\nonumber \\
\implies\frac{\alpha_{N}}{\alpha_{S}}\left(\frac{N_{c}}{N}\right)^{\alpha_{N}} & =\left(6B_{\ast}S_{c}\frac{N}{L^{1/\alpha_{B}}C}\right)^{\alpha_{S}}\label{eq:compute-optimality}
\end{align}
Equation \eqref{eq:loss-params-compute} and \eqref{eq:compute-optimality} together determine the compute-efficient frontier.

\subsection{Efficient Training}
Now we assemble the implications of \eqref{eq:loss-params-compute} and \eqref{eq:compute-optimality}.
First, note that inserting \eqref{eq:compute-optimality} into \eqref{eq:loss-params-compute} yields
\begin{equation}
L\left(N_{{\rm eff}}\left(C\right),C\right)=\left(1+\frac{\alpha_{N}}{\alpha_{S}}\right)L\left(N_{{\rm eff}},\infty\right),
\end{equation}
which implies that for compute-efficient training, we should train to a \textbf{fixed percentage} $\frac{\alpha_{N}}{\alpha_{S}}\approx10\%$ above the converged loss.
Next, let's determine how the optimal loss depends on the compute budget. Eliminating $N$ yields a power-law dependence of performance on compute:
\begin{align}
L\left(C\right) & =\left(\frac{C_{c}}{C}\right)^{\alpha_{C}}
\end{align}
where we defined
\begin{align}
\alpha_{C} & =1/\left(1/\alpha_{S}+1/\alpha_{B}+1/\alpha_{N}\right)\approx0.052\\
C_{c} & =6N_{c}B_{\ast}S_{c}\left(1+\frac{\alpha_{N}}{\alpha_{S}}\right)^{1/\alpha_{S}+1/\alpha_{N}}\left(\frac{\alpha_{S}}{\alpha_{N}}\right)^{1/\alpha_{S}}.
\end{align}
Similarly, we can eliminate $L$ to find $N\left(C\right)$:
\begin{align}
\frac{N\left(C\right)}{N_{c}}=\left(\frac{C}{C_{c}}\right)^{\alpha_{C}/\alpha_{N}}\left(1+\frac{\alpha_{N}}{\alpha_{S}}\right)^{1/\alpha_{N}}
\end{align}
and
\begin{align}
S\left(C\right) & =\frac{C_{c}}{6N_{c}B_{\ast}}\left(1+\frac{\alpha_{N}}{\alpha_{S}}\right)^{-1/\alpha_{N}}\left(\frac{C}{C_{c}}\right)^{\alpha_{C}/\alpha_{S}}
\end{align}

\subsection{Comparison to Inefficient}
Typically, researchers train models until they appear to be close to convergence. In this section, we compare the efficient training procedure described above to this more typical setup.
We define a the convergence factor $f$ as the percent deviation from the converged loss:
\begin{equation}
L\left(N,C\right)=\left(1+f\right)L\left(N,\infty\right).
\end{equation}
For compute-efficient training we have $f=\alpha_{N}/\alpha_{S}\approx10\%$ from the previous section, but researchers typically use a much smaller value. Here, we choose $f^{\prime}=2\%$ as an estimate.
For a fixed value of the loss, we predict:
\begin{align}
\frac{N_{f}}{N_{f^{\prime}}} & =\left(\frac{1+f}{1+f^{\prime}}\right)^{1/\alpha_{N}}\approx2.7\\
\frac{S_{f}}{S_{f^{\prime}}} & =\left(\frac{1+\frac{1}{f}}{1+\frac{1}{f^{\prime}}}\right)^{1/\alpha_{S}}\approx0.13\\
\frac{C_{f}}{C_{f^{\prime}}} & =\frac{N_{f}}{N_{f^{\prime}}}\frac{S_{f}}{S_{f^{\prime}}}\approx0.35
\end{align}
So that compute-efficient training uses 7.7x fewer parameter updates, 2.7x more parameters, and 65\% less compute to reach the same loss.

\subsection{Suboptimal Model Sizes}
\label{sec:suboptimal-models}
We can solve A.1 to find an expression for the amount of compute needed to reach a given value of the loss $L$ with a model of size $N$:
\be
C\left(N,L\right)=\left(6B_{\ast}S_{c}\frac{N}{L^{1/\alpha_{B}}}\right)\left(L-\left(\frac{N_{c}}{N}\right)^{\alpha_{N}}\right)^{-1/\alpha_{S}}.
\ee
Using A.6 and A.9, we can eliminate $L$ in favor of $N_{{\rm eff}}\left(L\right)$, the model size which reaches $L$ most efficiently. From there, we find an expression for the excess compute needed as a consequence of using a suboptimal model size: 
\be
\frac{C\left(N,N_{{\rm eff}}\right)}{C\left(N_{{\rm eff}},N_{{\rm eff}}\right)}=\frac{N}{N_{{\rm eff}}}\left[1+\frac{\alpha_{S}}{\alpha_{N}}\left(1-\left(\frac{N_{{\rm eff}}}{N}\right)^{\alpha_{N}}\right)\right]^{-1/\alpha_{S}}.
\ee
The result is shown in Figure X. Models between 0.6x and 2.2x the optimal size can be used with only a 20\% increase in compute budget. Using a smaller model is useful when accounting for the cost inference. A larger model can be trained the the same level of performance in fewer steps, allowing for more parallelism and faster training if sufficient harware is available (see Figure Y):
\be
\frac{S\left(N,N_{{\rm eff}}\right)}{S\left(N_{{\rm eff}},N_{{\rm eff}}\right)}=\left[1+\frac{\alpha_{S}}{\alpha_{N}}\left(1-\left(\frac{N_{{\rm eff}}}{N}\right)^{\alpha_{N}}\right)\right]^{-1/\alpha_{S}}.
\ee
A 2.2x larger model requires 45\% fewer steps at a cost of 20\% more training compute. Note that this equation should not be trusted for very large models, as it is only valid in the power-law region of the learning curve after initial transient effects.

\section{Caveats}

In this section we list some potential caveats to our analysis.

\begin{itemize}
\item At present we do not have a solid theoretical understanding for any of our proposed scaling laws.  The scaling relations with model size and compute are especially mysterious.  It may be possible to understand scaling at very large $D$ holding model size fixed \cite{1710.03667}, and also the shape of learning curves late in training, by modeling the loss with a noisy quadratic.  But the scaling with $D$ at very large model size still remains mysterious.  Without a theory or a systematic understanding of the corrections to our scaling laws, it's difficult to determine in what circumstances they can be trusted.  

\item  We are not especially confident in the prediction of $B_{\rm crit}(L)$ for values of the loss far outside the range we have explored.  Changes in $B_{\rm crit}$ could have a significant impact on trade-offs between data parallelism and the number of serial training steps required, which would have a major impact on training time.

\item We did not thoroughly investigate the small data regime, and our fits for $L(N,D)$ were poor for the smallest values of $D$ (where an epoch corresponded to only $40$ steps). Furthermore, we did not experiment with regularization and data augmentation. Improvements in these could alter our results,  quantitatively or qualitatively. 

\item We used the estimated training compute $C \approx 6 N B S$, which did not include contributions proportional to  $n_{\rm ctx}$ (see Section \ref{sec:ParameterComputeCounts}).  So our scalings with compute may be confounded in practice in the regime of very large $n_{\rm ctx}$, specifically where $n_{\rm ctx} \gtrsim 12 d_{\rm model}$.

\item We tuned learning rates, and we experimented with learning rate schedules.  But we may have neglected to tune some hyperparameter (e.g. intialization scale or  momentum) that have an important effect on scaling.

\item The optimal choice of learning rate is sensitive to the target loss. When training close to convergence, it may be necessary to use a smaller learning rate to avoid divergences.  But when conducting a short training run (eg due to compute limitations), it may be possible to use a larger learning rate.  We did not experiment with higher learning rates for training runs that did not proceed to convergence.
\end{itemize}

\section{Supplemental  Figures}

 \begin{figure}
\noindent \centering{} 
\includegraphics[width=0.48\textwidth]{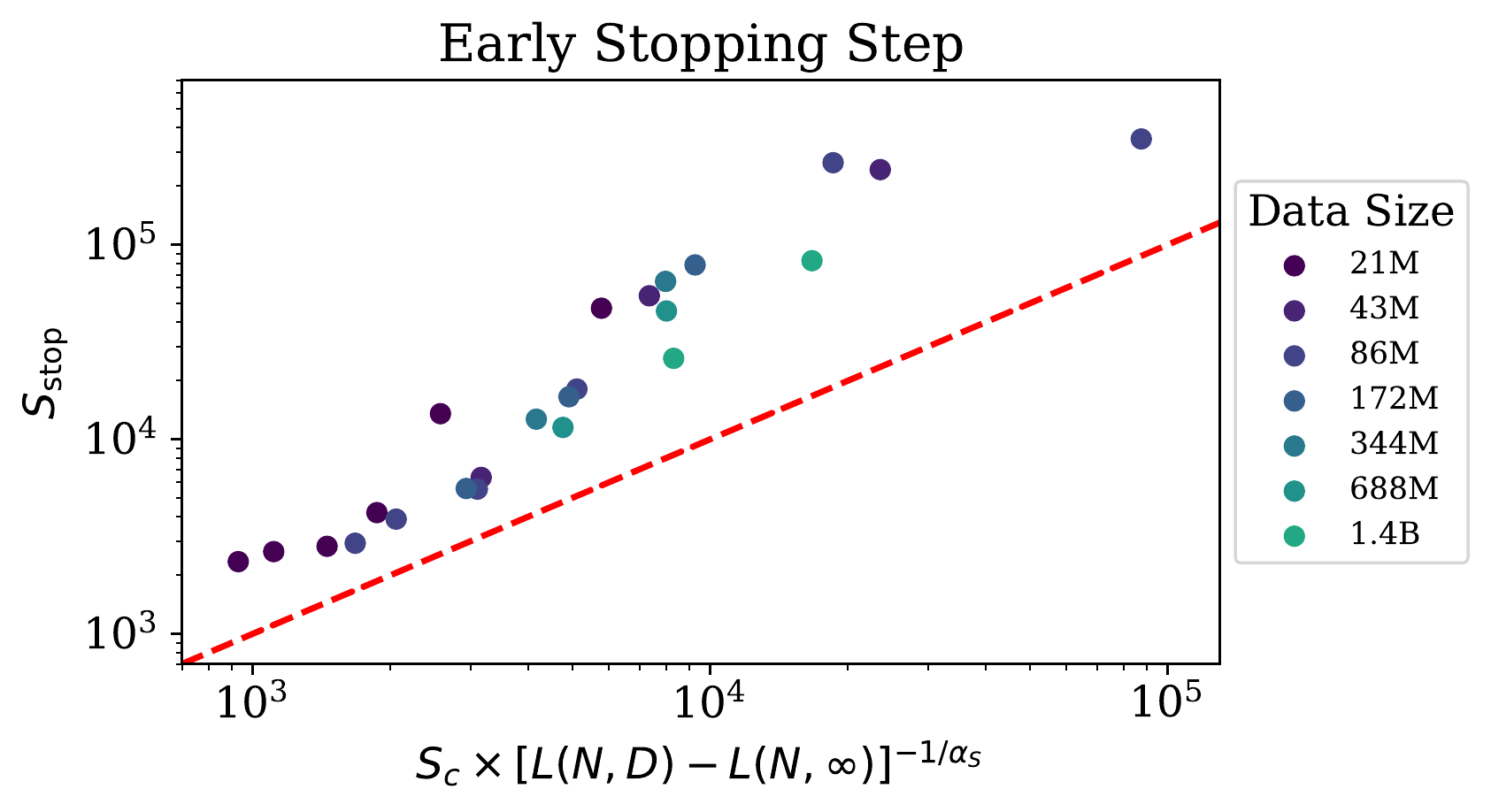}\hfill
\includegraphics[width=0.48\textwidth]{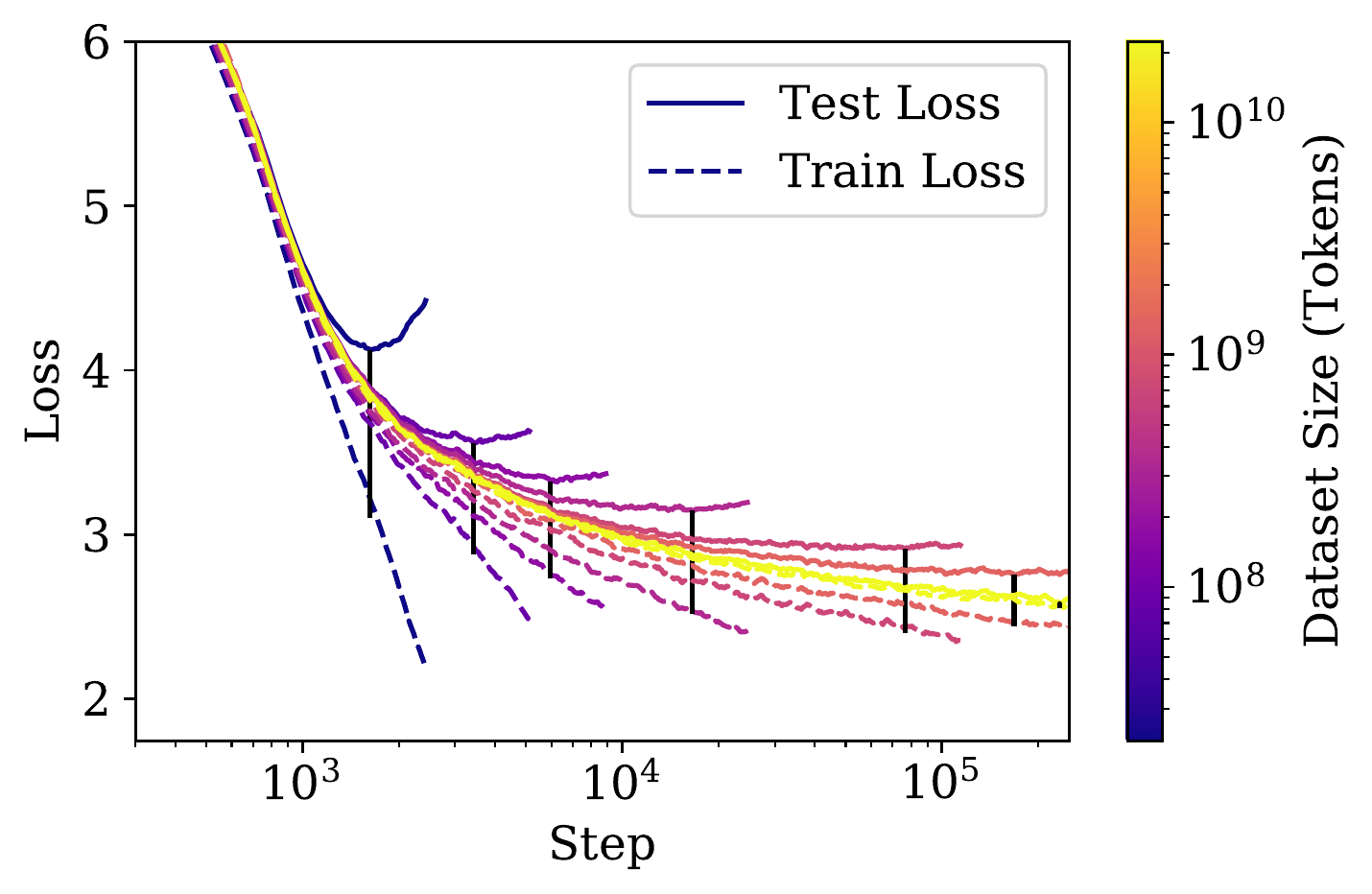}
\caption[Early stopping lower bound and training curves for overfit models]{ {\bf Left:}  We characterize the step on which early stopping occurs, as a function of the extent of overfitting.  The red line indicates a {lower bound} for early stopping that is derived in Section \ref{sec:EarlyStop}.  {\bf Right:} We display train and test loss for a series of 300M parameter models trained on different sized dataset sub-samples.  The test loss typically follows that of a run done with unrestricted data until diverging. Note that the degree of overfitting (as compared to the infinite data limit) is significantly overestimated by $L_{\rm test} - L_{\rm train}$ (denoted by a black bar for each run). \label{fig:OverfittingandEarlyStopping}}
\end{figure}

\begin{figure}
\noindent \centering{} 
\includegraphics[width=0.48\textwidth]{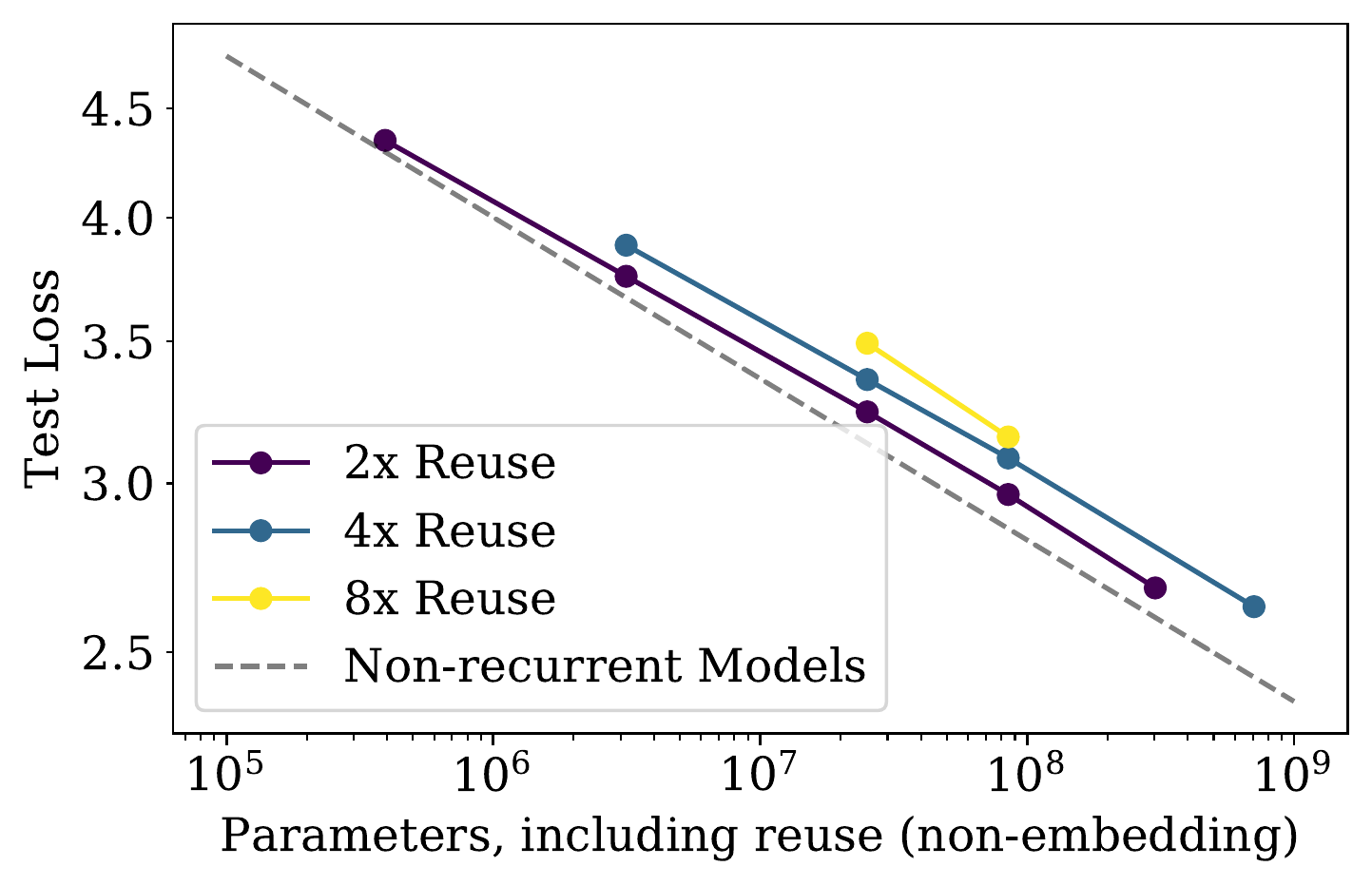}\hfill
\includegraphics[width=0.48\textwidth]{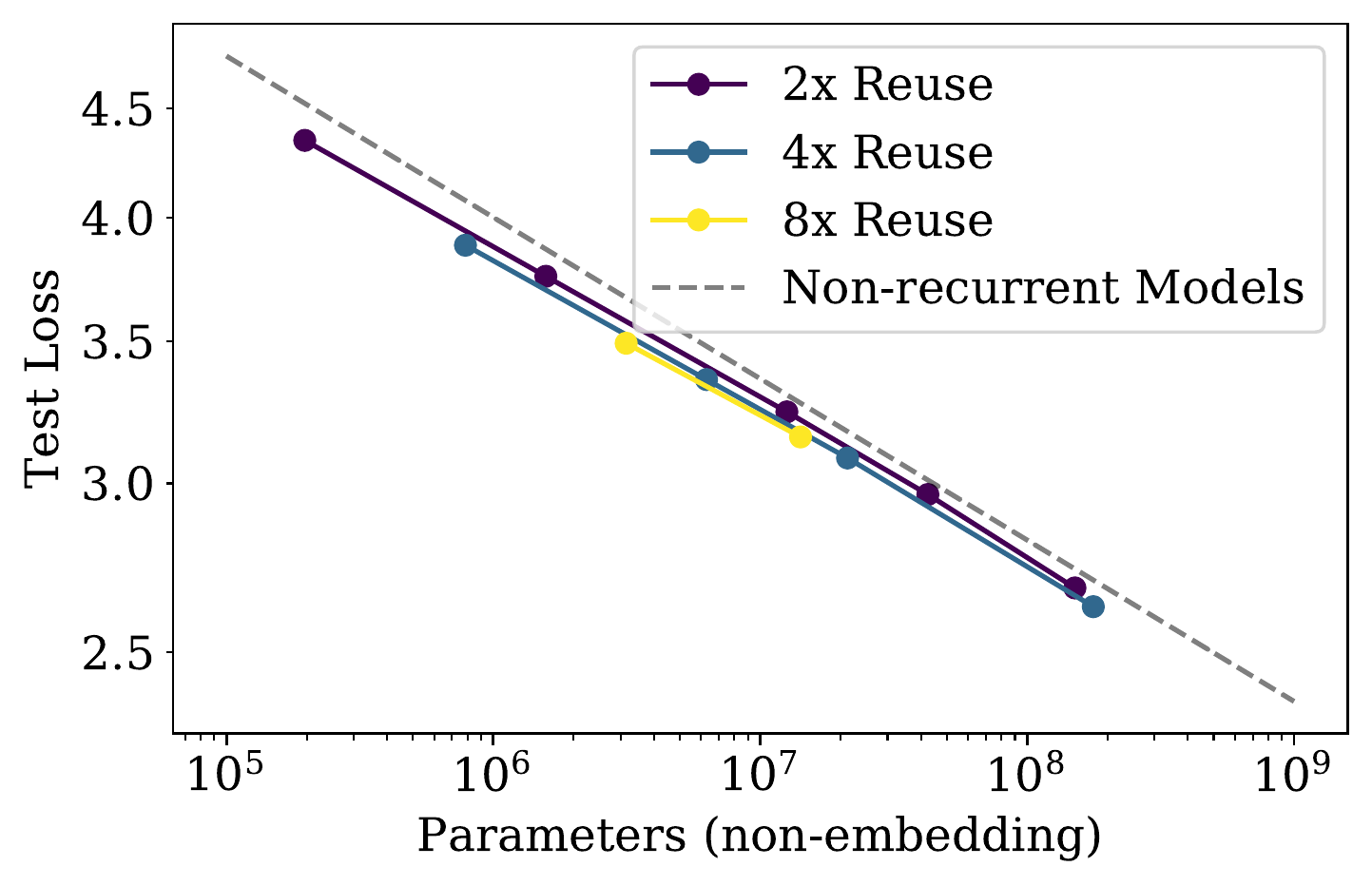}
\caption[Universal transformers]{We compare recurrent Transformers \cite{DBLP:journals/corr/abs-1807-03819}, which re-use parameters, to standard Transformers. Recurrent Transformers perform slightly better when comparing models with equal parameter count, but slightly worse when accounting for reuse and comparing per FLOP. \label{fig:RecurrentTransformers}}
\end{figure}

\subsection{Early Stopping and Test vs Train}
\label{sec:OverfittingandEarlyStopping}

In section \ref{sec:EarlyStop} we described the result shown in Figure \ref{fig:OverfittingandEarlyStopping}, which provides a prediction for a lower bound on the early stopping step.  We also show the train and test loss for a given model size when training on different sized datasets.

\subsection{Universal Transformers}

We  compare the performance of standard Transformers  to recurrent Transformers \cite{DBLP:journals/corr/abs-1807-03819} in Figure \ref{fig:RecurrentTransformers}.  These models re-use parameters, and so perform slightly better as a function of $N$, but slightly worse as a function of compute $C$.  We include several different different possibilities for parameter re-use.

\subsection{Batch Size}

\begin{figure}
\noindent \centering{} 
\includegraphics[width=0.48\textwidth]{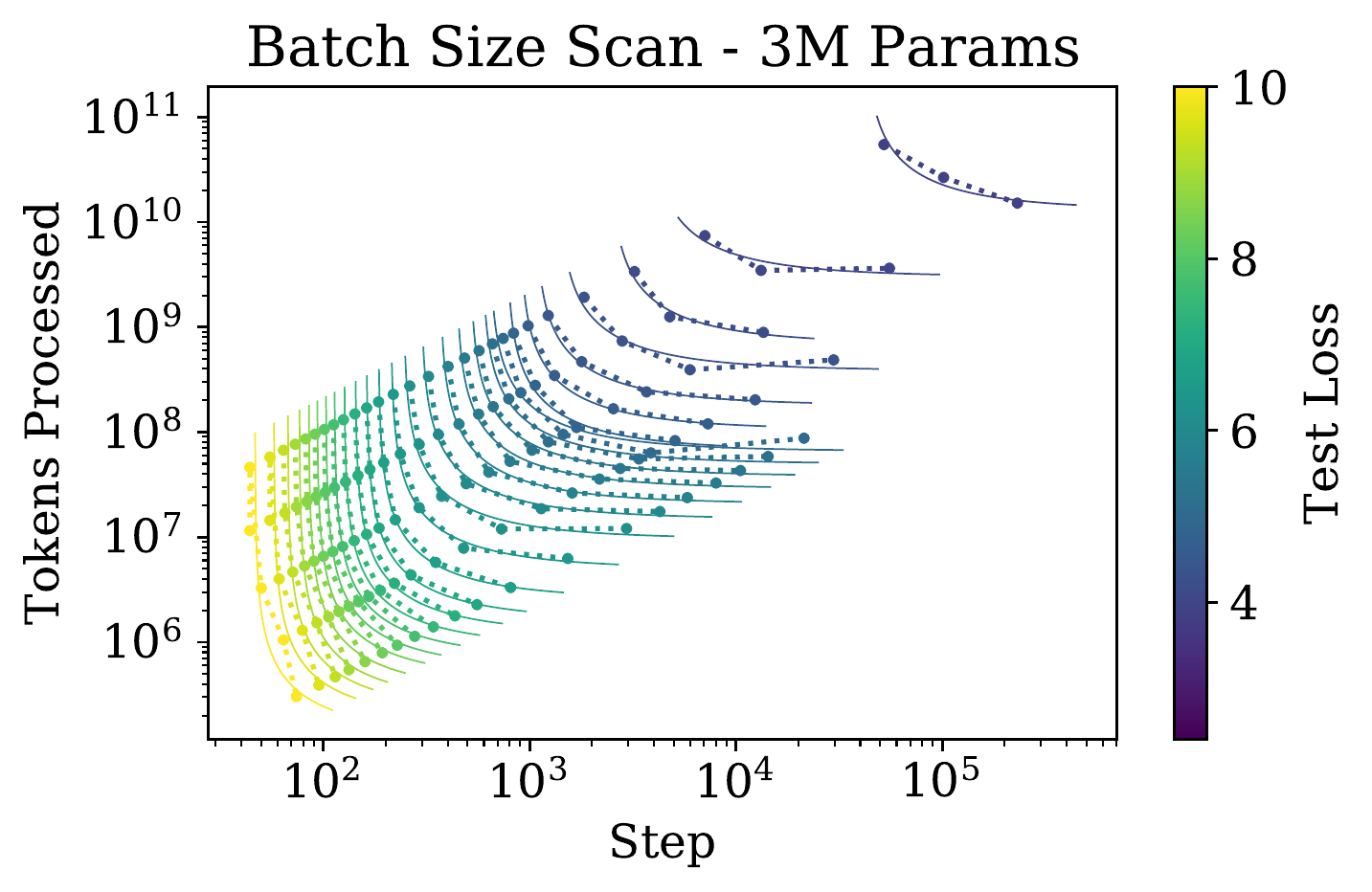} 
\includegraphics[width=0.48\textwidth]{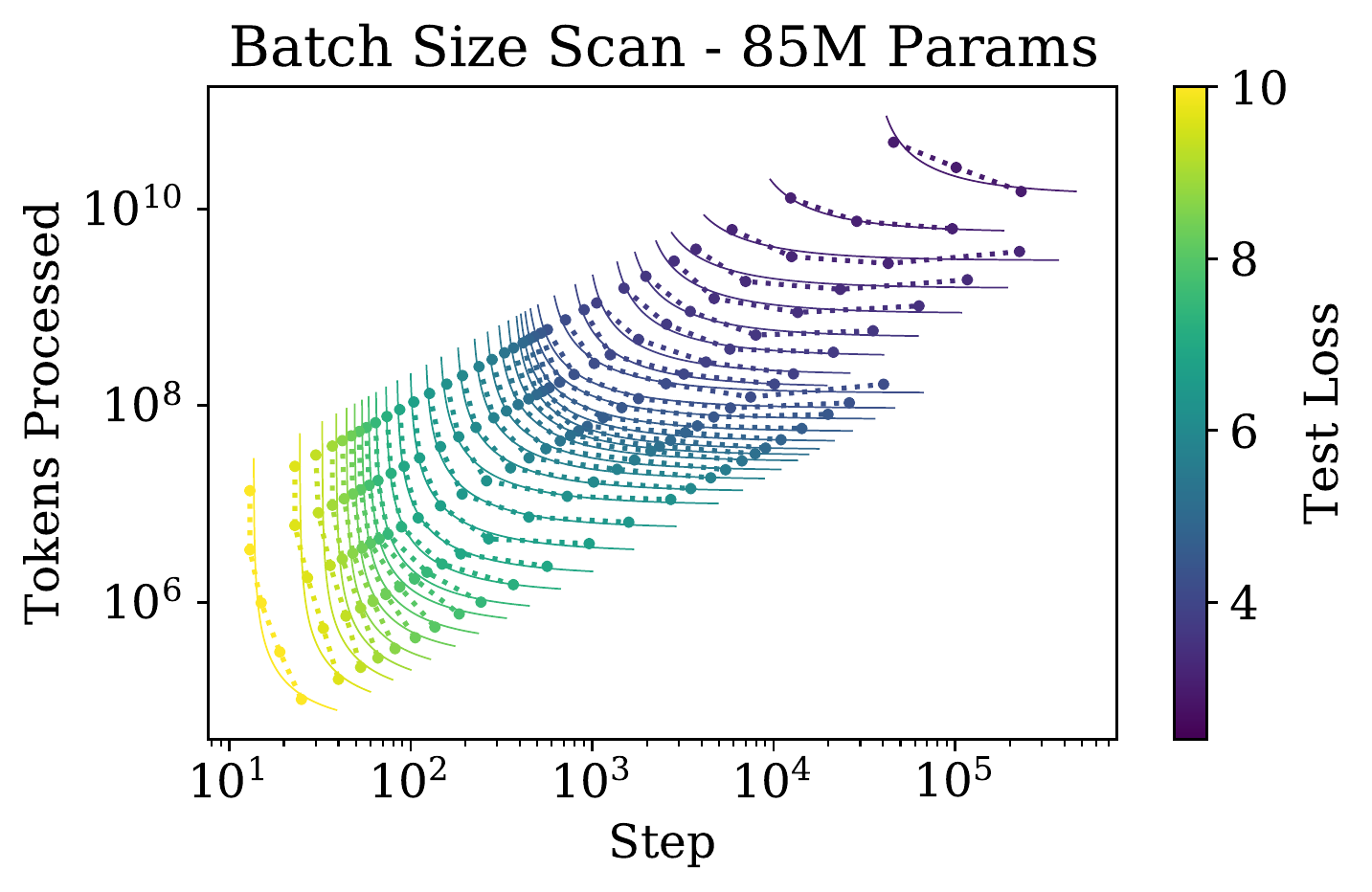}
 \caption[Batch size scans]{These figures demonstrate fits to Equation \eqref{eq:TimeComputeTradeoff} for a large number of values of the loss $L$, and for two different Transformer model sizes.  These fits were used to measure $B_{\rm crit}(L)$ for Figure \ref{fig:OptimalBatchSize}.  \label{fig:BatchPareto}}
\end{figure}

We measure the critical batch size using the data displayed in figure \ref{fig:BatchPareto}.  This made it possible to estimate $B_{\rm crit}(L)$ in figure \ref{fig:OptimalBatchSize}.

\subsection{Sample Efficiency vs Model Size}

\begin{figure}
\noindent \centering{}
\includegraphics[width=0.49\textwidth]{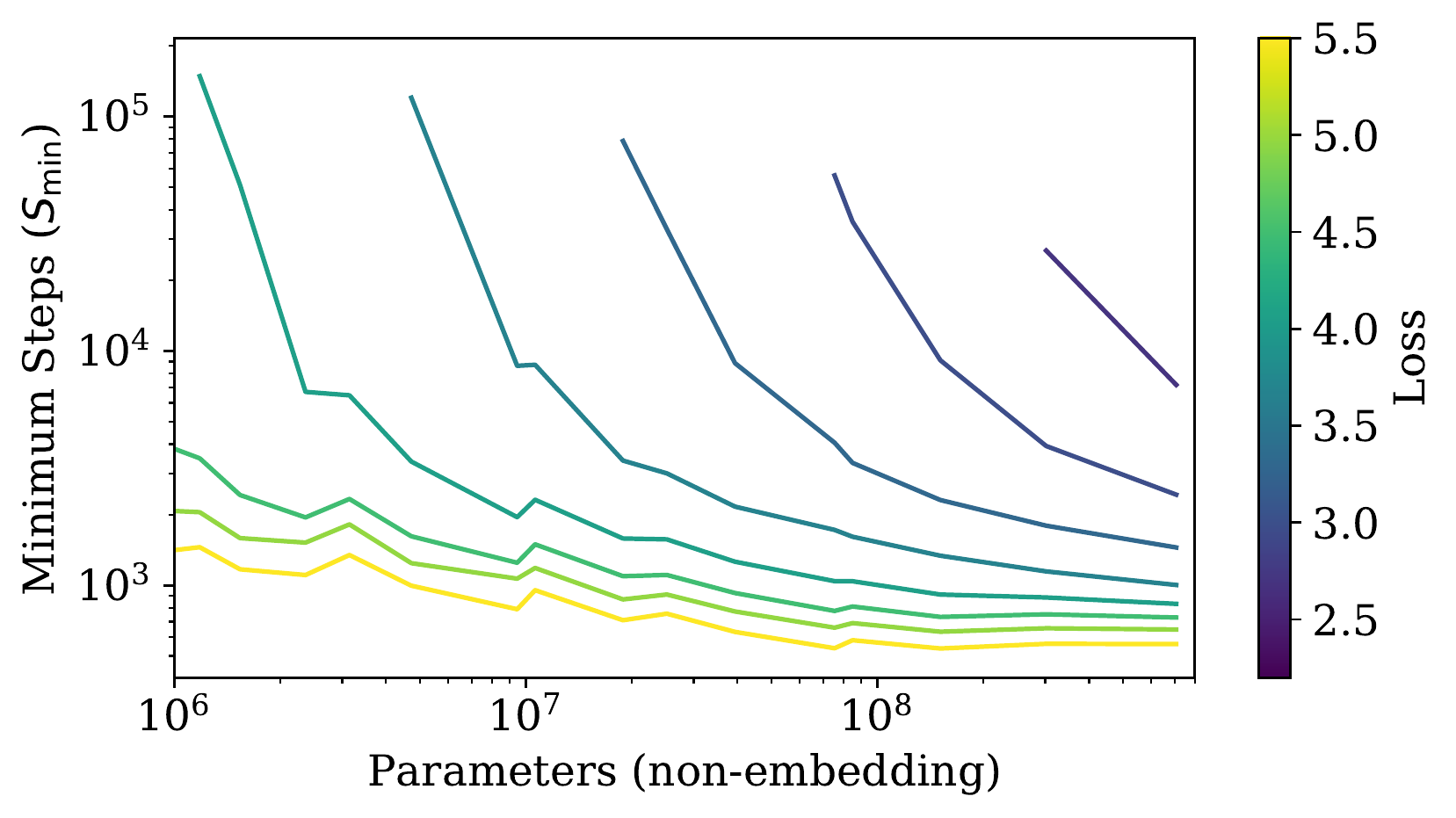}
\includegraphics[width=0.49\textwidth]{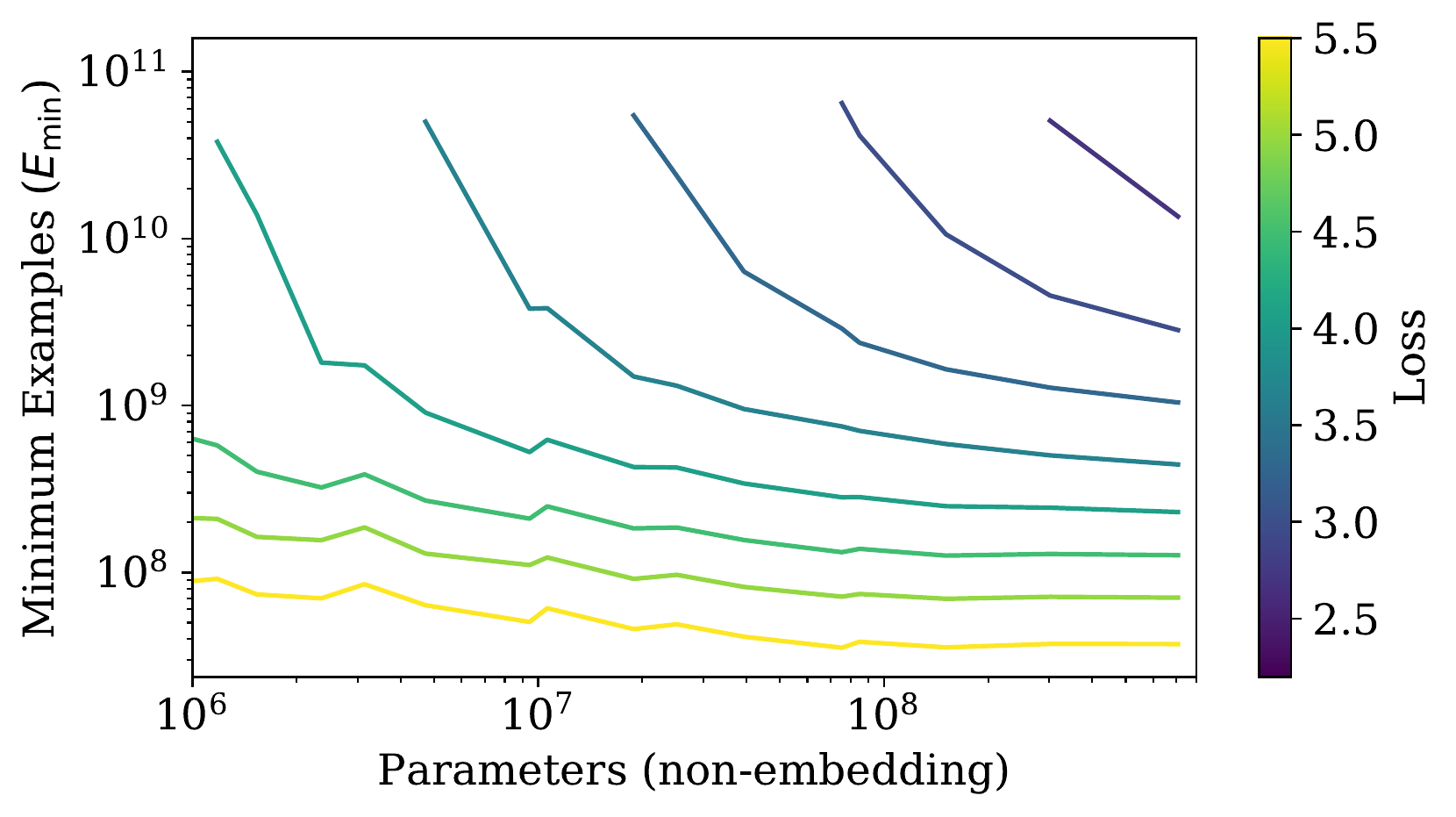}
\caption[Another look at sample efficiency]{The number of minimum serial steps needed to reach any fixed value of the test loss decreases precipitously with model size.  Sample efficiency (show here for training far below the critical batch size) improves greatly as well, improving by a factor of almost 100 when comparing the smallest possible model to a very large one.   \label{fig:SampleEfficiency}}
\end{figure}

It is easy to see from figure \ref{fig:EfficiencyIllustration} that larger models train faster, and are therefore more sample efficient.  We provide another way of looking at this phenomenon in figure \ref{fig:SampleEfficiency}, which shows when different models reach various fixed values of the loss.

\begin{figure}
\noindent \centering{} \includegraphics[width=0.50\textwidth]{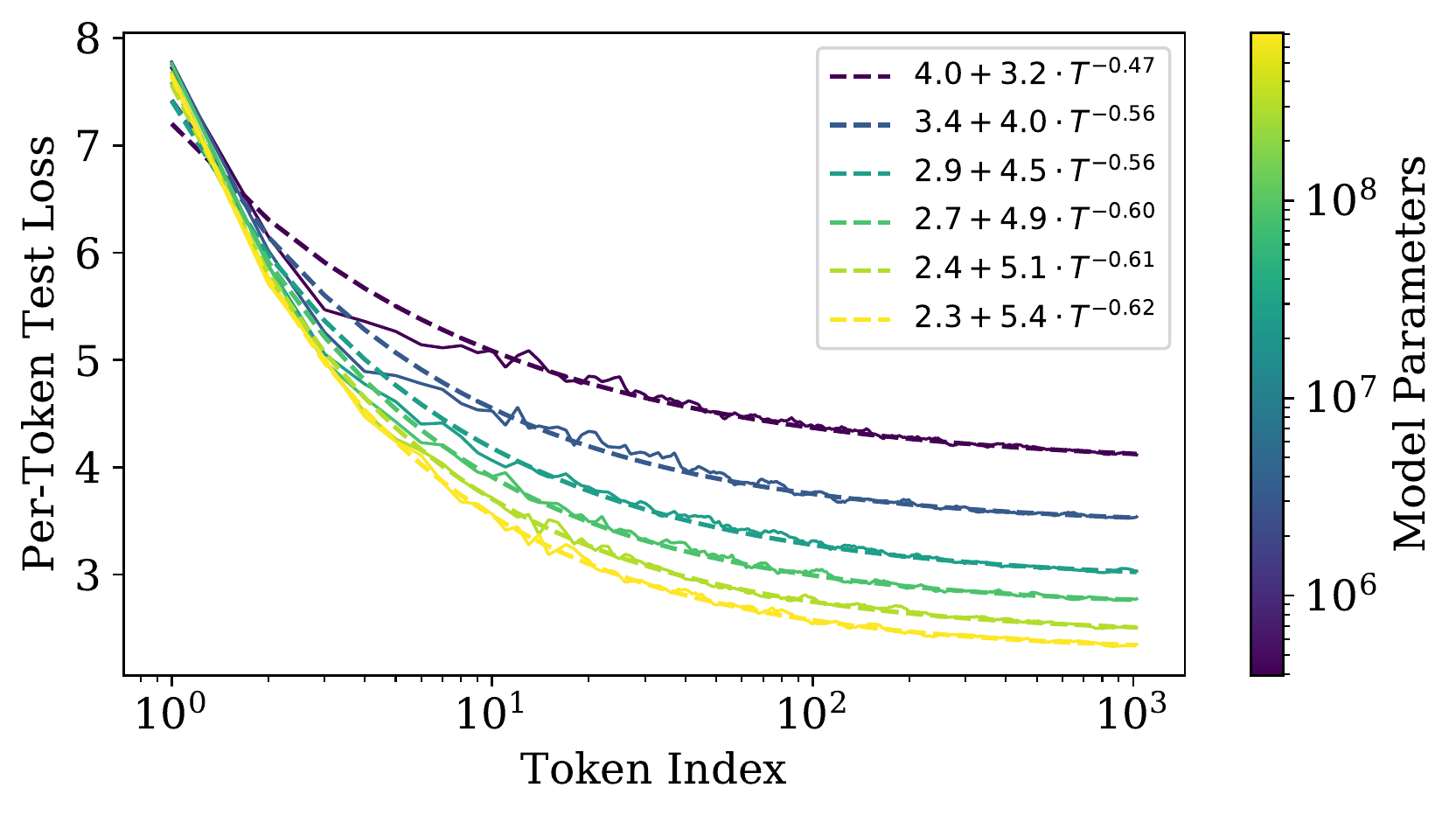} 
\includegraphics[width=0.48\textwidth]{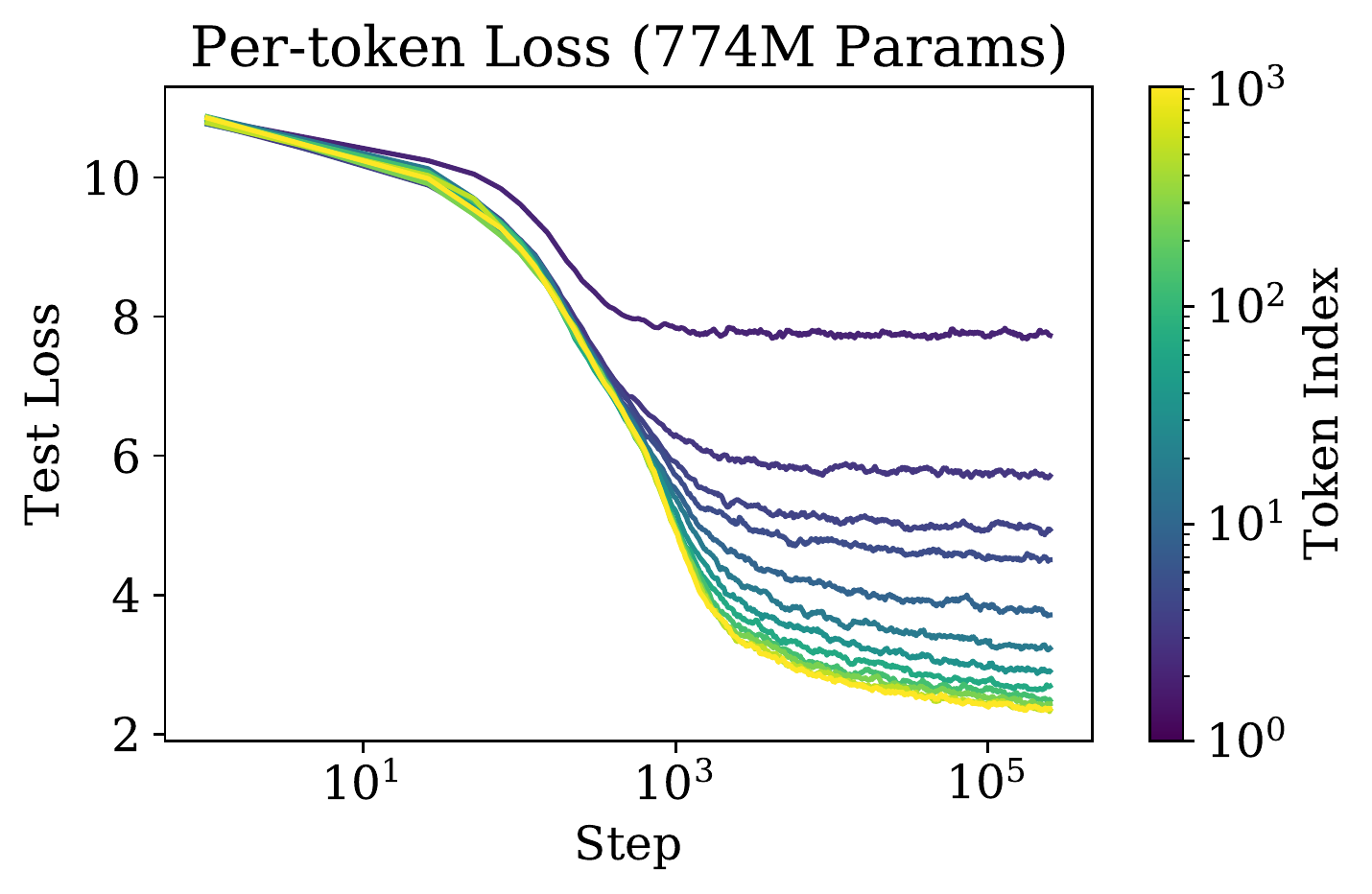}  \caption[Power-law dependence of performance on position in context]{ This figure provides information about the performance per token as a function of model size and training time.  {\bf Left:} Loss per token as a function of its position $T$ in the 1024-token context.  Loss scales predictably as a power-law in $T$.  {\bf Right: } Test loss per token as a function of training step.  \label{fig:MoreTokenAnalysis}}
\end{figure}

\subsection{Context Dependence}
\label{sec:ContextDependence}

\begin{figure}
\noindent \centering{}  \includegraphics[width=0.54\textwidth]{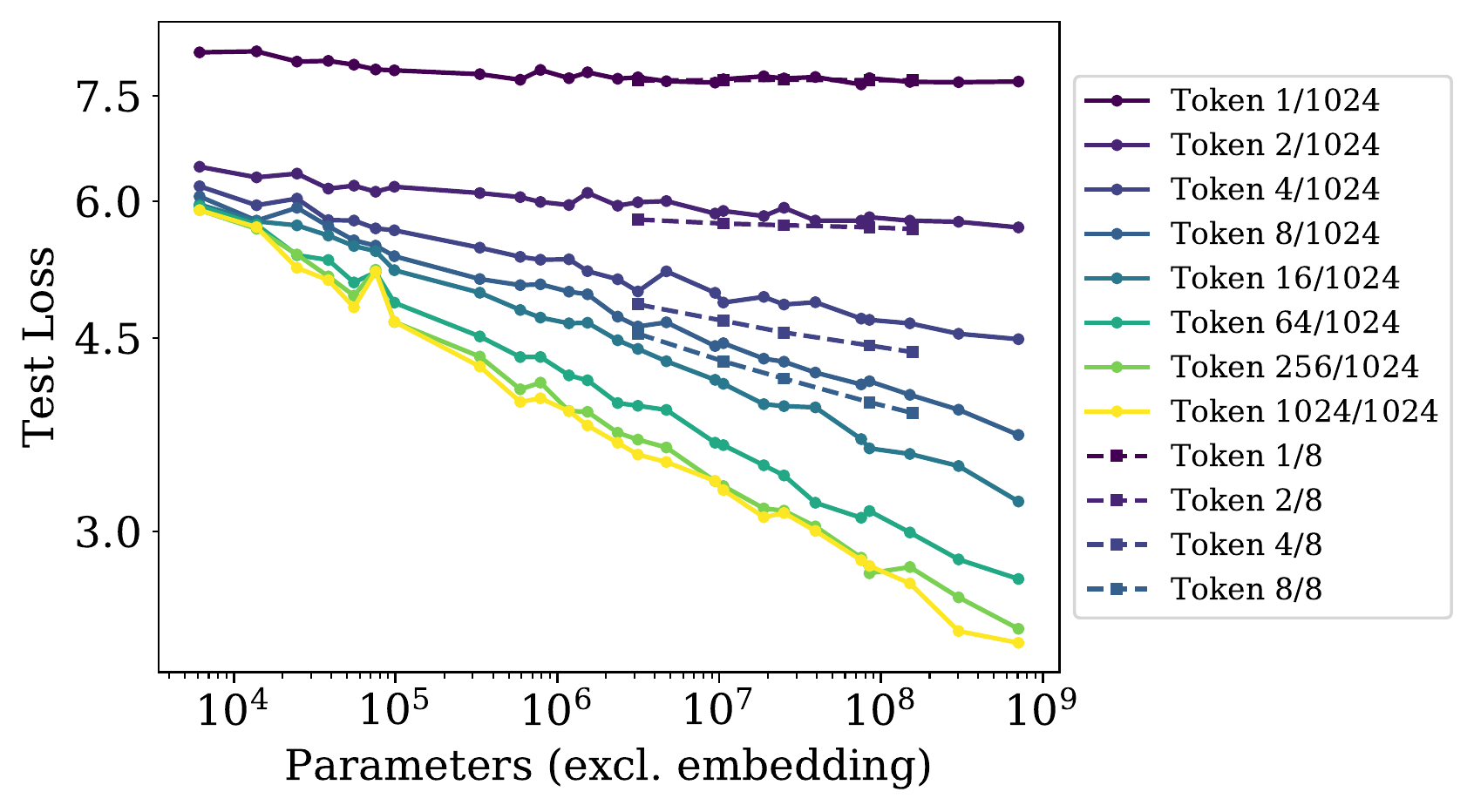}  \caption[Performance at different context positions versus model size]{In addition to the averaged loss, individual tokens within the 1024-token context also improve smoothly as model size increases.  Training runs with shorter context $n_{\rm ctx} = 8$ (dashed lines) perform better on early tokens, since they can allocate all of their capacity to them.  \label{fig:PerformancevsModelSizevsContext}}
\end{figure}

The trends for loss as a function of model size are displayed for different tokens in the context in Figure \ref{fig:PerformancevsModelSizevsContext}.  We see that models trained on $n_{\rm ctx} = 1024$ show steady improvement with model size on all but the first token.  

Fixing model size, it appears that the loss scales as a power-law as a function of position $T$ in the context, see Figure \ref{fig:MoreTokenAnalysis}.  This may be a consequence of underlying power-law correlations in language \cite{ebeling1994entropy, altmann2012origin, lin2016criticality}, or a more general feature of the model architecture and optimization.  It provides some suggestion for the potential benefits (or lack thereof) from training on larger contexts.  Not only do larger models converge to better performance at $T=1024$, but they also improve more quickly at early tokens, suggesting that larger models are more efficient at detecting patterns with less contextual information.  In the right-hand plot we show how per-token performance varies for a fixed model as a function of the training step.  The model begins by learning short-range information, and only learns longer-range correlations later in training.   

We have also included models trained with a tiny context $n_{\rm ctx} = 8$ in order  to compare with our longer context models.  Even modestly sized models trained on $n_{\rm ctx} = 8$ can dominate our largest $n_{\rm ctx} = 1024$ models on very early tokens.  This also suggests that further improvements should be possible with much larger models trained on large contexts.

\begin{figure}
\noindent \centering{} 
\includegraphics[width=0.46\textwidth]{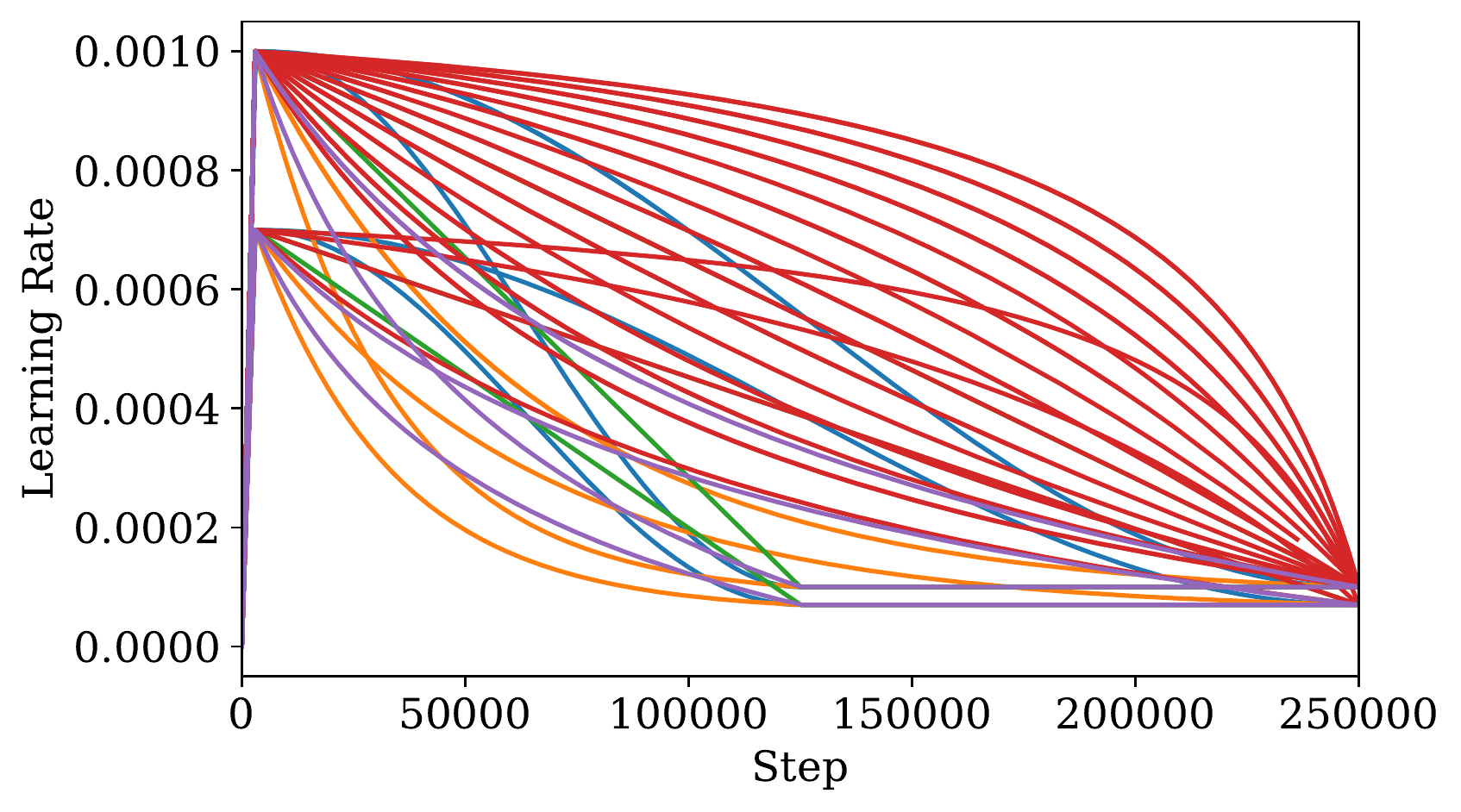} 
\includegraphics[width=0.46\textwidth]{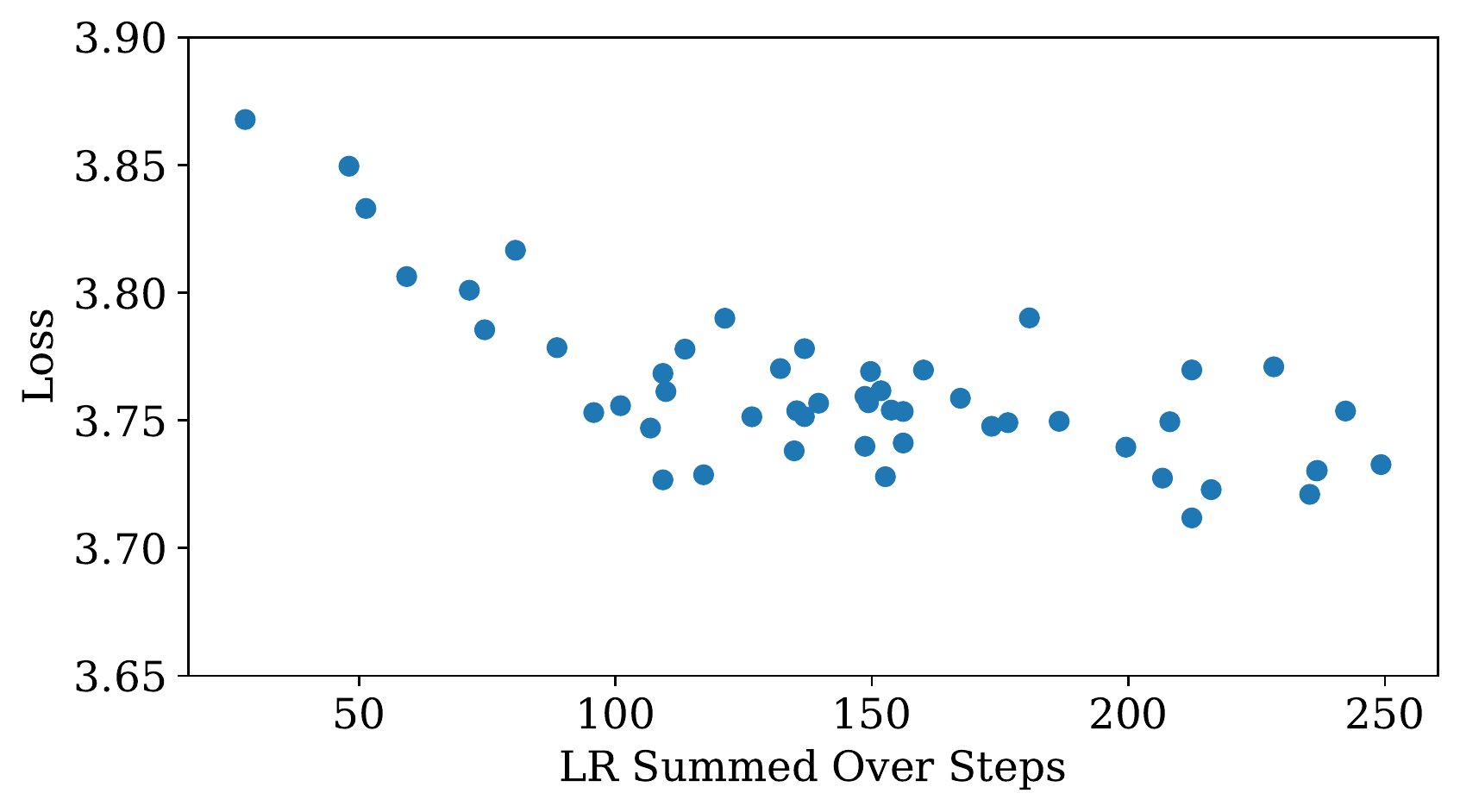} 
 \caption[Learning rate schedule scan]{
 We test a variety of learning rate schedules including cosine decay, linear decay, as well as other faster/slower decays schedules on a 3 million parameter model, shown on the left.
 For these experiments we do not decay to zero, since we find that this tends to give a fixed improvement close to the end of training.
 We find that, as long as the learning rate is not too small and does not decay too quickly, performance does not depend strongly on learning rate.
 Run-to-run variation is at the level of ~0.05 in the loss, so averaging multiple runs is necessary to validate performance changes smaller than this level.
 \label{fig:LearningRateSchedules} }
\end{figure}

\subsection{Learning Rate Schedules and Error Analysis}
\label{app:OptimizationDetailsandErrorAnalysis}

We experimented with a variety of learning rates and schedules.  A host of schedules and resulting test performances for a small language model are plotted in Figure \ref{fig:LearningRateSchedules}. We conclude that the choice of learning rate schedule is mostly irrelevant, as long as the total summed learning rate is sufficiently large, and the schedule includes a warmup period and a final decay to near-vanishing learning rate.  Variations among schedules appear to be statistical noise, and provide a rough gauge for the scale of variation between different training runs.  Experiments on larger models suggest that the variation in the final test loss between different random seeds is roughly constant in magnitude for different model sizes.

We found that larger models require a smaller learning rate to prevent divergence, while smaller models can tolerate a larger learning rate.  To implement this, the following rule of thumb was used for most runs:
\begin{equation}
\mathrm{LR}(N) \approx 0.003239 + -0.0001395  \log(N)
\end{equation}
We expect that this formula could be improved.  There may be a dependence on network width, likely set by the initialization scale.  The formula also breaks down for $N>10^{10}$ parameters.  Nevertheless, we found that it works sufficiently well for the models we considered.

\subsection{Fit Details and Power Law Quality}

\begin{figure}
\noindent \centering{} 
\includegraphics[width=0.40\textwidth]{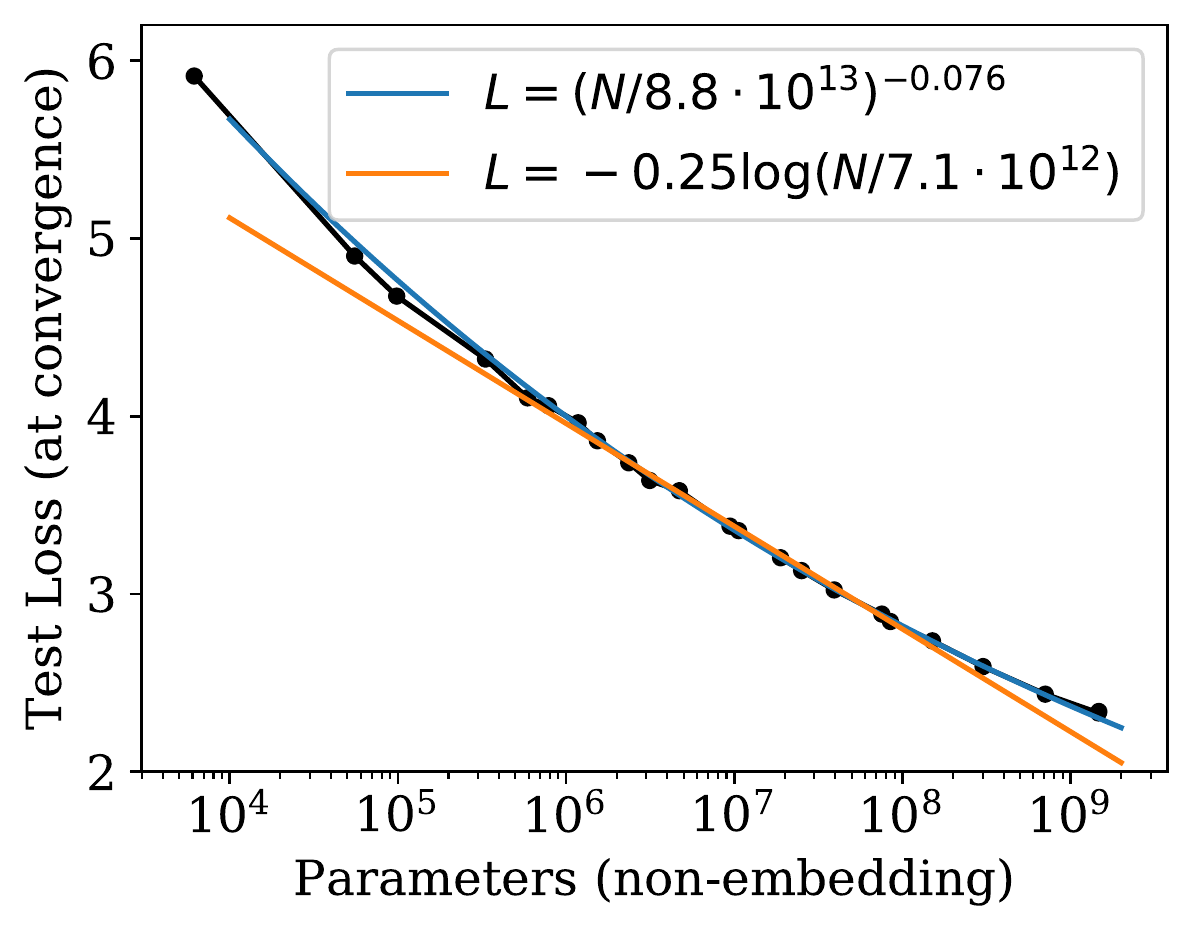}
\caption[Comparison of Power-Law and Logarithmic Fits]{The trend for performance as a function of parameter count, $L(N)$, is fit better by a power law than by other functions such as a logarithm at a qualitative level.  \label{fig:PoorLogFit}}
\end{figure}

\begin{figure}
\noindent \centering{}
\includegraphics[width=0.5\textwidth]{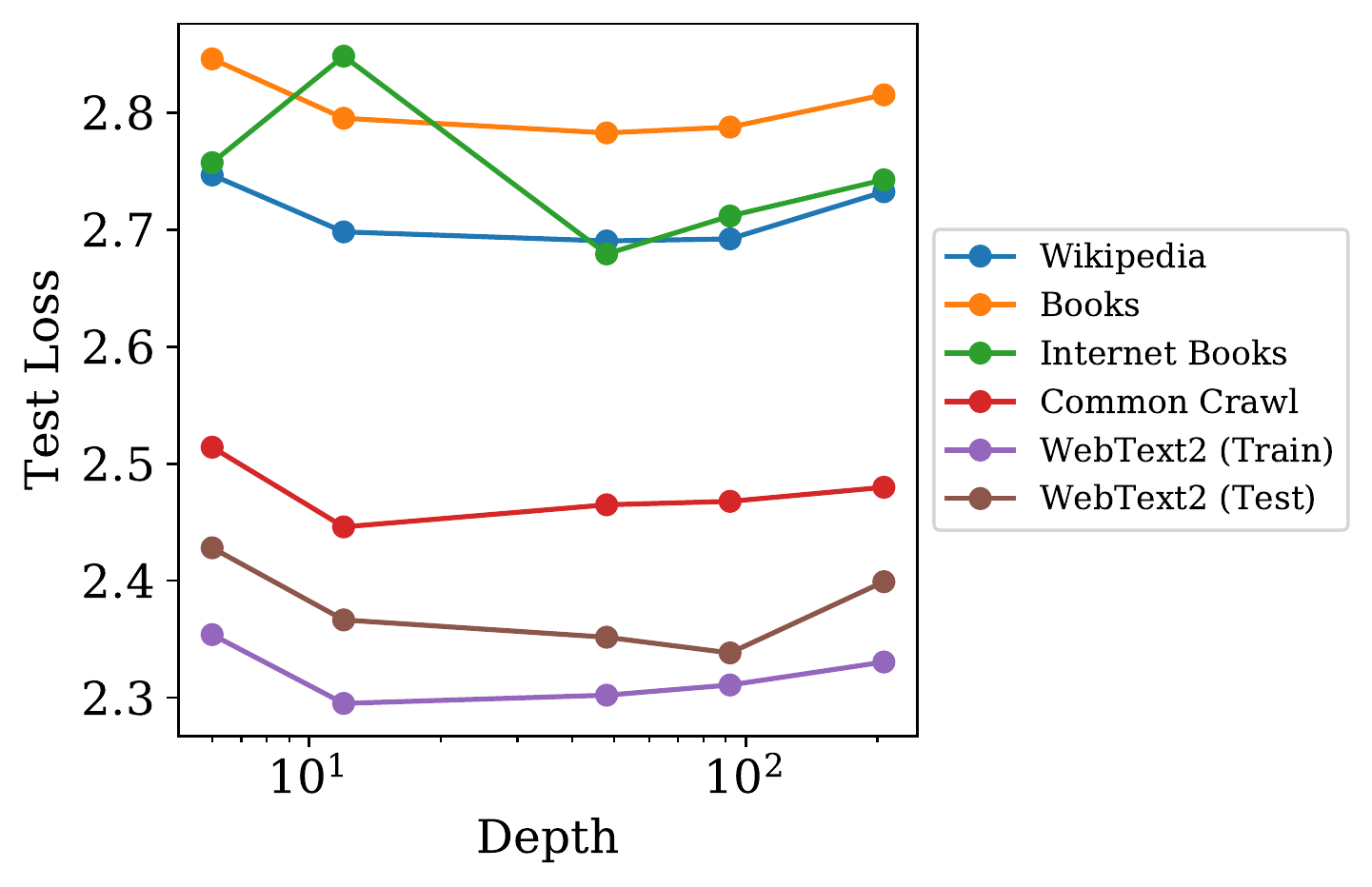}
\caption[Generalization versus depth]{We show evaluations on a series of datasets for models with approximately 1.5 Billion parameters.  We observe no effect of depth on generalization; generalization performance depends primarily on training distribution performance. The 12-layer model overfit the Internet Books dataset and we show the early-stopped performance; we have not seen this surprising result in other experiments. \label{fig:DepthVsGeneralization}}
\end{figure}

We experimented with a number of functional forms for the fits to $L(N), L(C)$, and $L(D)$; the power-law fits were qualitatively much more accurate than other functions such as logarithms (see Figure \ref{fig:PoorLogFit}).

For $L(C)$, we do not include small models with only 1 layer in the fit, as the transition from 1 to 2 layers causes a noticable lump in the data. For $L(N)$ we also do not include very small models with only 1 layer in the fit, and we exclude the largest models that have not trained fully to convergence.  Fit parameters change marginally if we do include them, and the trend extrapolates well in both directions regardless.

\subsection{Generalization and Architecture}
\label{sec:DepthVsGeneralization}

In figure \ref{fig:DepthVsGeneralization} we show that generalization to other data distributions does not depend on network depth when we hold the total parameter count fixed.  It seems to depend only on the performance on the training distribution.

\listoffigures
\listoftables

\bibliographystyle{halpha}
\bibliography{bibliography}

\end{document}